\documentclass{article}
\usepackage{preprint}

\usepackage{amsmath,amsfonts,bm}

\def\eqref#1{equation~\ref{#1}}

\def\1{\bm{1}}

\DeclareMathAlphabet{\mathsfit}{\encodingdefault}{\sfdefault}{m}{sl}
\SetMathAlphabet{\mathsfit}{bold}{\encodingdefault}{\sfdefault}{bx}{n}

\usepackage{hyperref}
\usepackage{url}
\usepackage[utf8]{inputenc}
\usepackage[T1]{fontenc}
\usepackage{hyperref}
\usepackage{url}
\usepackage{booktabs}
\usepackage{amsfonts}
\usepackage{amsmath}
\usepackage{soul}

\usepackage{amssymb}
\usepackage{pifont}
\usepackage{bm}
\providecommand{\xmark}{\ding{55}}
\usepackage{amsthm}
\newtheorem{assumption}{Assumption}

\usepackage{nicefrac}
\usepackage{microtype}
\usepackage[table]{xcolor}
\usepackage[pdftex]{graphicx}
\usepackage{subcaption}
\usepackage{multirow}
\usepackage{makecell}
\usepackage{array}
\definecolor{goodc}{HTML}{A8E6A6}
\definecolor{midc}{HTML}{F5E6A6}
\definecolor{badc}{HTML}{F5B9B5}
\definecolor{ncmp}{gray}{0.85}
\newcommand{\cc}[1]{%
  \ifnum#1<50
    \cellcolor{badc!\the\numexpr(50-#1)*2\relax!midc!50!white}%
  \else
    \cellcolor{goodc!\the\numexpr(#1-50)*2\relax!midc!50!white}%
  \fi
}
\newcommand{\best}[1]{\textbf{\boldmath #1}}
\newcolumntype{M}{>{\centering\arraybackslash}p{1.65cm}}

\usepackage{enumitem}
\usepackage{comment}
\usepackage{acronym}
\usepackage{algorithm}
\usepackage{algpseudocode}
\usepackage{afterpage}
\usepackage{placeins}
\usepackage{tikz}
\usepackage{pgfplots}
\pgfplotsset{compat=1.18}
\usepgfplotslibrary{fillbetween}
\usepgfplotslibrary{groupplots}
\usetikzlibrary{arrows.meta, positioning, backgrounds, calc, fit, patterns, decorations.pathreplacing}

\definecolor{backbonefill}{RGB}{180,215,235}    %
\definecolor{repairfill}{RGB}{215,200,235}      %
\definecolor{evaluatorfill}{RGB}{195,222,215}   %
\definecolor{groupfill}{gray}{0.982}
\definecolor{groupdraw}{gray}{0.85}
\tikzset{
  palarr/.style={->, thick, >={Stealth[length=1.6mm]}, line width=0.5pt},
  palbox/.style={rounded corners=2pt, draw, line width=0.6pt, inner sep=0pt,
                 minimum width=9mm, minimum height=7mm, font=\small},
  palbackbone/.style={palbox, fill=backbonefill,
    path picture={
      \begin{scope}[shift={(path picture bounding box.center)}]
        \foreach \sx/\tx in {-0.275/0, 0/0.275}
          \foreach \sy in {-0.21,0,0.21}
            \foreach \ty in {-0.21,0,0.21}
              \draw[line width=0.3pt] (\sx,\sy) -- (\tx,\ty);
        \foreach \x in {-0.275,0,0.275}
          \foreach \y in {-0.21,0,0.21}
            \node[circle, fill=white, draw=black, line width=0.5pt,
                  inner sep=0pt, minimum size=1.1mm] at (\x,\y) {};
      \end{scope}}},
  palrepair/.style={palbox, fill=repairfill},
  palevaluator/.style={palbox, fill=evaluatorfill},
  palgrouptitle/.style={font=\small\bfseries},
  pallbl/.style={font=\small},
  palstep/.style={draw=black!55, dashed, line width=0.3pt, rounded corners=2pt,
                  inner xsep=1mm, inner ysep=1mm},
  legendbox/.style={rounded corners=1pt, draw=black, line width=0.4pt, inner sep=0pt,
                    minimum width=4.5mm, minimum height=3.5mm, font=\scriptsize},
  legendbackbone/.style={legendbox, fill=backbonefill,
    path picture={
      \begin{scope}[shift={(path picture bounding box.center)}]
        \foreach \sx/\tx in {-0.12/0, 0/0.12}
          \foreach \sy in {-0.09,0,0.09}
            \foreach \ty in {-0.09,0,0.09}
              \draw[line width=0.15pt] (\sx,\sy) -- (\tx,\ty);
        \foreach \x in {-0.12,0,0.12}
          \foreach \y in {-0.09,0,0.09}
            \node[circle, fill=white, draw=black, line width=0.3pt,
                  inner sep=0pt, minimum size=0.5mm] at (\x,\y) {};
      \end{scope}}},
  legendrepair/.style={legendbox, fill=repairfill},
  legendevaluator/.style={legendbox, fill=evaluatorfill},
}

\newcommand{\ours}{PAL}

\acrodef{PAL}{Projection-Adaptive Loss}
\acrodef{SOTA}{State-of-the-Art}
\acrodef{ALM}{Adaptive Augmented Lagrangian Method}
\acrodef{LM}{Levenberg--Marquardt}
\acrodef{SQP}{Sequential Quadratic Programming}
\acrodef{IP}{Interior-Point}
\acrodef{VJP}{vector-Jacobian product}
\acrodef{SDF}{Signed Distance Function}
\acrodef{BWB}{Blended-Wing-Body}
\acrodef{VAE}{Variational Autoencoder}
\acrodef{CFRP}{Carbon-Fibre-Reinforced Polymer}
\acrodef{MLP}{Multilayer Perceptron}
\acrodef{OOM}{Out Of Memory}
\acrodef{ACOPF}{AC Optimal Power Flow}

\title{Only Project Once: Projection-Adaptive Loss for Exact Constraint Satisfaction}

\author{%
  Tim Aebersold\textsuperscript{1}\thanks{Corresponding author. \texttt{taebersold@ethz.ch}} \quad
  Soheyl Massoudi\textsuperscript{1} \quad
  Mark Fuge\textsuperscript{1} \\[2pt]
  \textsuperscript{1}ETH Z\"urich \\[4pt]
  {\small\faGithub\ \url{https://github.com/IDEALLab/projection-adaptive-loss}}
}

\usepackage{fontawesome5}
\hypersetup{colorlinks=true,linkcolor=blue!50!black,citecolor=blue!50!black,urlcolor=blue!50!black}

\begin{document}

\maketitle

\begin{abstract}

Precise constraint satisfaction is a prerequisite to deploying learned models in many areas, motivating methods that repair raw neural predictions with a repair procedure. Current methods unroll multiple repair steps in training and softly penalize constraint violations that remain after the unroll. This is compute- and memory-intensive, lacks robustness when the repair fails to converge, and surrenders most of the constraint satisfaction work to the repair. Our central finding is that, contrary to common practice, a single detached projection step suffices in training. We accomplish this with a \acf{PAL}, which uses the constraint residual after this single step to adaptively weigh constraint penalties on the raw prediction. In experiments, \ac{PAL} is the only method that retains virtually perfect feasibility on extremely nonlinear constraints, and matches or outperforms current methods on synthetic and engineering benchmarks. Because it only requires a single detached projection step, \ac{PAL} trains 2.5$\times$ faster than the canonical repair-based method (DC3) on its own \acs{ACOPF} benchmark. \ac{PAL} can also be trained when constraints are expensive to evaluate (e.g., via neural surrogates), a setting where current unrolled methods are memory-intractable.

\end{abstract}

\section{Introduction}

\label{sec:intro}

Precise enforcement of hard constraints remains a key challenge in many applications of machine learning. For example, in atmospheric and fluids modeling, neural surrogates must adhere to conservation laws for stable roll-out over long time horizons~\citep{wattmeyer2023ace,krishnapriyan2021characterizing}. In power systems and robotics, learned policies must reliably conform to operational or safety limits to be deployed~\citep{pan2022deepopf,brunke2022safe}, and in mechanical and aerospace design, generated designs must not violate stress or aerodynamic limits~\citep{hoyer2019neural,bouhlel2020scalable}. Current methods that enforce constraint satisfaction can typically be divided into two categories. \emph{Penalty methods} use fixed or adaptive constraint weights in the training loss and typically do not exactly satisfy constraints. More recent methods apply a \emph{repair procedure} to raw model predictions to ensure constraints are satisfied. 

\begin{figure}[H]
\centering
\resizebox{0.72\textwidth}{!}{%
\begin{tikzpicture}[x=1cm, y=1cm, node distance=3mm]

\node[palbackbone] (bb2) at (0,-1.80) {};
\node[left =4mm of bb2, pallbl] (x2in)  {$z$};
\node[right=4mm of bb2, pallbl] (y2out) {$\hat{y}$};
\draw[palarr] (x2in) -- (bb2);
\draw[palarr] (bb2)  -- (y2out);
\node[palevaluator] (ev2) at (2.55,-1.80) {$\Phi$};
\node[palrepair, right=0.5mm of ev2] (rs2) {$\mathcal{P}$};
\draw[palarr] ([xshift=+2.75mm]ev2.north) to[bend left=45] ([xshift=-2.75mm]rs2.north);
\draw[palarr] ([xshift=-2.75mm]rs2.south) to[bend left=45] ([xshift=+2.75mm]ev2.south);
\node[palstep, inner xsep=1.5mm, inner ysep=2.5mm, fit={(ev2)(rs2)}] (sotastep) {};
\node[below=0.3mm of sotastep.south, font=\small] {$\times N$};
\draw[palarr] (y2out.east) -- (sotastep.west);
\node[pallbl] (yN) at (5.0,-1.80) {$\tilde{y}^{(N)}$};
\draw[palarr] (sotastep.east) -- (yN.west);
\node[pallbl] (loss2) at (7.05,-1.80) {$\mathcal{L} \sim f + w_{c}\,\tilde{c}^{(N)}$};
\draw[palarr] (yN) -- (loss2);

\node[palbackbone] (bb3) at (0,-4.15) {};
\node[left =4mm of bb3, pallbl] (x3in)  {$z$};
\node[right=4mm of bb3, pallbl] (y3out) {$\hat{y}$};
\draw[palarr] (x3in) -- (bb3);
\draw[palarr] (bb3)  -- (y3out);
\node[palevaluator, right=4mm of y3out] (ev_sh) {$\Phi$};
\draw[palarr] (y3out) -- (ev_sh);
\node[right=4mm of ev_sh, pallbl] (chat) {$f,\,\hat{c}$};
\draw[palarr] (ev_sh) -- (chat);
\node[palrepair] (rs3) at (chat |- 0,-5.15) {$\mathcal{P}$};
\node[pallbl,       right=3mm of rs3]     (yN1)     {$\tilde{y}^{(1)}$};
\node[palevaluator, right=3mm of yN1]     (ev_post) {$\Phi$};
\node[pallbl,       right=3mm of ev_post] (ctilde)  {$\tilde{c}^{(1)}$};
\node[pallbl,       right=3mm of ctilde]  (wc_lbl)  {$\boldsymbol{\bar{w}_{c}}$};
\draw[palarr] (rs3)     -- (yN1);
\draw[palarr] (yN1)     -- (ev_post);
\draw[palarr] (ev_post) -- (ctilde);
\draw[palarr] (ctilde)  -- (wc_lbl);
\draw[palarr] (chat.south) -- (rs3.north);
\node[palstep, fit={(rs3)(yN1)(ev_post)(ctilde)(wc_lbl)}] (palstep) {};
\node[below=0.3mm of palstep.south, font=\small]
  {$\times 1$ \ (detached)};
\node[pallbl] (loss3) at ($(wc_lbl |- 0,-4.15) + (-4.5mm,0)$) {$\mathcal{L} \sim f + \boldsymbol{\bar{w}_{c}}\,\hat{c}$};
\draw[palarr] (chat.east) -- (loss3.west);
\draw[palarr] (wc_lbl.north) -- (wc_lbl |- loss3.south);

\begin{scope}[on background layer]
  \filldraw[rounded corners=3pt, fill=groupfill, draw=groupdraw, line width=0.8pt]
    (-1.625, -2.95) rectangle ( 8.625, -0.55);
  \filldraw[rounded corners=3pt, fill=groupfill, draw=groupdraw, line width=0.8pt]
    (-1.625, -6.15) rectangle ( 8.625, -3.22);
\end{scope}
\node[font=\scriptsize, anchor=south east] at (8.5, -2.92) {$w_c$: fixed soft weight};

\node[palgrouptitle] at (3.5, -0.87) {SOTA};
\node[palgrouptitle] at (3.5, -3.54) {PAL (Ours)};

\node[rounded corners=2pt, draw=black!15, line width=0.7pt, inner sep=1mm]
     at (3.5, -6.82) (legendbox) {%
  \begin{tikzpicture}
    \node[legendbackbone]  (lbb) {};
    \node[font=\scriptsize, right=0.5mm of lbb]         (lbb_lbl) {Backbone $N_\theta$};
    \node[legendrepair,    right=5mm of lbb_lbl]  (lrs)     {$\mathcal{P}$};
    \node[font=\scriptsize, right=0.5mm of lrs]         (lrs_lbl) {Repair Step};
    \node[legendevaluator, right=5mm of lrs_lbl]  (lev)     {$\Phi$};
    \node[font=\scriptsize, right=0.5mm of lev]         (lev_lbl) {Evaluator};
  \end{tikzpicture}%
};
\end{tikzpicture}}
\caption{\ours{} vs.\ \ac{SOTA} repair-aware training. From raw model predictions $\hat y$, \ac{SOTA} iterates projector $\mathcal{P}$ and evaluator $\Phi$ for $N$ times per training iteration and then uses a soft penalty on post-repair constraint violation $\tilde c ^{(N)}$. We instead use a single and detached projection step to $\tilde y ^{(1)}$. \ours{} then uses the post-projection residual $\tilde c ^{(1)}$ to adaptively drive $\bar w_c$, which acts as an adaptive penalty on the raw prediction's constraint residual $\hat c$.}
\label{fig:hero}
\end{figure}
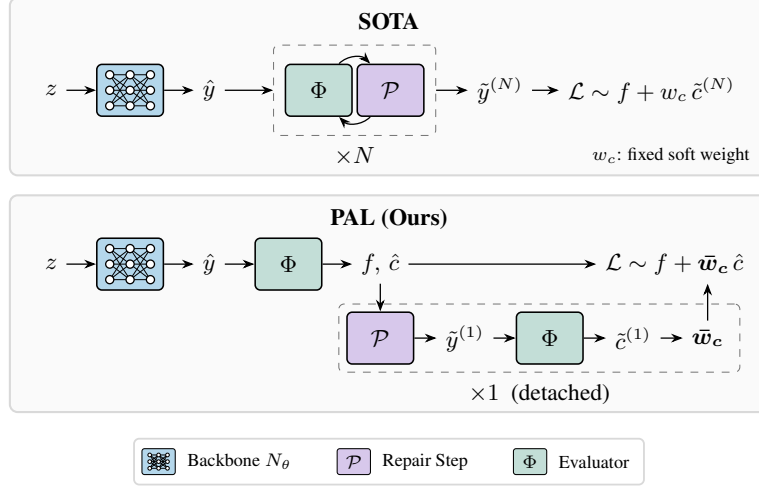

In this latter category, the current predominant approach to train these methods is to unroll multiple repair steps per training iteration. Any residual violation at the end of the unroll is then softly penalized in the loss with a fixed weight, as shown in Fig.~\ref{fig:hero}.

In our view, this has three major drawbacks: (i) Unrolled repair is compute- and memory-intensive, which limits applicability to relatively cheap-to-evaluate constraints (e.g., closed-form expressions). (ii) Current methods softly penalize post-repair constraint residuals. When the projection iterates do not converge (e.g., under extremely non-linear constraints), a soft penalty does not provide enough training pressure to satisfy these constraints. (iii) We believe that the repair should only act as a final clean-up, and the neural net should learn the bulk of constraint satisfaction. Methods that train with unrolled repair tend to transfer the majority constraint satisfaction work to the repair step.

To address these issues, our contributions are as follows:

\begin{enumerate}
\item \textit{Single-Step Repair in Training.} We show that, contrary to the prevailing wisdom, only a \emph{single} detached projection step suffices in training while matching or outperforming multi-step unrolled-repair baselines (\S\ref{sec:synthetic_results}---\S\ref{sec:curvature}). This avoids backpropagation through the repair, and enables tractability in applications where unrolled-repair training is infeasible (\S\ref{sec:engineering_results}).

\item \textit{Loss Formulation.} We propose a \acf{PAL} which enables reliable training with single-step repair. It is agnostic to the choice of projection step (\acs{LM}, \acs{SQP}, \acl{IP}, \ldots). This formulation shifts the feasibility load to the neural net and only treats projection as final clean-up (Fig.~\ref{fig:residual_intuition}) and avoids degradation of feasibility when the projector sequence fails to converge (\S\ref{sec:curvature}).

\end{enumerate}

\section{Preliminaries}

\label{sec:prelim}

\textbf{Problem.} We focus on amortized optimization, which targets settings where instances of the same problem family are solved repeatedly under varying conditions where learned models have previously been shown to predict solutions with optimality comparable to classical solvers at many times lower inference cost~\citep{donti2021dc3, lastrucci2025enforce}. We consider the constrained optimization problem

\[
\min_{y \in \mathbb{R}^n} f(y, z)
\quad \text{s.t.} \quad
h(y, z) = 0,
\qquad
g(y, z) \le 0,
\]

where $y$ is the design variable, $z$ an input condition or learned embedding, $f$ the objective, and $h$ and $g$ the equality and inequality constraints. In the following notation, we suppress $z$ for brevity. We write $c(y) \ge 0$ for the vector of constraint violations, with entries $|h_k(y)|$ for the equalities and $\max(g_k(y), 0)$ for the inequalities, and $\mathcal{M} := \{y : c(y) = 0\}$ for the feasible set.

\textbf{Backbone and Evaluator.} A neural network $N_\theta : z \mapsto \hat y$ produces predictions, and an evaluator $\Phi(y) \mapsto (f, h, g)$ scores a candidate $y$ by returning the objective and constraint values. We require the evaluator $\Phi$ to be differentiable in $y$.

\textbf{Projector.} A projector $\mathcal{P}$ produces iterates $\tilde y^{(i+1)} = \mathcal{P}(\tilde y^{(i)}, c(\tilde y^{(i)}))$ from $\tilde y^{(0)} = \hat y$ toward $\mathcal{M}$. We write $\Delta y^{(i)} := \tilde y^{(i+1)} - \tilde y^{(i)}$ for the repair step's displacement and $R_{i,k} := c_k(\tilde y^{(i)})$ for the constraint residuals at iteration $i$. The projection is computed independently per batch element.

Appendix \ref{app:notation_summary} provides a full notation overview.

\section{Related Work}
\label{sec:related}

We focus on settings where $\Phi$ is expensive and lacks any exploitable algebraic form, but is smoothly differentiable. Approaches enforcing constraints by construction via hand-specified output layers~\citep{beucler2021enforcing} or differentiable optimization layers~\citep{amos2017optnet,agrawal2019differentiable} usually require known structure (linearity, convexity, \ldots) that $\Phi$ does not expose, and are thus out of scope here.

\paragraph{Penalty Methods.} A straightforward approach is to incorporate constraints directly into the training loss: $\mathcal{L}_{c} \sim w_c \cdot c(\hat{y})$, with penalty multipliers $w_c$ either fixed or propagated adaptively~\citep{platt1988constrained, basir2023adaptive}. However, these soft methods often fail to satisfy constraints exactly~\citep{donti2021dc3}. A naive solution is to ``bolt on'' a repair procedure at test-time. However, this can result in poor solution quality after projection because the trained model fails to anticipate the repair operation~\citep{chu2026snarenet,pinet}, a finding we corroborate in our own experiments. Recent approaches therefore use the repair procedure in end-to-end training. We now discuss prominent variants of this idea.

\paragraph{Repair Methods.}
Projection-based methods evaluate constraint residuals $c$ and their Jacobian $J$ to perform a linearized quasi-orthogonal projection from $(\tilde{y}^{(i)})$ onto the feasible set $\mathcal{M}$. Because the linearization introduces higher-order error, multiple projection steps are typically used at both training and inference time, often with auto-stopping once feasibility is reached. Recent examples are HardNet~\citep{min2024hardnet} (restricted to linear constraints), $\Pi$net~\citep{pinet} (restricted to analytically known convex constraints), ENFORCE~\citep{lastrucci2025enforce}, or SnareNet~\citep{chu2026snarenet}, which uses \ac{LM} projection, improving projection stability when the constraint Jacobian is near-degenerate. We use a similar projector (see Eq.~\ref{eq:pal_projection_loop}), albeit with adaptive damping according to \citet{yamashita2001rate}. To solve the issue of slow convergence in early training, some methods use adaptive constraint relaxation in early training~\citep{chu2026snarenet}, or warm-up using plain penalty methods~\citep{lastrucci2025enforce}. DC3~\citep{donti2021dc3} decomposes optimization variables into independent ($\hat{y}_\text{indep}$) and dependent ($y_\text{dep}$) subsets, predicting the former via the neural backbone and recovering the latter via projection to satisfy equality constraints, with inequalities handled via gradient descent. DC3 requires that equality constraints have known dependence on $y_{\text{dep}}$. Instead of projection, FSNet~\citep{nguyen2025fsnet} employs gradient descent to minimize $\|h(y^{(i)})\|^2 + \|g(y^{(i)})\|^2$ at repair step $i$. FSNet carries formal convergence guarantees, and unlike projection methods, avoids full Jacobian construction, which scales favorably with constraint count. Its core limitation is iteration count. It requires more repair steps $N$ than iterative projection methods, making backpropagation memory-intensive. L-BFGS~\citep{liu1989lbfgs} with early stopping and truncated unrolled differentiation mitigate but do not resolve this. 

\paragraph{Gap.}

Existing methods (DC3, SnareNet, ENFORCE, FSNet, \ldots) as shown under \acs{SOTA} in Fig.~\ref{fig:hero} share three limitations. First, they unroll multiple repair steps per training iteration and backpropagate through them, which is intractable for very expensive evaluators $\Phi$. Second, they train on post-repair constraint residuals using soft penalties with fixed scalar weights $w_c, w_d$:
  \[
      \mathcal{L} = f(\tilde y) + w_d \|\hat y - \tilde y^{(N)}\|^2 + w_c \|c(\tilde y^{(N)})\|^2 .
  \] 

When the unrolled projector drives $c(\tilde y^{(N)}) \approx 0$ during training, the constraint loss term ($w_c$) exerts no pressure on the backbone and the repair carries all of the feasibility work as is shown later in Fig.~\ref{fig:residual_intuition}. When the unroll does not converge, on the other hand, the residual that remains competes with $f(\tilde y)$ through the fixed weight $w_c$, and feasibility degrades as we show in \S\ref{sec:curvature}.

\section{Method}
\label{sec:method}

We now specify the adaptive penalty update and loss formulation that form our core contribution in detail. The procedure is shown with a damped \ac{LM} repair step (which is described in more detail in Appendix \ref{app:projector}, but can be used analogously with \acs{SQP} and \ac{IP}.

\afterpage{%
\begin{algorithm}[t]
\caption{\ours{} training.}
\label{alg:pal_train}
\begin{algorithmic}[1]
\Require Backbone $N_\theta$, Evaluator $\Phi$, Epochs $T$, Hyperparameters $\tau_c, \tau_d, \gamma, D_{\max}, \delta, \eta$
\State Initialize $\theta$;\ \  $\bar w_{c,k}, \bar w_d \gets 0$.
\For{$t = 1, \dots, T$}
  \State $\hat y \gets N_\theta$ \Comment{Neural backbone prediction}
  \State $c(\hat y),\, f(\hat y) \gets \Phi(\hat y)$ \Comment{Evaluator forward pass}
  \State $J \gets \partial c / \partial \hat y$ \Comment{$K$ reverse-mode \acs{VJP}s through $\Phi$}
  \State $\tilde y^{(1)} \gets \textsc{Project}\big(\hat y,\, J,\, c(\hat y);\, \delta\big)$ \Comment{Single-step projection with LM, SQP, IP, ...}
  \State $\mathcal{L} \gets \textsc{Loss}\big(\hat y,\, \tilde y^{(1)},\, c(\hat y),\, \bar w_{c,k},\, \bar w_d\big)$ \Comment{Eq.~\ref{eq:pal_loss}}
  \State $\theta \gets \theta - \eta\, \nabla_\theta \mathcal{L}$ \Comment{Training step with LR $\eta$}
  \State $c(\tilde y^{(1)}) \gets \Phi_c(\tilde y^{(1)})$ \Comment{Detached evaluator forward}
  \State $\bar w_{c,k}, \bar w_d \gets \textsc{LogGapUpdate}\big(\bar w_{c,k}, \bar w_d,\, c(\tilde y^{(1)}),\, \tilde y^{(1)} - \hat y;\, \tau_c, \tau_d, \gamma, D_{\max}\big)$ \Comment{Eq.~\ref{eq:pal_weight_update}}
\EndFor
\State \Return $\theta$
\end{algorithmic}
\end{algorithm}%
}

\subsection{\ours{} Training}
\label{sec:pal_training}

Algorithm~\ref{alg:pal_train} summarizes \ours{} training. We now discuss the three subroutines \textsc{Project}, \textsc{Loss} (Eq.~\ref{eq:pal_loss}), and \textsc{LogGapUpdate} (Eq.~\ref{eq:pal_weight_update}).

\paragraph{Projection.} 

Our core insight is that, in training, it is sufficient to apply a \emph{single} repair step instead of a full unroll. Given a prediction $\hat y \sim N_\theta$, we perform a single forward pass on the evaluator, $\Phi(\hat y) \mapsto c(\hat y), f(\hat y)$, and $K$ backward passes to assemble the Jacobian $J$. We then perform a single projection iterate. In our experiments, we use \acl{LM} with damping according to \citet{yamashita2001rate} as the default. \ours{} is also compatible with other projectors (e.g., \ac{SQP} or \acf{IP} as described in Appendix~\ref{app:repair_variants}). Note that at \emph{test time}, we use up to 10 projection steps with early stopping, just as our baselines do.

\paragraph{Loss.} After this single projection step, we compute the training loss as follows

\begin{equation}
\mathcal{L}(\hat y, \tilde y^{(1)})
=
{f(\hat y)} 
+ \sum_{k=1}^{K} \bar w_{c,k} \, c_k(\hat y)
+ \bar w_d\, \|\Delta y^{(1)}\|,
\label{eq:pal_loss}
\end{equation}

where we only utilize gradient signal from \emph{pre-projection} quantities in training: the pre-projection constraint violations $c(\hat y)$, displacement $\|\Delta y^{(1)}\| := \|\tilde y^{(1)} - \hat y\|$ with $\tilde y^{(1)}$ detached from the gradient graph, and $f(\hat y)$. We use $f(\hat y)$ rather than $f(\tilde y^{(1)})$ because it progresses more reliably in early training (see Appendix~\ref{app:loss_placement} for further details).

\paragraph{Projection-Adaptive Penalty Weight Update.} To compute the penalty weights  $\bar w_{c,k}$ and $\bar w_d$, we use two post-projection metrics which directly indicate the quality of the repair step:

\begin{enumerate}
    \item The nonlinear residual after a single projection step $R_{1,k}$. It serves as a proxy for how easy constraint $k$ is to repair. A perfectly linear, uncoupled constraint, for example, can be driven to $R_{1,k}\approx0$ in a single step. A coupled, nonlinear constraint might, even for small pre-projection violations $c(\hat y)$, be difficult to project, leaving non-zero $R_{1,k}$.
    \item The displacement $\|\Delta y^{(1)}\|$. It is the distance in $y$ the first repair step takes. For the repair to act only as final clean-up, we require $\|\Delta y^{(1)}\|$ to be small.
\end{enumerate}

In training, both these quantities should evolve to be within thresholds $\tau_d$ (displacement) and $\tau_c$ (constraint residual). To achieve $\|\Delta y^{(1)}\| < \tau_d$ and $R_{1,k} < \tau_c$ for all $k$, we use the following adaptive penalty scheme: 

\begin{equation}
\label{eq:pal_weight_update}
\begin{aligned}
\bar w_{c,k} &\;\leftarrow\; \Big(\bar w_{c,k} + \gamma\,\min\!\big(\log_{10}\!\tfrac{R_{1,k}}{\tau_c},\, D_{\max}\big)\Big)_{+} \\
\bar w_d   &\;\leftarrow\; \Big(\bar w_d   + \gamma\,\min\!\big(\log_{10}\!\tfrac{\|\Delta y^{(1)}\|}{\tau_d},\, D_{\max}\big)\Big)_{+}
\end{aligned}
\end{equation}

where $\gamma$ is the penalty growth rate and $D_{\max}$ is the per-step growth cap (in decades).  The update acts as follows: if any $R_{1,k} > \tau_c$, i.e., constraint $k$ is hard to project, $\bar w_{c,k}$ grows. If $\|\Delta y^{(1)}\| > \tau_d$, i.e., if the repair step displacement is too large, $\bar w_d$ grows. On the other hand, where $\|\Delta y^{(1)}\| < \tau_d$ or $R_{1,k} < \tau_c$, the respective weights $\bar w_d$ and $\bar w_{c,k}$ will gradually decay, freeing learning signal towards the objective. We use log-scaling to make the updates of $\bar w_{c,k}$ and $\bar w_d$ meaningful across constraints with heterogeneous native scales and to avoid vanishing weight growth for $\tau_{c,k} < R_{1,k} \ll 1$. The growth rate $\gamma$ sets the adaptation timescale, and $D_{\max}$ promotes stability in early training. We initialize $\bar w_d = 0$ and $\bar w_{c,k} = 0$ and clip to positive values to prevent rewarding constraint violations. In practice, $\tau_c$ is set to the desired constraint tolerance (which is typically known in advance for any application), i.e., we aim for the first projection step to reach feasibility in training. Note that $\tau_c$ does not affect the precision reachable at test time. Despite single-step repair in training, constraint violation can be driven to machine precision given enough projection steps, as we show in Appendix~\ref{app:precision}. We characterize the influence of the newly introduced hyperparameters in Ablations (\S~\ref{sec:ablations}), where a wide band of settings yields comparable performance. 

\begin{figure}[!b]
\begin{center}
\includegraphics{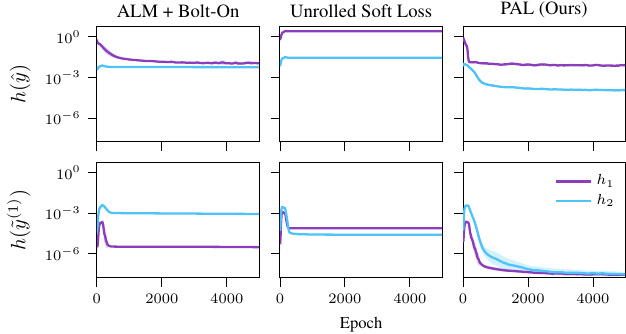}\hspace*{3.5em}
\end{center}
\caption{Two-constraint toy problem where $h_1$ is stiff but linear and therefore trivial to project, and $h_2$ is non-stiff but nonlinear and difficult to project (more details in Appendix~\ref{app:intuition_setup}). \textbf{Top:} constraint violations before projection $h(\hat y)$. \textbf{Bottom:} constraint violations after the first projection step $h(\tilde y^{(1)})$, equivalent to $R_{1,k}$. \textbf{Left:} \ac{ALM} (adaptive penalty method) trains without projection and penalizes constraints only according to pre-projection magnitude $h(\hat y)$, i.e., cannot distinguish constraints by repair difficulty. \textbf{Center:} Under soft-loss training (SnareNet, ENFORCE, FSNet, \ldots), the projector drives $h(\tilde y^{(N)}) \approx 0$ in training so that constraint pressure on the neural backbone is low, resulting in high violations before and after the first projection step. \textbf{Right:} \ours{} exerts the right amount of training pressure onto $h(\hat y)$ (and thus the neural backbone) to drive post-projection residuals $h_k(\tilde y^{(1)})$ below threshold $\tau_c$ (here $10^{-7}$).}
\label{fig:residual_intuition}
\end{figure}

We observe that the per-constraint terms $\bar w_{c,k}\, c_k(\hat y)$ and the displacement term $\bar w_d\, \|\Delta y^{(1)}\|$ are often aligned, i.e., reducing one also reduces the other. We show in our Ablations (\S~\ref{sec:ablations}) that using only one of the two often suffices, but we nevertheless employ both simultaneously as a two-layered safety net, e.g., for stiff constraints ($\|\partial c_k / \partial y\| \gg 1$) for which a small $\|\Delta y^{(1)}\|$ can still leave a large residual $R_{1,k}$ due to damping and nonlinearity.  The displacement penalty $\bar w_d \|\Delta y^{(1)}\|$ also ensures $f(\hat y) \approx f(\tilde y^{(1)})$ once the projection step is small, so the backbone's objective is a reliable proxy for the post-projection objective without backpropagating $f$ through the repair step. 

In Fig.~\ref{fig:residual_intuition}, we show how our loss formulation shifts feasibility work between learning (neural backbone $N_\theta$) and repair ($\mathcal{P}$).\footnote{Side note on representation: We hypothesize that pushing the neural backbone $N_\theta$ to predict near-feasible $\hat y$ could also encourage the learned embeddings of $N_\theta$ to approximately encode the constraint manifold $\mathcal{M}$, a property other repair-aware losses do not actively encourage.} 

\section{Experiments}

\label{sec:experiments}

\paragraph{Baselines.} We compare against ENFORCE~\citep{lastrucci2025enforce}, FSNet~\citep{nguyen2025fsnet}, SnareNet~\citep{chu2026snarenet}, and DC3~\citep{donti2021dc3}where applicable. We also include \ac{ALM}~\citep{basir2023adaptive} (an adaptive penalty method) with Bolt-On repair, i.e., the repair step is only used at test time. The ALM+Bolt-On baseline uses the same \ac{LM} repair step with adaptive damping as \ours{}. All other methods use their own repair methods (DC3 uses undamped completion, FSNet uses gradient descent, SnareNet uses damped \ac{LM} with fixed damping, etc.).

\paragraph{Metrics.} All experiments report the \emph{Objective} as $f(\tilde y^{(N)})$ and \emph{Feasibility (\%)}, the fraction of evaluation samples whose final output $\tilde y^{(N)}$ satisfies all constraints to within the benchmark-specific tolerance (typically $10^{-3}$ or $10^{-4}$). To compare compute, we report \emph{Evaluator forward passes} $N_\Phi^{\mathrm{f}}$ and \emph{Evaluator backward passes} $N_\Phi^{\mathrm{b}}$, normalized by batch size.

\subsection{Comparison on Synthetic Stress-Test Benchmarks}
\label{sec:synthetic_results}

In Table~\ref{tab:methods-sweep-synthetic} we compare results on six benchmarks, each adding a specific stressor. \texttt{S1} is the simplest with a single nonlinear equality constraint and a non-convex objective. \texttt{S2} adds inequality constraints and a non-convex chained Rosenbrock objective. \texttt{S3} features an ill-conditioned objective with constraint coupling. \texttt{S4} has nonlinear coupling between two variable blocks. \texttt{S5} is an overdetermined nonlinear equality system, whose feasible set reduces to a single point per condition $z$. \texttt{S6} is high-dimensional with $K = 90$ constraints, including redundant inequalities. By construction, projection alone cannot reach the constrained optimum on \texttt{S1}--\texttt{S4} and \texttt{S6}. All synthetic benchmarks have a known constrained optimum $f^\star = 0$. More details on \texttt{S1}--\texttt{S6} can be found in Appendix~\ref{app:synthetic_setups}. All methods are hyperparameter-tuned with Bayesian optimization, which is described in further detail in Appendix~\ref{app:hp_tuning}.

All baselines reach high feasibility except DC3 due to its undamped completion step, which fails on ill-conditioned problems. With \ours{} this failure mode can be replicated when the \ac{LM} projection is used without damping (see App. \ref{app:ablations}). 

\ours{} is competitive with or outperforms repair-aware baselines on optimality across \texttt{S1}--\texttt{S6}. ENFORCE is the best baseline, except on \texttt{S5} where it is structurally inapplicable. ALM+Bolt-On reaches near-perfect feasibility but very poor optimality. Since the projector is held constant between \ours{} and ALM+Bolt-On, this isolates \ac{ALM}'s lack of repair-awareness as the cause for ALM+Bolt-On's poor performance, which corroborates prior observations that training without repair-awareness can be counterproductive~\citep{chu2026snarenet,pinet}. SnareNet underperforms on objective across the sweep. FSNet performs well, except on \texttt{S5}. 

Per training step, \ours{} requires two forward passes through $\Phi$ (at $\hat y$ and $\tilde y^{(1)}$) and $K+1$ backward passes at $\hat y$, of which $K$ build the Jacobian $J$. In practice, we cache $J$ and reuse it to backpropagate the training loss (Eq.~\ref{eq:pal_loss}) without a second pass through $\Phi$. \ours{} uses the fewest evaluator passes of the repair-aware methods for problems with few constraints (moderate $K$) such as \texttt{S1}. At, high $K$ (e.g., \texttt{S6}) FSNet, which sidesteps Jacobian assembly, requires fewer backward passes. Among projection methods, \ours{} requires fewest. ALM+Bolt-On is cheapest overall but achieves poor results. Appendix~\ref{app:compute_scaling} elaborates further on compute scaling.

\begin{table}[t]
\centering
\footnotesize
\setlength{\tabcolsep}{3pt}
\setlength{\aboverulesep}{0.15ex}
\setlength{\belowrulesep}{0.2ex}
\renewcommand{\arraystretch}{1.05}
\caption{Synthetic stress-test experiments. Means and standard deviations across 10 seeds. ``---'': failure on 2 or more seeds. Forward and backward evaluator passes $N_\phi ^f$ and $N_\phi ^b$ per training step are only shown for extreme cases \texttt{S1} ($K=1$) and \texttt{S6} ($K=90$), with full results in Table~\ref{tab:methods-sweep-synthetic-full}.}
\label{tab:methods-sweep-synthetic}
\begin{tabular}{@{}ll *{6}{M}@{}}
\toprule
 & \textbf{Bench.}
  & \makecell{\textbf{ALM+}\\\textbf{Bolt-On}} & \textbf{ENFORCE} & \textbf{DC3} & \textbf{FSNet} & \textbf{SnareNet} & \makecell{\textbf{PAL}\\{}[Ours]} \\
\midrule
\multirow{7}{*}{\rotatebox[origin=c]{90}{Feas.\ (\%)}}
  & \texttt{S1} & \cc{100} $100\pm0$ & \cc{100} $100\pm0$ & — & \cc{100} $100\pm0$ & \cc{100} $100\pm0$ & \cc{100} $100\pm0$ \\
  & \texttt{S2} & \cc{100} $100\pm0$ & \cc{100} $100\pm0$ & \cc{84} $97\pm7$ & \cc{100} $100\pm0$ & \cc{100} $100\pm0$ & \cc{100} $100\pm0$ \\
  & \texttt{S3} & \cc{100} $100\pm0$ & \cc{100} $100\pm0$ & \cc{100} $100\pm0$ & \cc{100} $100\pm0$ & \cc{100} $100\pm0$ & \cc{100} $100\pm0$ \\
  & \texttt{S4} & \cc{98} $100\pm1$ & \cc{100} $100\pm0$ & — & \cc{100} $100\pm0$ & \cc{100} $100\pm0$ & \cc{100} $100\pm0$ \\
  & \texttt{S5} & \cc{100} $100\pm0$ & — & — & \cc{4} $51\pm31$ & \cc{100} $100\pm0$ & \cc{100} $100\pm0$ \\
  & \texttt{S6} & \cc{100} $100\pm0$ & \cc{100} $100\pm0$ & — & \cc{100} $100\pm0$ & \cc{100} $100\pm0$ & \cc{100} $100\pm0$ \\
\cmidrule(lr){2-8}
  & \textit{Mean} & \cc{100} \best{$100$} & — & — & \cc{66} $92$ & \cc{100} \best{$100$} & \cc{100} \best{$100$} \\
\midrule
\multirow{7}{*}{\rotatebox[origin=c]{90}{Obj.}}
  & \texttt{S1} & \cc{44} $0.07\pm0.12$ & \cc{100} $0.00\pm0.00$ & — & \cc{100} $0.00\pm0.00$ & \cc{12} $0.78\pm0.10$ & \cc{55} $0.03\pm0.04$ \\
  & \texttt{S2} & \cc{26} $0.27\pm0.17$ & \cc{100} $0.00\pm0.00$ & \cc{100} $0.00\pm0.00$ & \cc{100} $0.00\pm0.00$ & \cc{33} $0.16\pm0.00$ & \cc{100} $0.00\pm0.00$ \\
  & \texttt{S3} & \cc{100} $0.00\pm0.00$ & \cc{100} $0.00\pm0.00$ & \cc{100} $0.00\pm0.00$ & \cc{100} $0.00\pm0.00$ & \cc{70} $0.01\pm0.00$ & \cc{100} $0.00\pm0.00$ \\
  & \texttt{S4} & \cc{0} $19.47\pm5.25$ & \cc{70} $0.01\pm0.01$ & — & \cc{100} $0.00\pm0.00$ & \cc{41} $0.09\pm0.01$ & \cc{100} $0.00\pm0.00$ \\
  & \texttt{S5} & \cc{100} $0.00\pm0.00$ & — & — & \cc{0} $4.41\pm2.16$ & \cc{100} $0.00\pm0.00$ & \cc{100} $0.00\pm0.00$ \\
  & \texttt{S6} & \cc{35} $0.14\pm0.03$ & \cc{100} $0.00\pm0.00$ & — & \cc{100} $0.00\pm0.00$ & \cc{16} $0.60\pm0.06$ & \cc{55} $0.03\pm0.01$ \\
\cmidrule(lr){2-8}
  & \textit{Mean} & \cc{0} $3.32$ & — & — & \cc{13} $0.73$ & \cc{26} $0.27$ & \cc{70} \best{$0.01$} \\
\midrule
\multirow{2}{*}{\rotatebox[origin=c]{90}{$N_\Phi^{\mathrm{f}}$}}
  & \texttt{S1} & \cc{100} $1.0\pm0.0$ & \cc{53} $6.2\pm0.1$ & — & \cc{0} $72\pm77$ & \cc{9} $35.0\pm5.0$ & \cc{82} $2.0\pm0.0$ \\
  & \texttt{S6} & \cc{100} $1.0\pm0.0$ & \cc{50} $7.1\pm0.1$ & — & \cc{30} $15.5\pm0.0$ & \cc{0} $81.1\pm2.9$ & \cc{82} $2.0\pm0.0$ \\
\midrule
\multirow{2}{*}{\rotatebox[origin=c]{90}{$N_\Phi^{\mathrm{b}}$}}
  & \texttt{S1} & \cc{100} $1.0\pm0.0$ & \cc{64} $4.1\pm0.1$ & — & \cc{39} $11.0\pm5.6$ & \cc{10} $33.6\pm5.0$ & \cc{82} $2.0\pm0.0$ \\
  & \texttt{S6} & \cc{100} $1.0\pm0.0$ & \cc{0} $128\pm2$ & — & \cc{49} $7.2\pm0.0$ & \cc{0} $253\pm12$ & \cc{0} $91.0\pm0.0$ \\
\bottomrule
\end{tabular}
\end{table}

\subsection{Case Studies in Science and Engineering}
\label{sec:engineering_results}

Table~\ref{fig:engineering_results} shows results for three engineering case studies. \texttt{E1} optimizes aircraft designs using expensive neural \ac{SDF} geometry representation with surrogates for structural properties and aerodynamics~\citep{sung2025blendednet}. \texttt{E2} (urban planning) uses a video-diffusion model~\citep{perini2026windinet} to evaluate pedestrian wind comfort in cities. Both of these benchmarks are compute-intensive. \texttt{E3} is a variant of the \ac{ACOPF} benchmark that motivated DC3 \cite{donti2021dc3}, but our variant has more constraints (576 vs. 256). This benchmark is difficult due to the high number of constraints and nonlinear, rank-deficient equalities. Detailed descriptions of \texttt{E1}---\texttt{E3} can be found in Appendix~\ref{app:engineering_setups}. 

On both \texttt{E1} and \texttt{E2}, the unroll-repair baselines are \ac{OOM} even at a mini-batch size of 1, except SnareNet on \texttt{E1} which \ac{OOM}'s at a mini-batch size of 8. With our 4$\times$GH200, \ours{} can be used with batches of 32 on \texttt{E1}, and 16 on \texttt{E2}. On \texttt{E1} and \texttt{E2}, the penalty methods are the least memory-hungry, but fall short in optimality and feasibility, respectively. On \texttt{E3}, DC3's flagship benchmark, \ours{} leads in both feasibility and optimality while training $2.5\times$ faster. We reproduce the violation magnitudes of DC3 within their reported precision~\citep{donti2021dc3}. On \texttt{E3}, the penalty-based methods train in minutes because they avoid $K$ backward passes, but fall short catastrophically in constraint satisfaction.

\newlength{\engrimgh}\setlength{\engrimgh}{1.82cm}
\newlength{\engrimgcell}\setlength{\engrimgcell}{3.4cm}

\newcommand{\prov}[1]{\textcolor{blue}{#1}}
\newcommand{\xxx}{\textcolor{red}{XXX}}
\newcommand{\notrun}{\textcolor{black!40}{$\cdot$}}
\newcommand{\covglyphlw}{0.5pt}
\newcommand{\covglyphh}{1.3ex}
\DeclareRobustCommand{\nonconv}{\raisebox{0.1ex}{\begin{tikzpicture}[baseline=0pt,line width=\covglyphlw,line join=round]\draw (0,0) -- (\covglyphh,0) -- (0.5*\covglyphh,0.866*\covglyphh) -- cycle;\end{tikzpicture}}}
\DeclareRobustCommand{\inapplic}{\raisebox{0.1ex}{\begin{tikzpicture}[baseline=0pt,line width=\covglyphlw,line join=round]\draw (0.5*\covglyphh,0) -- (\covglyphh,0.5*\covglyphh) -- (0.5*\covglyphh,\covglyphh) -- (0,0.5*\covglyphh) -- cycle;\end{tikzpicture}}}

\newcommand{\engrimgcellbox}[1]{\begin{minipage}[b][\engrimgh][c]{\engrimgcell}\centering #1\end{minipage}}
\newcommand{\engrimgEone}{\engrimgcellbox{\raisebox{0.75mm}{\hspace*{-1.5mm}\reflectbox{\includegraphics[trim=72 27 6 32,clip,height=0.88\engrimgh,keepaspectratio]{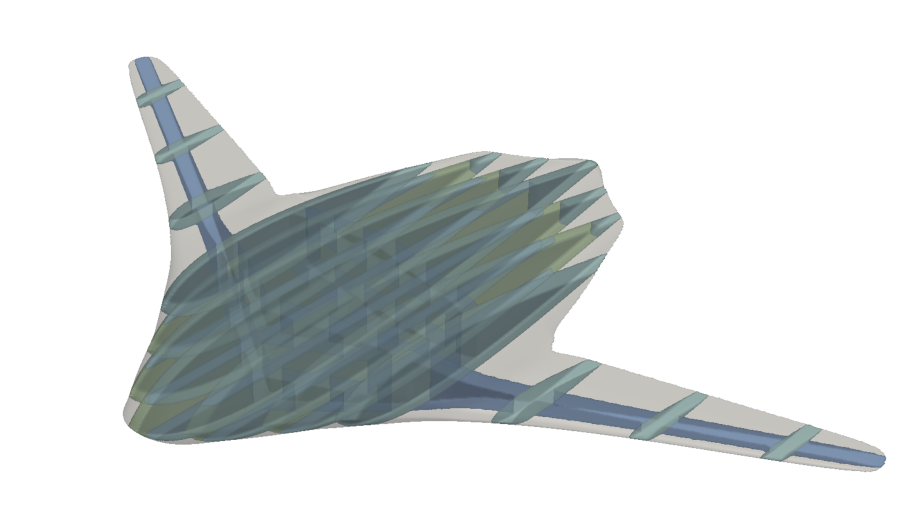}}\hspace*{1.5mm}}}}
\newcommand{\engrimgEtwo}{\engrimgcellbox{\begin{tikzpicture}\node[anchor=south west,inner sep=0] (img) at (0,0) {\includegraphics[trim=10 6 8 50,clip,height=\engrimgh,keepaspectratio]{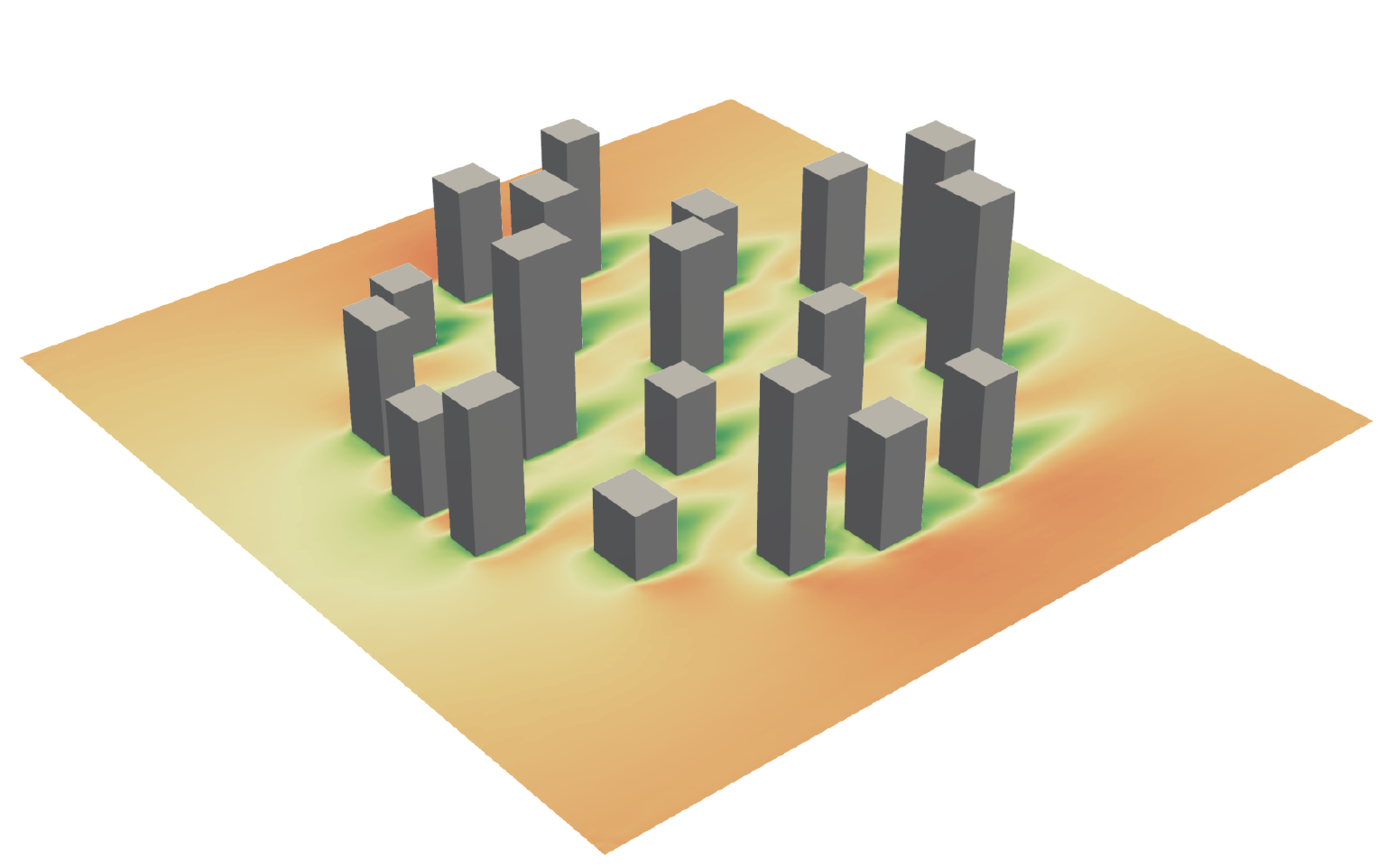}};\fill[white] (img.south west) rectangle ++(4.5mm,4.5mm);\end{tikzpicture}}}
\newcommand{\engrimgEthree}{\engrimgcellbox{\resizebox{!}{0.98\engrimgh}{\input{figures/e3_acopf_oneline}}}}
\newcommand{\engrgh}[1]{\multicolumn{3}{c}{\footnotesize\bfseries #1}}

\begin{table}[t]
\centering
\caption{Case studies in science and engineering. Mem: peak GPU memory (GB) at \texttt{batchsize=1}. Time: training wall-clock. 10 seeds each, IPOPT 5 restarts $\times$ 64 queries. $\ast$: Maximum constraint violation. Bold marks the best value among feasible methods. $\ddagger$: 49 GB on 3 seeds and >96 GB (OOM) on 7 seeds in the first two training epochs. $\S$: \texttt{batchsize=8}. $\dagger$: 0\% feasibility. \nonconv: Non-convergence. \inapplic: Structurally inapplicable. Baseline failure modes are discussed in Appendix~\ref{app:failure-modes}.}
\begin{minipage}{\textwidth}\centering
\footnotesize
\setlength{\tabcolsep}{2pt}
\resizebox{\textwidth}{!}{%
\begin{tabular}{@{}l @{\hspace{5pt}}ccc @{\hspace{5pt}}ccc @{\hspace{5pt}}ccc@{}}
& \multicolumn{3}{c}{\engrimgEone} & \multicolumn{3}{c}{\engrimgEtwo} & \multicolumn{3}{c}{\engrimgcellbox{\resizebox{!}{0.98\engrimgh}{\input{figures/e3_acopf_oneline}}}} \\[2pt]
\toprule
& \engrgh{\texttt{E1} Unmanned Aircraft} & \engrgh{\texttt{E2} Urban Planning} & \engrgh{\texttt{E3} \acs{ACOPF} (IEEE-57)} \\
\cmidrule(lr){2-4}\cmidrule(lr){5-7}\cmidrule(lr){8-10}
Method
 & Obj$_{\times10^{-2}}$ & Feas\% & Mem
 & Obj$_{\times10^{1}}$  & Feas\% & Mem
 & Obj$_{\times10^{4}}$  & Viol$^{\ast}_{\times10^{-2}}$ & Time \\
\midrule
\ours
 & $\mathbf{-3.8{\pm}0.1}$ & $\mathbf{100{\pm}0}$ & $4.4$
 & $-2.71{\pm}0.02$ & $\mathbf{98{\pm}3}$ & $25.7$
 & $\mathbf{4.28{\pm}0.05}$ & $\mathbf{0.15{\pm}0.08}$ & $\mathbf{3\text{h}50\text{m}}$ \\
\specialrule{0.02em}{\aboverulesep}{\belowrulesep}
ALM
 & $-2.6{\pm}1.5$ & $99{\pm}1$ & $\mathbf{4.1}$
 & $\mathbf{-2.74{\pm}0.01}$ & $58{\pm}33$ & $\mathbf{15.7}$
 & $3.60{\pm}2.00^{\dagger}$ & $1100{\pm}731$ & $1\text{m}18\text{s}^{\dagger}$ \\
ALM+Bolt-On
 & $-2.6{\pm}1.5$ & $\mathbf{100{\pm}0}$ & $\mathbf{4.1}$
 & $-2.73{\pm}0.01$ & $69{\pm}25$ & $\mathbf{15.7}$
 & $4.52{\pm}0.26^{\dagger}$ & $38{\pm}56$ & $1\text{m}18\text{s}^{\dagger}$ \\
\specialrule{0.02em}{\aboverulesep}{\belowrulesep}
DC3
 & \inapplic & \inapplic & \inapplic
 & \inapplic & \inapplic & \inapplic
 & $4.34{\pm}0.06$ & $0.53{\pm}0.23$ & $9\text{h}52\text{m}$ \\
FSNet
 & OOM & OOM & $>96$
 & OOM & OOM & $>96$
 & $1.15{\pm}0.46^{\dagger}$ & $>10^{17}$ & $3\text{h}29\text{m}^{\dagger}$ \\
SnareNet
 & {OOM} & {OOM} & {$>96^\S$}
 & OOM & OOM & $>49^{\ddagger}$
 & \nonconv & \nonconv & \nonconv \\
ENFORCE
 & OOM & OOM & $>96$
 & OOM & OOM & $>96$
 & \nonconv & \nonconv & $8\text{h}54\text{m}$ \\
\specialrule{0.02em}{\aboverulesep}{\belowrulesep}
IPOPT
 & $-3.7{\pm}0.1$ & $\mathbf{100}$ & ---
 & \nonconv & \nonconv & \nonconv
 & \multicolumn{3}{c}{see \citet{donti2021dc3}} \\
\bottomrule
\end{tabular}}
\end{minipage}

\label{fig:engineering_results}
\end{table}

Overall, \ours{} is the only method that combines high feasibility with a competitive objective on all three engineering benchmarks. Regarding the lacking robustness of unroll baselines, we hypothesize on the varying causes of failures (IPOPT on \texttt{E2} and ENFORCE, FSNet and SnareNet on \texttt{E3}) in Appendix~\ref{app:failure-modes}. These results also show that, as one might expect, unrolled-repair methods are not tractable on problems with expensive constraint simulators (\texttt{E1} and \texttt{E2}).

\subsection{Highly Nonlinear Constraints}
\label{sec:curvature}

\begin{figure}[b]
\centering
\input{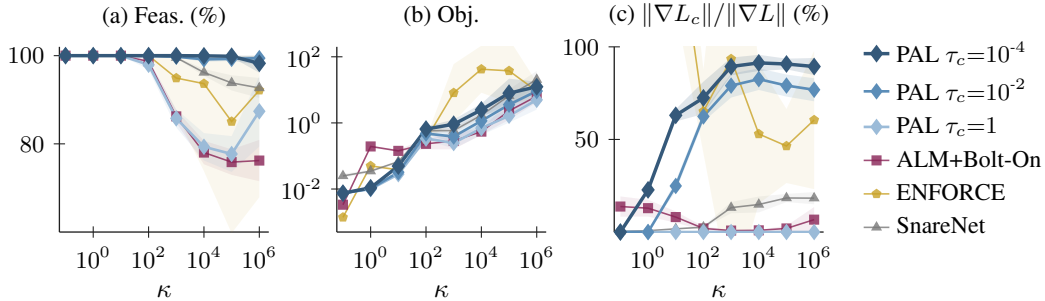}
\caption{Feasibility, objective and constraint training pressure $\|\nabla_\theta L_c\|/\|\nabla_\theta L\|$ over curvature $\kappa$. \ours{} retains best-in-class feasibility at high curvature despite a single repair step, because its adaptive loss raises training pressure in response to first-step residuals.}
\label{fig:curvature_sweep}
\end{figure}

Figure  \ref{fig:curvature_sweep} shows how \ours{} and unrolled soft-loss baselines behave under extreme constraint nonlinearity on a synthetic problem with tunable curvature $\kappa$. In addition to feasibility and optimality, we report the ratio $\|\nabla_\theta L_c\|/\|\nabla_\theta L\|$ in late training. It represents the share of overall training pressure exerted by the constraint loss. Details on the problem setup and full results are shown in Appendix~\ref{app:curvature}. 

\ours{} reaches optimality on par with the full-unroll methods (Fig.~\ref{fig:curvature_sweep}b) and retains the highest feasibility of all methods at extreme curvature of $\kappa=10^6$ (Fig.~\ref{fig:curvature_sweep}a). \ours{} achieves this because its adaptive weight update applies increasing pressure to reduce pre-projection constraints when the linear first projection step leaves large residuals, as shown in Fig.~\ref{fig:curvature_sweep}c. The soft-loss baselines do not possess this mechanism: at high curvature, when the unrolled projection steps alone fail to converge to feasible solutions, constraints and optimality compete on equal footing due to the fixed-weight summed loss. This can be seen also by the lower $\|\nabla_\theta L_c\|/\|\nabla_\theta L\|$ in  Fig.~\ref{fig:curvature_sweep}c. For \ours{} at $\tau_c=1$, the adaptive weight growth is effectively disabled and we observe similar degradation.

These results demonstrate that, despite the single-step training, \ours{} is \textbf{not} limited to near-linear constraints. In fact, it is more robust in cases where unrolled projection alone is not sufficient for reliable constraint satisfaction due to the use of our projection-adaptive weight loss.

\subsection{Ablation Studies}

\label{sec:ablations}

In Appendix~\ref{app:ablations}, we ablate \ours{}'s loss and weight update mechanisms and their hyperparameters. Slowing the adaptive weight growth via $\gamma$ or $D_{\max}$ (recall Eq.~\ref{eq:pal_weight_update} for definitions) can improve optimality, while an excessive growth rate $\gamma$ can collapse optimality unless mitigated by low $D_{\max}$. We find that outcomes are insensitive to the threshold $\tau_d$ across several orders of magnitude. If $\tau_d$ is set to a sufficiently small numerical value, further tightening has no effect. Using $f(\tilde y^{(1)})$ in the training loss in place of $f(\hat y)$ degrades both optimality and feasibility. We hypothesize on a potential cause in detail in Appendix \ref{app:loss_placement}. Disabling either $\bar w_c$ or $\bar w_d$ preserves performance, suggesting redundancy of these loss terms. Disabling both collapses optimality and degrades pre-projection feasibility, yet the projector alone still recovers ${\sim}99\%$ feasibility. This validates our load-share argument: the projector achieves feasibility in isolation, but ideal results require non-trivial constraint pressure on the backbone.

\paragraph{Alternative Repair Steps.} In Appendix~\ref{app:repair_variants} we compare \ours{} with \acs{SQP} and \acl{IP} repair in place of \ac{LM}. Both roughly match \ac{LM} across the synthetic benchmarks, showing that \ours{} is agnostic to the type of projection step, as long as it is robust on ill-conditioned Jacobians. (Newton projection, for example, which is a special case of \ac{LM} without damping, is not. We discuss damping hyperparameters of the \ac{LM} step in Appendix \ref{app:ablations}).  %

\paragraph{Repair Steps in Inference.} In Appendix~\ref{app:precision} we vary the number of repair steps $N$ at test time and show that single-step training does not limit the precision of constraint satisfaction \ours{} reaches at inference.

\section{Conclusion}

\label{sec:conclusion}

\paragraph{Contributions and Outcome.} 

We introduced \ours{}, a scheme where single-step projection metrics drive adaptive constraint weights. We show that with this method, fully unrolled repair is not necessary to achieve good feasibility or optimality, and a single step is in fact sufficient. In fact, \ours{} is the \emph{only} method that achieves best-in-class results simultaneously on synthetic stress-testing (Tab.~\ref{tab:methods-sweep-synthetic}), engineering applications (Tab.~\ref{fig:engineering_results}), and under extremely high curvature (Fig.~\ref{fig:curvature_sweep}). 

In \S\ref{sec:engineering_results} we show that using only a single detached repair step enables tractable training when constraints are expensive to evaluate, like with high-fidelity physics surrogates. Current full-unroll methods were not able to train on these problems due to memory. Using only a single detached step also results in fewer constraint evaluator calls (\S\ref{sec:synthetic_results}), and \ours{} trains 2.5$\times$ faster  than DC3 on its flagship benchmark (while still providing better feasibility and optimality). 

Lastly, \ours{}'s adaptive weight scheme shifts feasibility towards the neural backbone (Fig.~\ref{fig:residual_intuition}) which soft-loss unrolled baselines do not do. Under extremely non-linear constraints, this manifests in \ours{} retaining near-full feasibility while unrolled soft-loss baselines degrade (\S\ref{sec:curvature}). 

\paragraph{Limitations.}
Several limitations remain: (1) Like all other projection-based methods, we require the constraint evaluation to be differentiable. (2) In all repair-based methods, feasibility is defined w.r.t.\ the evaluator $\Phi$. If $\Phi$ is inaccurate, constraints will be satisfied according to an incorrect reference. (3) Assembly of the Jacobian requires $K$ backward passes through the evaluator $\Phi$ for all  projection-based methods. \ours{} alleviates this issue somewhat as it only requires a single projection step, but problems where the number of constraints $K$ is extremely large are still compute-intensive. And (4), \ours{} introduces several hyperparameters ($\gamma$, $\tau_d$, $\tau_c$, $D_{\max}$, $\delta$, $\lambda_{\min}$), though in practice $\tau_c$ follows immediately from problem-specific tolerances, $D_{\max}$ can be left at a conservative default, and $\delta$ and $\lambda_{\min}$ have negligible impact.

\paragraph{Future Work.}
A first natural extension is to study how \ours{} performs on tasks beyond amortized optimization, where the objective $f$ is a data-driven loss (e.g., MSE). Second, efficiency improvements or auto-calibration of the remaining hyperparameters could make single-step repair with \ours{} even more easily usable in scientific and engineering applications. Examples include estimating $\bar w_d$ and $\bar w_c$ from sparse projection only every few steps, or auto-calibrating $\gamma$ and $D_{\max}$ from early-trajectory metrics or $\tau_d$ from the relative error between $f(\hat y)$ and $f(\tilde y^{(1)})$). Third, since the repair step is never backpropagated through, \ours{} does not require a differentiable projector. This opens the door to bespoke, non-differentiable repair methods that are specialized to niche applications beyond the projection methods explored here.

\subsubsection*{Acknowledgments}
We would like to thank Giacomo Lastrucci, one of the authors of ENFORCE~\citep{lastrucci2025enforce}, for the inspiration and productive discussion, and Gerhard Br\"aunlich for support with the differentiable geometry tooling underlying our engineering benchmarks.

\bibliography{references}

\begin{thebibliography}{36}
\providecommand{\natexlab}[1]{#1}
\providecommand{\url}[1]{\texttt{#1}}
\expandafter\ifx\csname urlstyle\endcsname\relax
  \providecommand{\doi}[1]{doi: #1}\else
  \providecommand{\doi}{doi: \begingroup \urlstyle{rm}\Url}\fi

\bibitem[Agrawal et~al.(2019)Agrawal, Amos, Barratt, Boyd, Diamond, and
  Kolter]{agrawal2019differentiable}
Akshay Agrawal, Brandon Amos, Shane Barratt, Stephen Boyd, Steven Diamond, and
  J.~Zico Kolter.
\newblock Differentiable convex optimization layers.
\newblock In \emph{Advances in Neural Information Processing Systems 32
  (NeurIPS)}, 2019.

\bibitem[{AI4OPT}(2024)]{ml4opf}
{AI4OPT}.
\newblock {ML4OPF}: Machine learning for optimal power flow.
\newblock \url{https://github.com/AI4OPT/ML4OPF}, 2024.

\bibitem[Amos \& Kolter(2017)Amos and Kolter]{amos2017optnet}
Brandon Amos and J.~Zico Kolter.
\newblock {OptNet}: Differentiable optimization as a layer in neural networks.
\newblock In \emph{Proceedings of the 34th International Conference on Machine
  Learning (ICML)}, pp.\  136--145, 2017.

\bibitem[Basir \& Senocak(2023)Basir and Senocak]{basir2023adaptive}
Shamsulhaq Basir and Inanc Senocak.
\newblock An adaptive augmented {L}agrangian method for training physics and
  equality constrained artificial neural networks.
\newblock \emph{arXiv preprint arXiv:2306.04904}, 2023.

\bibitem[Beucler et~al.(2021)Beucler, Pritchard, Rasp, Ott, Baldi, and
  Gentine]{beucler2021enforcing}
Tom Beucler, Michael Pritchard, Stephan Rasp, Jordan Ott, Pierre Baldi, and
  Pierre Gentine.
\newblock Enforcing analytic constraints in neural networks emulating physical
  systems.
\newblock \emph{Physical Review Letters}, 126\penalty0 (9):\penalty0 098302,
  2021.
\newblock \doi{10.1103/PhysRevLett.126.098302}.

\bibitem[Bouhlel et~al.(2020)Bouhlel, He, and Martins]{bouhlel2020scalable}
Mohamed~Amine Bouhlel, Sicheng He, and Joaquim R. R.~A. Martins.
\newblock Scalable gradient-enhanced artificial neural networks for airfoil
  shape design in the subsonic and transonic regimes.
\newblock \emph{Structural and Multidisciplinary Optimization}, 61\penalty0
  (4):\penalty0 1363--1376, 2020.
\newblock \doi{10.1007/s00158-020-02488-5}.

\bibitem[Brunke et~al.(2022)Brunke, Greeff, Hall, Yuan, Zhou, Panerati, and
  Schoellig]{brunke2022safe}
Lukas Brunke, Melissa Greeff, Adam~W. Hall, Zhaocong Yuan, Siqi Zhou, Jacopo
  Panerati, and Angela~P. Schoellig.
\newblock Safe learning in robotics: From learning-based control to safe
  reinforcement learning.
\newblock \emph{Annual Review of Control, Robotics, and Autonomous Systems},
  5:\penalty0 411--444, 2022.
\newblock \doi{10.1146/annurev-control-042920-020211}.

\bibitem[Chu et~al.(2026)Chu, Boukas, and Udell]{chu2026snarenet}
Ya-Chi Chu, Alkiviades Boukas, and Madeleine Udell.
\newblock {SnareNet}: Flexible repair layers for neural networks with hard
  constraints.
\newblock \emph{arXiv preprint arXiv:2602.09317}, 2026.

\bibitem[Donti et~al.(2021)Donti, Rolnick, and Kolter]{donti2021dc3}
Priya~L. Donti, David Rolnick, and J.~Zico Kolter.
\newblock {DC3}: A learning method for optimization with hard constraints.
\newblock In \emph{International Conference on Learning Representations
  (ICLR)}, 2021.

\bibitem[Gill et~al.(2005)Gill, Murray, and Saunders]{gill2005snopt}
Philip~E. Gill, Walter Murray, and Michael~A. Saunders.
\newblock {SNOPT}: An {SQP} algorithm for large-scale constrained optimization.
\newblock \emph{SIAM Review}, 47\penalty0 (1):\penalty0 99--131, 2005.
\newblock \doi{10.1137/S0036144504446096}.

\bibitem[Grontas et~al.(2026)Grontas, Terpin, Balta, D'Andrea, and
  Lygeros]{pinet}
Panagiotis~D. Grontas, Antonio Terpin, Efe~C. Balta, Raffaello D'Andrea, and
  John Lygeros.
\newblock {$\Pi$net}: Optimizing hard-constrained neural networks with
  orthogonal projection layers.
\newblock In \emph{The Fourteenth International Conference on Learning
  Representations (ICLR)}, 2026.

\bibitem[HaCohen et~al.(2024)HaCohen, Chiprut, Brazowski, Shalem, Moshe,
  Richardson, Levin, Shiran, Zabari, Gordon, Panet, Weissbuch, Kulikov,
  Bitterman, Melumian, and Bibi]{hacohen2024ltxvideo}
Yoav HaCohen, Nisan Chiprut, Benny Brazowski, Daniel Shalem, Dudu Moshe, Eitan
  Richardson, Eran Levin, Guy Shiran, Nir Zabari, Ori Gordon, Poriya Panet,
  Sapir Weissbuch, Victor Kulikov, Yaki Bitterman, Zeev Melumian, and Ofir
  Bibi.
\newblock {LTX-Video}: Realtime video latent diffusion.
\newblock \emph{arXiv preprint arXiv:2501.00103}, 2024.

\bibitem[Hoyer et~al.(2019)Hoyer, Sohl-Dickstein, and
  Greydanus]{hoyer2019neural}
Stephan Hoyer, Jascha Sohl-Dickstein, and Sam Greydanus.
\newblock Neural reparameterization improves structural optimization.
\newblock \emph{arXiv preprint arXiv:1909.04240}, 2019.
\newblock NeurIPS 2019 Workshop on Solving Inverse Problems with Deep Networks.

\bibitem[Kingma \& Welling(2014)Kingma and Welling]{wellingvae}
Diederik~P. Kingma and Max Welling.
\newblock Auto-encoding variational bayes.
\newblock In \emph{International Conference on Learning Representations
  (ICLR)}, 2014.

\bibitem[Krishnapriyan et~al.(2021)Krishnapriyan, Gholami, Zhe, Kirby, and
  Mahoney]{krishnapriyan2021characterizing}
Aditi Krishnapriyan, Amir Gholami, Shandian Zhe, Robert Kirby, and Michael~W.
  Mahoney.
\newblock Characterizing possible failure modes in physics-informed neural
  networks.
\newblock In \emph{Advances in Neural Information Processing Systems},
  volume~34, pp.\  26548--26560, 2021.

\bibitem[Lastrucci \& Schweidtmann(2025)Lastrucci and
  Schweidtmann]{lastrucci2025enforce}
Giacomo Lastrucci and Artur~M. Schweidtmann.
\newblock {ENFORCE}: Nonlinear constrained learning with adaptive-depth neural
  projection.
\newblock \emph{arXiv preprint arXiv:2502.06774}, 2025.

\bibitem[Lincoln(2011)]{pypower}
Richard Lincoln.
\newblock {PYPOWER}: Port of {MATPOWER} to {Python}.
\newblock \url{https://github.com/rwl/PYPOWER}, 2011.

\bibitem[Liu \& Nocedal(1989)Liu and Nocedal]{liu1989lbfgs}
Dong~C. Liu and Jorge Nocedal.
\newblock On the limited memory {BFGS} method for large scale optimization.
\newblock \emph{Mathematical Programming}, 45\penalty0 (1--3):\penalty0
  503--528, 1989.
\newblock \doi{10.1007/BF01589116}.

\bibitem[Mallappa et~al.(2024)Mallappa, Mostafa, Galkin, Phielipp, and
  Majumdar]{floorset2024}
Uday Mallappa, Hesham Mostafa, Mikhail Galkin, Mariano Phielipp, and Somdeb
  Majumdar.
\newblock {FloorSet} -- a {VLSI} floorplanning dataset with design constraints
  of real-world {SoCs}.
\newblock In \emph{Proceedings of the 43rd IEEE/ACM International Conference on
  Computer-Aided Design (ICCAD)}, pp.\  1--9, 2024.
\newblock \doi{10.1145/3676536.3676814}.

\bibitem[McDonald \& Gloudemans(2022)McDonald and
  Gloudemans]{mcdonald2022openvsp}
Robert~A. McDonald and J.~R. Gloudemans.
\newblock Open vehicle sketch pad: An open source parametric geometry and
  analysis tool for conceptual aircraft design.
\newblock In \emph{AIAA SciTech 2022 Forum}, 2022.
\newblock \doi{10.2514/6.2022-0004}.

\bibitem[Mildenhall et~al.(2020)Mildenhall, Srinivasan, Tancik, Barron,
  Ramamoorthi, and Ng]{mildenhall2020nerf}
Ben Mildenhall, Pratul~P. Srinivasan, Matthew Tancik, Jonathan~T. Barron, Ravi
  Ramamoorthi, and Ren Ng.
\newblock {NeRF}: Representing scenes as neural radiance fields for view
  synthesis.
\newblock In \emph{European Conference on Computer Vision (ECCV)}, 2020.

\bibitem[Min \& Azizan(2024)Min and Azizan]{min2024hardnet}
Youngjae Min and Navid Azizan.
\newblock {HardNet}: Hard-constrained neural networks with universal
  approximation guarantees.
\newblock \emph{arXiv preprint arXiv:2410.10807}, 2024.

\bibitem[Nguyen \& Donti(2025)Nguyen and Donti]{nguyen2025fsnet}
Hoang~T. Nguyen and Priya~L. Donti.
\newblock {FSNet}: Feasibility-seeking neural network for constrained
  optimization with guarantees.
\newblock In \emph{Advances in Neural Information Processing Systems
  (NeurIPS)}, 2025.

\bibitem[Nocedal \& Wright(2006)Nocedal and Wright]{nocedal2006numerical}
Jorge Nocedal and Stephen~J. Wright.
\newblock \emph{Numerical Optimization}.
\newblock Springer Series in Operations Research and Financial Engineering.
  Springer, New York, NY, 2 edition, 2006.
\newblock ISBN 978-0-387-30303-1.

\bibitem[Pan et~al.(2023)Pan, Chen, Zhao, and Low]{pan2022deepopf}
Xiang Pan, Minghua Chen, Tianyu Zhao, and Steven~H. Low.
\newblock {DeepOPF}: A feasibility-optimized deep neural network approach for
  {AC} optimal power flow problems.
\newblock \emph{IEEE Systems Journal}, 17\penalty0 (1):\penalty0 673--683,
  2023.
\newblock \doi{10.1109/JSYST.2022.3201041}.

\bibitem[Perini et~al.(2026)Perini, Bischof, Arar, Duran, Kraus, Mishra, and
  Bickel]{perini2026windinet}
Janne Perini, Rafael Bischof, Moab Arar, Ay\c{c}a Duran, Michael~A. Kraus,
  Siddhartha Mishra, and Bernd Bickel.
\newblock Pretrained video models as differentiable physics simulators for
  urban wind flows.
\newblock \emph{arXiv preprint arXiv:2603.21210}, 2026.

\bibitem[Platt \& Barr(1988)Platt and Barr]{platt1988constrained}
John~C. Platt and Alan~H. Barr.
\newblock Constrained differential optimization.
\newblock In \emph{Advances in Neural Information Processing Systems
  (NeurIPS)}, 1988.

\bibitem[Sullivan \& Kaszynski(2019)Sullivan and
  Kaszynski]{sullivan2019pyvista}
C.~Bane Sullivan and Alexander~A. Kaszynski.
\newblock {PyVista}: {3D} plotting and mesh analysis through a streamlined
  interface for the {Visualization Toolkit (VTK)}.
\newblock \emph{Journal of Open Source Software}, 4\penalty0 (37):\penalty0
  1450, 2019.
\newblock \doi{10.21105/joss.01450}.

\bibitem[Sung et~al.(2025)Sung, Spreizer, Elrefaie, Jones, and
  Ahmed]{sung2025blendednet}
Nicholas Sung, Steven Spreizer, Mohamed Elrefaie, Matthew~C. Jones, and Faez
  Ahmed.
\newblock {BlendedNet++}: A large-scale blended wing body aerodynamics dataset
  and benchmark.
\newblock \emph{arXiv preprint arXiv:2512.03280}, 2025.

\bibitem[Tancik et~al.(2020)Tancik, Srinivasan, Mildenhall, Fridovich-Keil,
  Raghavan, Singhal, Ramamoorthi, Barron, and Ng]{tancik2020fourier}
Matthew Tancik, Pratul~P. Srinivasan, Ben Mildenhall, Sara Fridovich-Keil,
  Nithin Raghavan, Utkarsh Singhal, Ravi Ramamoorthi, Jonathan~T. Barron, and
  Ren Ng.
\newblock Fourier features let networks learn high frequency functions in low
  dimensional domains.
\newblock In \emph{Advances in Neural Information Processing Systems
  (NeurIPS)}, 2020.

\bibitem[Thurner et~al.(2018)Thurner, Scheidler, Sch{\"a}fer, Menke, Dollichon,
  Meier, Meinecke, and Braun]{pandapower.2018}
L.~Thurner, A.~Scheidler, F.~Sch{\"a}fer, J.~Menke, J.~Dollichon, F.~Meier,
  S.~Meinecke, and M.~Braun.
\newblock pandapower — an open-source python tool for convenient modeling,
  analysis, and optimization of electric power systems.
\newblock \emph{IEEE Transactions on Power Systems}, 33\penalty0 (6):\penalty0
  6510--6521, 2018.
\newblock \doi{10.1109/TPWRS.2018.2829021}.

\bibitem[W{\"a}chter \& Biegler(2006)W{\"a}chter and Biegler]{wachter2006ipopt}
Andreas W{\"a}chter and Lorenz~T. Biegler.
\newblock On the implementation of an interior-point filter line-search
  algorithm for large-scale nonlinear programming.
\newblock \emph{Mathematical Programming}, 106\penalty0 (1):\penalty0 25--57,
  2006.
\newblock \doi{10.1007/s10107-004-0559-y}.

\bibitem[Watt-Meyer et~al.(2023)Watt-Meyer, Dresdner, McGibbon, Clark, Henn,
  Peters, Bretherton, Duncan, Brenowitz, Kashinath, Pritchard, and
  Bonev]{wattmeyer2023ace}
Oliver Watt-Meyer, Gideon Dresdner, Jeremy McGibbon, Spencer~K. Clark, Brian
  Henn, Matthew~E. Peters, Christopher~S. Bretherton, James Duncan, Noah~D.
  Brenowitz, Karthik Kashinath, Michael~S. Pritchard, and Boris Bonev.
\newblock {ACE}: A fast, skillful learned global atmospheric model for climate
  prediction.
\newblock \emph{arXiv preprint arXiv:2310.02074}, 2023.

\bibitem[Yamashita \& Fukushima(2001)Yamashita and
  Fukushima]{yamashita2001rate}
Nobuo Yamashita and Masao Fukushima.
\newblock On the rate of convergence of the {Levenberg-Marquardt} method.
\newblock In G.~Alefeld and X.~Chen (eds.), \emph{Topics in Numerical
  Analysis}, volume~15 of \emph{Computing Supplementa}, pp.\  239--249.
  Springer, Vienna, 2001.
\newblock \doi{10.1007/978-3-7091-6217-0_18}.

\bibitem[Yang et~al.(2022)Yang, Balaprakash, and Leyffer]{yang2022nnsurrogates}
Dominic Yang, Prasanna Balaprakash, and Sven Leyffer.
\newblock Modeling design and control problems involving neural network
  surrogates.
\newblock \emph{Computational Optimization and Applications}, 83:\penalty0
  759--800, 2022.
\newblock \doi{10.1007/s10589-022-00404-9}.

\bibitem[Zimmerman et~al.(2011)Zimmerman, Murillo-S{\'a}nchez, and
  Thomas]{matpower2011}
Ray~Daniel Zimmerman, Carlos~Edmundo Murillo-S{\'a}nchez, and Robert~John
  Thomas.
\newblock {MATPOWER}: Steady-state operations, planning, and analysis tools for
  power systems research and education.
\newblock \emph{IEEE Transactions on Power Systems}, 26\penalty0 (1):\penalty0
  12--19, 2011.
\newblock \doi{10.1109/TPWRS.2010.2051168}.

\end{thebibliography}

\bibliographystyle{preprint}

\appendix

\section{Notation Summary}
\label{app:notation_summary}
See Table~\ref{tab:notation-summary}.

\begin{table}[!h]
\centering
\small
\caption{Notation summary.}
\begin{tabular}{@{}cl@{}}
\toprule
Symbol & Meaning \\
\midrule
$y,\; \hat y,\; \tilde y^{(i)}$ & design variable, raw neural backbone prediction, $i$-th projection iterate \\
$z$ & input condition or latent embedding \\
$f,\; h,\; g$ & objective, equality constraints, inequality constraints \\
$c_k,\; c_\mathcal{A}$ & per-constraint residual, active residual vector \\
$\mathcal{A}$ & active index set \\
$J_\mathcal{A}$ & active constraint Jacobian $\partial c_\mathcal{A}/\partial y$ \\
$\mathcal{M}_\mathcal{A}$ & local feasible manifold $\{y : c_\mathcal{A}(y)=0\}$ \\
$N$ & number of projection steps \\
$\Delta y^{(i)},\; R_i$ & projection step displacement, constraint violation residual at iterate $i$ \\
$\lambda_i,\; \delta$ & \ac{LM} damping at step $i$, damping hyperparameter \\
$\tau_c,\; \tau_d$ & residual threshold, displacement threshold \\
$\bar w_{c,k},\; \bar w_d$ & adaptive constraint and displacement penalty weights \\
$\gamma,\; D_{\max}$ & log-gap rate, per-step growth ceiling (decades) \\
\bottomrule
\end{tabular}
\label{tab:notation-summary}
\end{table}

\section{Ablation Study}
\label{app:ablations}

In this appendix, we report full ablations for \ours{}. These ablations were carried out on the six synthetic benchmarks \texttt{S1}--\texttt{S6}, over $n=10$ seeds each. We report worst-bench (i.e., maximum) objective across the six synthetic benchmarks and minimum feasibility across the six benchmarks. Additionally, we report the pre-projection residual $\|c(\hat y)\|_\infty$ averaged across \texttt{S1}--\texttt{S6}. This is a measure of how much of the constraint-satisfaction load the backbone $N_\theta$ carries. Table \ref{tab:ablation} shows the full ablation summary.

\begin{table}[t]
\centering
\footnotesize
\setlength{\tabcolsep}{3pt}
\caption{Ablation of \ours{} hyperparameters on synthetic benchmarks ($n=10$ seeds). Blank cells inherit the \texttt{Base} value. Bold $=$ best per column. The $f$ column lists the argument used inside the objective term of the training loss ($\hat y$ vs $\tilde y^{(1)}$). \texttt{Dec.} marks whether the multiplier rule's decay half is active: $\checkmark$ = bidirectional (default), $\xmark$ = monotone-only growth.}
\label{tab:ablation}
\resizebox{\textwidth}{!}{%
\begin{tabular}{c@{\hspace{0.9em}}ccccccccc@{\quad}ccc}
\toprule
\texttt{ID}& $\delta$ & $\lambda_{\min}^{\text{proj}}$ & $\gamma$ & $D_{\text{max}}$ & $\tau_c$ & $\tau_d$ & LR & $f$ & Dec. & Obj.\ (max)\,$\downarrow$ & Feas.\ (\%)\,$\uparrow$ & $\|c(\hat y)\|_\infty\,\downarrow$  \\
\midrule
\texttt{Base}&$10^{-3}$ & $10^{-6}$ & $10^{-2}$ & 1 & $10^{-4}$ & $10^{-4}$ & $10^{-4}$ & $\hat y$ & \checkmark & 0.08 $\pm$ 0.01 & 100 $\pm$ 0\% & 0.06 $\pm$ 0.02  \\
\midrule
\texttt{A}&0 &  &  &  &  &  &  &  &  & 0.07 $\pm$ 0.01 & 100 $\pm$ 0\% & 0.06 $\pm$ 0.02  \\
\texttt{B}&$10^{-4}$ &  &  &  &  &  &  &  &  & 0.07 $\pm$ 0.01 & 100 $\pm$ 0\% & \textbf{0.05 $\pm$ 0.01}  \\
\texttt{C}&$10^{-2}$ &  &  &  &  &  &  &  &  & 0.10 $\pm$ 0.02 & 100 $\pm$ 0\% & 0.06 $\pm$ 0.02  \\
\texttt{D}&0 & 0 &  &  &  &  &  &  &  & 0.11 $\pm$ 0.09 & 100 $\pm$ 0\% & 0.06 $\pm$ 0.02  \\
\texttt{E}& &  & $10^{-3}$ &  &  &  &  &  &  & 0.06 $\pm$ 0.03 & 100 $\pm$ 0\% & 0.09 $\pm$ 0.06  \\
\texttt{F}& &  & $10^{-1}$ &  &  &  &  &  &  & 0.72 $\pm$ 0.04 & 100 $\pm$ 0\% & 0.06 $\pm$ 0.02  \\
\texttt{G}& &  & $10^{-1}$ & 0.1 &  &  &  &  &  & 0.08 $\pm$ 0.01 & 100 $\pm$ 0\% & 0.06 $\pm$ 0.02  \\
\texttt{H}& &  &  & 0.2 &  &  &  &  &  & 0.04 $\pm$ 0.03 & 100 $\pm$ 0\% & 0.06 $\pm$ 0.04  \\
\texttt{J}& &  &  & 5 &  &  &  &  &  & 0.23 $\pm$ 0.02 & 100 $\pm$ 0\% & 0.06 $\pm$ 0.02  \\
\texttt{K}& &  &  &  &  & $10^{-2}$ &  &  &  & 0.08 $\pm$ 0.01 & 100 $\pm$ 0\% & 0.07 $\pm$ 0.02  \\
\texttt{L}& &  &  &  &  &  & $10^{-3}$ &  &  & 0.13 $\pm$ 0.02 & 100 $\pm$ 0\% & 0.22 $\pm$ 0.06  \\
\texttt{M}& &  &  &  &  & $10^{-4}$ &  & $\tilde y^{(1)}$ &  & 0.84 $\pm$ 0.79 & 99 $\pm$ 1\% & 0.40 $\pm$ 0.01  \\
\texttt{N}& &  &  &  &  & $10^{-1}$ &  & $\tilde y^{(1)}$ &  & 0.84 $\pm$ 0.79 & 99 $\pm$ 1\% & 0.61 $\pm$ 0.06  \\
\texttt{O}& &  &  &  &  & $10^{-1}$ &  & $\hat y$ &  & \textbf{0.03 $\pm$ 0.04} & 100 $\pm$ 0\% & 0.14 $\pm$ 0.04  \\
\texttt{P}& &  &  &  & $10^{10}$ &  &  &  &  & 0.08 $\pm$ 0.03 & 100 $\pm$ 0\% & 0.07 $\pm$ 0.02  \\
\texttt{Q}& &  &  &  &  & $10^{10}$ &  &  &  & 0.06 $\pm$ 0.01 & 100 $\pm$ 0\% & 0.27 $\pm$ 0.02  \\
\texttt{R}& &  &  &  & $10^{10}$ & $10^{10}$ &  &  &  & 0.93 $\pm$ 0.52 & 99 $\pm$ 2\% & 1.19 $\pm$ 0.11  \\
\texttt{S}& &  &  &  &  &  &  &  & \xmark & 0.09 $\pm$ 0.01 & 100 $\pm$ 0\% & \textbf{0.05 $\pm$ 0.01}  \\
\bottomrule
\end{tabular}
}%
\end{table}

\paragraph{Projector Damping (\texttt{A}--\texttt{D}).}
Rows \texttt{A}--\texttt{D} show that performance is insensitive to the residual-scaling coefficient $\delta$ on these tasks. Rows \texttt{A}, \texttt{B}, \texttt{C} sweep $\delta$ over three orders of magnitude and match \texttt{Base} on objective and feasibility. Row \texttt{A} ($\delta=0$) retains the numerical floor $\lambda_{\min}^{\text{proj}}$. Row \texttt{D} removes both residual-scaled damping and the floor, but on \texttt{S2}--\texttt{S6} the masked $BB^\top$ is singular, the Cholesky solve fails, and the eigenvalue fallback of the projector applies its own floor of $10^{-4}$. Only \texttt{S1} runs undamped, and there two of ten seeds degrade in objective, which accounts for the larger spread of row \texttt{D}. A fully undamped step is exposed on ill-conditioned $BB^\top$, which is the failure mode of the undamped completion of DC3 in Table~\ref{tab:methods-sweep-synthetic}.

\paragraph{Multiplier-Update Timescale (\texttt{E}--\texttt{J}).}
The growth rate $\gamma$ and the per-step cap $D_{\max}$ jointly control how aggressively the constraint weights $\bar w_{c,k}$ react to residual overshoot. Lowering $\gamma$ marginally improves optimality (\texttt{E}), since slower weight growth leaves more early capacity for the objective. Raising $\gamma$ collapses optimality (\texttt{F}) as the penalty term overwhelms $f$. Pairing a high $\gamma$ with a low $D_{\max}$ (\texttt{G}) recovers \texttt{Base} performance, confirming that $\gamma$ and $D_{\max}$ act together rather than independently. Tightening $D_{\max}$ alone (\texttt{H}) halves \texttt{Base}'s worst-bench objective, while loosening it (\texttt{J}) is several times worse but not catastrophic.

\paragraph{Tolerances and Learning Rate (\texttt{K}--\texttt{L}).}
The displacement threshold $\tau_d$ can be loosened by two orders of magnitude without effect (\texttt{K}). It only needs to be ``small enough''. The learning-rate sensitivity in \texttt{L} is generic to neural training and not specific to \ours{}.

\paragraph{Pre-Projection Objective (\texttt{M}--\texttt{O}).}
Replacing $f(\hat y)$ with $f(\tilde y^{(1)})$ in the training loss degrades both optimality and feasibility across two values of $\tau_d$ (\texttt{M}, \texttt{N}), while a matched control with the same $\tau_d$ and $f(\hat y)$ (\texttt{O}) recovers near-baseline. We hypothesize this is related to how gradients flow back through the projection step, which we detail in Appendix~\ref{app:loss_placement}. 

\paragraph{Load-Share (\texttt{P}--\texttt{R}).}
Disabling either $\bar w_c$ or $\bar w_d$ alone (by setting its threshold to $10^{10}$) preserves objective and feasibility on this benchmark (\texttt{P}, \texttt{Q}), with the only visible effect being a higher pre-projection $\|c(\hat y)\|_\infty$ in \texttt{Q} where $\bar w_d$ is off. The two terms in Eq.~\ref{eq:pal_weight_update} therefore act as redundant safety nets here, although both may be needed for stiffer constraint systems. Disabling both (\texttt{R}) lifts all constraint pressure from the backbone: optimality collapses and pre-projection $\|c(\hat y)\|_\infty$ blows up, yet the projector still recovers ${\sim}99\%$ feasibility on its own. This is an empirical demonstration of our load-share argument: the projector is often capable enough to handle feasibility in isolation, but a more preferable outcome is for the neural backbone to carry significant constraint-satisfaction burden, and the projector (or, more generally: the repair step) to be utilized as clean-up only.

Together, the ablation isolates the design choices behind \ours{}'s loss formulation: the log-gap weight update with a growth cap for the multipliers (\texttt{E}--\texttt{J}), the pre-projection objective $f(\hat y)$ (\texttt{M}--\texttt{O}), and load-share, where backbone pressure remains necessary for optimality even when the projector alone recovers feasibility (\texttt{P}--\texttt{R}).

\section{{Constraint Curvature Ablation}}
\label{app:curvature}

Here, we compare how \ours{} and the baselines perform under extreme constraint curvature. As a benchmark, we construct a problem with a single equality constraint, whose curvature $\kappa$ ranges from $0$ to $10^{6}$. The problem dimensionality is $\dim(y)=8$. Equation~\ref{eq:curv_benchmark} shows the constraint $c(y)$ (variable-frequency sine on the hyperplane $a^\top y = 0$) with a fixed orthonormal pair $a \perp b$, drawn once and shared across all $\kappa$ and all seeds.

\begin{equation}
c(y) \;=\; a^\top y \;-\; \sin\!\big(\omega\, b^\top y\big), \qquad
\kappa \;=\; \omega^2, \qquad
f_x(y) \;=\; \big\lVert\, T^{-1}(y) - T^{-1}(x) \,\big\rVert_2^2 .
\label{eq:curv_benchmark}
\end{equation}

The parameter $\kappa$ is the curvature stemming from the high-frequency sine component of the constraint. The objective is formulated such that it has a unique optimum (non-multimodal) and a closed-form $f^\star$ at every $\kappa$, using the shear

\begin{equation}
T(v) = v + a \sin\!\big(\omega\, b^\top v\big), \qquad
T^{-1}(v) = v - a \sin\!\big(\omega\, b^\top v\big).
\label{eq:curv_warp}
\end{equation}

All methods use bayesian-optimization tuned hyperparameters (see Appendix~\ref{app:hp_tuning}). Plain \ac{ALM} is 0\% feasible at all $\kappa$, so we omit it. We use 10 seeds and 512 evaluations per seed, and all runs are in double precision. We measure feasibility here as violations below $10^{-4}$.

Additionally, we report the ratio $\|\nabla_\theta L_c\|/\|\nabla_\theta L\|$, which is the recorded mean over the last five logged epochs of each training seed. This ratio is effectively the overall share of training pressure exerted by the constraint violation. 

Table~\ref{tab:curv_full} shows feasibility, optimality gap, and constraint training pressure for all methods at varying $\kappa$. At zero curvature (purely linear constraint), all methods reach 100\% feasibility with virtually no optimality gap. For \ours{} at $\kappa = 0$, the constraint pressure $\|\nabla_\theta L_c\|/\|\nabla_\theta L\|$ is equal to zero across $\tau_c \in \{10^{-4}, 10^{-2}, 1\}$. This is because the purely linear constraint at $\kappa = 0$ is repaired exactly with a single projection step, so there is no single-step remainder. 

As can be expected, the constraint training pressure in \ours{} increases with higher curvature $\kappa$ (due to larger post-step residuals $R_1$) and with lower residual targets $\tau_c$, reaching up to $\|\nabla_\theta L_c\|/\|\nabla_\theta L\| = 91 \%$ at $\kappa=10^4$ and $\tau_c=10^{-4}$. This is in contrast to the full-unroll baselines. Here, training pressure is weaker at high curvature because post-repair residuals are soft-penalized (or, in the case of FSNet, not at all) instead of being driven by adaptive weight growth as in \ours{}. One exception is ENFORCE (which, through the soft displacement penalty with constant $w_d$, exerts pressure especially at low $\kappa$ where displacement $\|\tilde y - \hat y\|$ is large). But also for ENFORCE, constraint pressure, feasibility and optimality degrade at higher curvatures.

Overall, this experiment demonstrates that single-step repair in training does \textbf{not} limit \ours{} to near-linear constraints. This is by design, as the single-step residual under high curvature triggers improved load sharing in \ours{}. On the contrary, thanks to the use of adaptive weight updates, \ours{} exhibits \textbf{better} performance than existing full-unroll baselines under high-curvature constraints.

All runs use double precision, since above $\kappa = 10^4$ the constraint Jacobian is large even at small constraint values, and in single precision the rounding error in $y$ alone produces violations above the feasibility threshold. DC3 does not run with its undamped Newton completion, which diverges on ill-conditioned Jacobians. We instead use a damped Newton completion with Yamashita-Fukushima damping.

\begin{table}[t]
\centering
\footnotesize
\setlength{\tabcolsep}{8pt}
\caption{Full results for high-curvature experiment. $\kappa$ from 0 to $10^{6}$, 10 seeds, all methods trained and evaluated in \texttt{float64}. $\|\nabla_\theta L_c\|/\|\nabla_\theta L\|$ is the share of constraint training pressure. Bold marks the best feasibility at each $\kappa$ across all methods. $\dagger$: gradient ratio carried over from the source ALM training run. Plain ALM without Bolt-On is omitted because it is 0\% feasible at every $\kappa$. DC3 uses a Newton completion with Yamashita-Fukushima damping, since the undamped completion diverges for every $\kappa > 0$.}
\label{tab:curv_full}
\begin{tabular}{lcccccc}
\toprule
 & \multicolumn{3}{c}{\ours{} ($\tau_c{=}10^{-4}$)} & \multicolumn{3}{c}{\ours{} ($\tau_c{=}10^{-2}$)} \\
\cmidrule(lr){2-4}\cmidrule(lr){5-7}
$\kappa$ & Feas.\ (\%) & Obj. & $\tfrac{\|\nabla L_c\|}{\|\nabla L\|}$ (\%) & Feas.\ (\%) & Obj. & $\tfrac{\|\nabla L_c\|}{\|\nabla L\|}$ (\%) \\
\midrule
$\kappa{=}0$ & \textbf{\boldmath 100$\pm$0} & 0.01$\pm$0.00 & 0.0 & \textbf{\boldmath 100$\pm$0} & 0.01$\pm$0.00 & 0.0 \\
$\kappa{=}1$ & \textbf{\boldmath 100$\pm$0} & 0.01$\pm$0.00 & 23 & \textbf{\boldmath 100$\pm$0} & 0.01$\pm$0.00 & 0.0 \\
$\kappa{=}10$ & \textbf{\boldmath 100$\pm$0} & 0.05$\pm$0.01 & 63 & \textbf{\boldmath 100$\pm$0} & 0.03$\pm$0.00 & 25 \\
$\kappa{=}10^{2}$ & \textbf{\boldmath 100$\pm$0} & 0.66$\pm$0.13 & 72 & 99.9$\pm$0.1 & 0.48$\pm$0.05 & 62 \\
$\kappa{=}10^{3}$ & \textbf{\boldmath 100$\pm$0} & 0.96$\pm$0.40 & 89 & 99.8$\pm$0.2 & 0.42$\pm$0.20 & 79 \\
$\kappa{=}10^{4}$ & \textbf{\boldmath 100.0$\pm$0.1} & 2.69$\pm$1.10 & 91 & 99.2$\pm$0.5 & 1.23$\pm$0.52 & 83 \\
$\kappa{=}10^{5}$ & \textbf{\boldmath 99.9$\pm$0.2} & 11.84$\pm$9.98 & 91 & 99.3$\pm$0.5 & 4.22$\pm$2.60 & 79 \\
$\kappa{=}10^{6}$ & 98.3$\pm$2.1 & 13.16$\pm$4.98 & 89 & \textbf{\boldmath 99.4$\pm$1.5} & 10.50$\pm$6.20 & 77 \\
\addlinespace[6pt]
 & \multicolumn{3}{c}{\ours{} ($\tau_c{=}1$)} & \multicolumn{3}{c}{ALM+Bolt-On} \\
\cmidrule(lr){2-4}\cmidrule(lr){5-7}
$\kappa$ & Feas.\ (\%) & Obj. & $\tfrac{\|\nabla L_c\|}{\|\nabla L\|}$ (\%) & Feas.\ (\%) & Obj. & $\tfrac{\|\nabla L_c\|}{\|\nabla L\|}$ (\%) \\
\midrule
$\kappa{=}0$ & \textbf{\boldmath 100$\pm$0} & 0.01$\pm$0.00 & 0.0 & \textbf{\boldmath 100$\pm$0} & 0.00$\pm$0.00 & 14$^{\dagger}$ \\
$\kappa{=}1$ & \textbf{\boldmath 100$\pm$0} & 0.01$\pm$0.00 & 0.0 & \textbf{\boldmath 100$\pm$0} & 0.19$\pm$0.01 & 13$^{\dagger}$ \\
$\kappa{=}10$ & \textbf{\boldmath 100$\pm$0} & 0.03$\pm$0.00 & 0.0 & \textbf{\boldmath 100$\pm$0} & 0.14$\pm$0.01 & 8.1$^{\dagger}$ \\
$\kappa{=}10^{2}$ & 97.9$\pm$1.1 & 0.34$\pm$0.03 & 0.0 & 98.7$\pm$0.7 & 0.25$\pm$0.08 & 1.9$^{\dagger}$ \\
$\kappa{=}10^{3}$ & 85.8$\pm$2.1 & 0.26$\pm$0.10 & 0.0 & 86.2$\pm$1.4 & 0.34$\pm$0.22 & 0.8$^{\dagger}$ \\
$\kappa{=}10^{4}$ & 79.3$\pm$3.1 & 0.78$\pm$0.43 & 0.0 & 78.0$\pm$2.7 & 0.56$\pm$0.12 & 0.9$^{\dagger}$ \\
$\kappa{=}10^{5}$ & 77.7$\pm$4.1 & 1.80$\pm$0.75 & 0.0 & 75.9$\pm$2.9 & 2.71$\pm$1.50 & 1.9$^{\dagger}$ \\
$\kappa{=}10^{6}$ & 87.4$\pm$8.0 & 6.12$\pm$4.17 & 0.0 & 76.2$\pm$4.7 & 6.97$\pm$2.78 & 6.7$^{\dagger}$ \\
\addlinespace[6pt]
 & \multicolumn{3}{c}{DC3} & \multicolumn{3}{c}{ENFORCE} \\
\cmidrule(lr){2-4}\cmidrule(lr){5-7}
$\kappa$ & Feas.\ (\%) & Obj. & $\tfrac{\|\nabla L_c\|}{\|\nabla L\|}$ (\%) & Feas.\ (\%) & Obj. & $\tfrac{\|\nabla L_c\|}{\|\nabla L\|}$ (\%) \\
\midrule
$\kappa{=}0$ & \textbf{\boldmath 100$\pm$0} & 0.76$\pm$0.03 & 114 & \textbf{\boldmath 100$\pm$0} & 0.00$\pm$0.00 & 234 \\
$\kappa{=}1$ & 76.4$\pm$1.8 & 0.86$\pm$0.03 & 143 & \textbf{\boldmath 100$\pm$0} & 0.05$\pm$0.00 & 341 \\
$\kappa{=}10$ & 75.6$\pm$2.5 & 0.99$\pm$0.05 & 125 & \textbf{\boldmath 100$\pm$0} & 0.04$\pm$0.00 & 170 \\
$\kappa{=}10^{2}$ & 89.7$\pm$2.0 & 1.26$\pm$0.06 & 88 & \textbf{\boldmath 100$\pm$0} & 0.32$\pm$0.02 & 65 \\
$\kappa{=}10^{3}$ & 90.4$\pm$2.0 & 1.20$\pm$0.22 & 92 & 94.9$\pm$8.3 & 37.90$\pm$58.84 & 93 \\
$\kappa{=}10^{4}$ & 87.5$\pm$3.4 & 1.47$\pm$0.70 & 94 & 93.7$\pm$19.9 & 233$\pm$610 & 53 \\
$\kappa{=}10^{5}$ & 91.5$\pm$5.7 & 3.12$\pm$0.46 & 94 & 85.1$\pm$25.3 & 393$\pm$982 & 46 \\
$\kappa{=}10^{6}$ & 92.2$\pm$8.4 & 5.33$\pm$0.74 & 93 & 92.2$\pm$24.3 & 10.95$\pm$7.14 & 61 \\
\addlinespace[6pt]
 & \multicolumn{3}{c}{FSNet} & \multicolumn{3}{c}{SnareNet} \\
\cmidrule(lr){2-4}\cmidrule(lr){5-7}
$\kappa$ & Feas.\ (\%) & Obj. & $\tfrac{\|\nabla L_c\|}{\|\nabla L\|}$ (\%) & Feas.\ (\%) & Obj. & $\tfrac{\|\nabla L_c\|}{\|\nabla L\|}$ (\%) \\
\midrule
$\kappa{=}0$ & \textbf{\boldmath 100$\pm$0} & 0.00$\pm$0.00 & 0.0 & \textbf{\boldmath 100$\pm$0} & 0.03$\pm$0.00 & 0.4 \\
$\kappa{=}1$ & \textbf{\boldmath 100$\pm$0} & 0.00$\pm$0.00 & 0.0 & \textbf{\boldmath 100$\pm$0} & 0.03$\pm$0.00 & 0.7 \\
$\kappa{=}10$ & 96.7$\pm$9.3 & 72.83$\pm$52.42 & 0.0 & \textbf{\boldmath 100$\pm$0} & 0.07$\pm$0.00 & 1.7 \\
$\kappa{=}10^{2}$ & 96.0$\pm$11.5 & 86.24$\pm$29.21 & 0.0 & \textbf{\boldmath 100$\pm$0} & 0.58$\pm$0.08 & 2.6 \\
$\kappa{=}10^{3}$ & 69.6$\pm$47.8 & 88.06$\pm$20.81 & 0.0 & 99.6$\pm$0.2 & 0.60$\pm$0.10 & 13 \\
$\kappa{=}10^{4}$ & 51.5$\pm$27.1 & 82.84$\pm$26.01 & 0.0 & 96.2$\pm$0.8 & 1.73$\pm$0.30 & 15 \\
$\kappa{=}10^{5}$ & 39.4$\pm$38.1 & 104$\pm$28 & 0.0 & 93.8$\pm$1.7 & 7.22$\pm$2.99 & 18 \\
$\kappa{=}10^{6}$ & 17.1$\pm$34.3 & 107$\pm$26 & 0.0 & 92.7$\pm$2.3 & 23.67$\pm$12.51 & 18 \\
\bottomrule
\end{tabular}%
\end{table}

\section{{Alternative Repair Steps}}
\label{app:repair_variants}

We swap the damped \ac{LM} projector of Equation~\ref{eq:pal_projection_loop} for two standard single-step repair operators, an \acf{SQP} step with elastic slacks~\citep{nocedal2006numerical,gill2005snopt} and an \acf{IP} step~\citep{nocedal2006numerical,wachter2006ipopt}. Each variant exposes one additional hyperparameter: the slack penalty for \ac{SQP} (we use $\rho = 10^{3}$) and the barrier weight for the \ac{IP} step (we use $\mu_0 = 0.01$). 

Both are trained, just as with our \ac{LM} repair step, with one detached repair step and iterate the same step at inference with a budget of 10 steps. The loss formulation and weight update rule remain unchanged from \ours{}. Beyond the above parameters $\rho$ and $\mu$, which were set to reasonable defaults but not tuned, we did not modify any other hyperparameters.

Table~\ref{tab:repair_variants} reports the objective at convergence for the three repair steps. Feasibility is $1.00 \pm 0.00$ for every method on every benchmark, with the single exception of the \ac{IP} step on \texttt{S3} ($0.86 \pm 0.05$). We speculate that its barrier steps stall on the near-parallel, ill-conditioned rows of \texttt{S3} and exhaust the ten-step inference repair budget.

The \ac{SQP} variant matches or slightly improves on \ac{LM} on five of the six benchmarks, tying on \texttt{S2}, \texttt{S3}, and \texttt{S5} and improving on \texttt{S1}, \texttt{S4}, and \texttt{S6}, and never underperforms \ac{LM}. This is consistent with \ac{SQP}'s coupled handling of the inequalities.

Overall, \ours{} accommodates all three repair operators without adaptation, which we interpret as \ours{} being agnostic to the specific repair step, provided the step is a linearized projection-type repair that is robust to ill-conditioned Jacobians.

\begin{table}[t]
\centering
\footnotesize
\caption{Objective at convergence ($f^\star = 0$) for the three repair step mechanisms on the synthetic benchmarks \texttt{S1}--\texttt{S6}. Shown is mean $\pm$ standard deviation across 10 seeds. All methods use default hyperparameters and consistently reach $100\%$ constraint satisfaction on \texttt{S1}--\texttt{S6}, except for IP on \texttt{S3} ($86 \pm 5\%$).}
\label{tab:repair_variants}
\begin{tabular}{@{}lccc@{}}
\toprule
 & \ours{}-\acs{LM} & \ours{}-\acs{SQP} & \ours{}-\acs{IP} \\
\midrule
\texttt{S1} & $0.02 \pm 0.02$ & $\textbf{\boldmath $0.01 \pm 0.01$}$ & $\textbf{\boldmath $0.01 \pm 0.01$}$ \\
\texttt{S2} & $\textbf{\boldmath $0.00 \pm 0.00$}$ & $\textbf{\boldmath $0.00 \pm 0.00$}$ & $0.02 \pm 0.00$ \\
\texttt{S3} & $\textbf{\boldmath $0.00 \pm 0.00$}$ & $\textbf{\boldmath $0.00 \pm 0.00$}$ & $\textbf{\boldmath $0.00 \pm 0.00$}$ \\
\texttt{S4} & $\textbf{\boldmath $0.01 \pm 0.00$}$ & $\textbf{\boldmath $0.01 \pm 0.00$}$ & $0.05 \pm 0.02$ \\
\texttt{S5} & $\textbf{\boldmath $0.00 \pm 0.00$}$ & $\textbf{\boldmath $0.00 \pm 0.00$}$ & $\textbf{\boldmath $0.00 \pm 0.00$}$ \\
\texttt{S6} & $0.08 \pm 0.01$ & $\textbf{\boldmath $0.07 \pm 0.01$}$ & $0.21 \pm 0.02$ \\
\bottomrule
\end{tabular}
\end{table}

\section{Additional Details on $f(\tilde y)$ vs. $f(\hat y)$ for Training Loss}
\label{app:loss_placement}

In this appendix, we hypothesize about the causes of stalled training when $f(\tilde y ^{(1)})$ is used in the \ours{} training loss. Intuitively, one would think that $f(\tilde y)$ is the logical choice, since it is the objective after projection. One would want to take the change in objective by the projection into account. This is also the motivation for using $f(\tilde y ^{(N)})$ in the baseline \ac{SOTA} methods we considered in \S~\ref{sec:related}.

However, we have found that using $f(\tilde y)$ directly in the loss can have disastrous consequences on training convergence of the objective when projection is enabled in early training. In short, this is because using $f(\tilde y)$ involves backpropagating its gradients through the projector. Here, we sketch out in rough terms that, because of how the projector is formulated for us (damped \ac{LM}), the gradients $\frac{\partial f(\tilde y)}{\partial \hat y}$ get \emph{projected} onto directions \emph{tangent to the constraint manifold} $\mathcal{M}$, which, in early training\footnote{On the other hand, later in training, this can be desirable, since one would want objective gradients to reach $N_\theta$ only in directions tangent to $\mathcal{M}$ at $\tilde y$.}, can be disastrous because the direction toward the global constrained optimum is, in general, not tangent to $\mathcal{M}$ at the repaired $\tilde y ^{(1)}$ when (as is usual in early training) the unrepaired prediction $\hat y$ is bad (very far from constraint optimum).

We first derive how the projector formulation leads to this effect, demonstrate it on our simple example described in Appendix~\ref{app:intuition_setup}, and then speculate why \ac{SOTA} methods employing similar projectors do not report on this effect (as far as we are aware).

\paragraph{The Differential of One \ac{LM} Step.} We recall the definition of the constraint Jacobian $J = \partial c / \partial y$ (size $K \times n$, with $K$ the number of active equality constraints and $n$ the dimension of $y$), $M = JJ^\top + \lambda I_K$, and $\lambda = \delta\,\|c(\hat y)\|^2$. The single \ac{LM} step is
\begin{equation}
\tilde y^{(1)}(\hat y) \;=\; \hat y \;-\; J^\top M^{-1} c.
\end{equation}
Applying the product rule gives
\begin{equation}
\frac{\partial \tilde y^{(1)}}{\partial \hat y}
\;=\;
I
\;-\; J^\top M^{-1} J
\;-\; \underbrace{(\partial J^\top / \partial \hat y)\, M^{-1} c}_{\mathcal{r}_1}
\;-\; \underbrace{J^\top\,(\partial M^{-1} / \partial \hat y)\, c}_{\mathcal{r}_2}.
\label{eq:lmstep_dy}
\end{equation}
where both remainders $\mathcal{r}_1$ and $\mathcal{r}_2$ vanish as $\|c\| \to 0$, so that:
\begin{equation}
\frac{\partial \tilde y^{(1)}}{\partial \hat y}
\;=\;
\mathbf{P}_\lambda \;+\; O(\|c\|),
\qquad
\mathbf{P}_\lambda
\;:=\;
I - J^\top (JJ^\top + \lambda I)^{-1} J.
\label{eq:Plambda}
\end{equation}
By the chain rule, $\partial f(\tilde y^{(1)})/\partial \hat y \approx \nabla f(\tilde y^{(1)})^\top \mathbf{P}_\lambda$, so the structure of $\mathbf{P}_\lambda$ determines the gradient-signal flow to the neural backbone $N_\theta$ when using $f(\tilde y^{(1)})$ in the training loss. Here, we just consider the case of using $f(\tilde y^{(1)})$, i.e., a single projection step. 

\paragraph{Graphical Intuition.} Take the simplest non-trivial case: one active equality constraint ($K=1$) with output dimension $n=2$. At $\tilde y^{(1)}$, let $u$ be the unit tangent to the feasible manifold and $v$ the unit normal (so $\nabla c \parallel v$). In the $(u, v)$ basis, $J = (0, \sigma)$ with $\sigma = \|\nabla c\|$, hence $JJ^\top = \sigma^2$ and
\[
\mathbf{P}_\lambda
\;=\;
\begin{pmatrix} 1 & 0 \\[2pt] 0 & \dfrac{\lambda}{\sigma^2+\lambda} \end{pmatrix}.
\]
Writing $\nabla f(\tilde y^{(1)}) = (f_u,\, f_v)^\top$ in the same basis,
\[
\mathbf{P}_\lambda^\top \nabla f(\tilde y^{(1)})
\;=\;
\begin{pmatrix} f_u \\[2pt] \dfrac{\lambda}{\sigma^2+\lambda}\, f_v \end{pmatrix}.
\]
The tangent component $f_u$ is preserved. The normal component $f_v$ is attenuated by $\lambda/(\sigma^2+\lambda)$, which vanishes for small damping terms $\lambda$. (We use $\lambda = \delta \|c\|^2$ with $\delta = 10^{-3}$, which results in $\lambda \ll 1$). Using an undamped projection ($\lambda=0$) step would entirely block $f_v$. 

In contrast, when we use the \emph{pre-projection} objective $f(\hat y)$, then $\partial f(\hat y)/\partial \hat y = \nabla f(\hat y)^\top$ is unperturbed since it is not routed through $\mathbf{P}_\lambda$. Objective and penalty can compete in the same subspace and the adaptive weight $\bar w_c$ arbitrates that competition. When the objective is placed at $\tilde y^{(1)}$, it is band-limited to the tangent subspace and no longer participates in the direction that controls feasibility pressure on the backbone.

\begin{figure}[!htbp]
\centering
\resizebox{0.72\linewidth}{!}{%
\begin{tikzpicture}[
    scale=1.2, font=\small,
    pt/.style={fill=black, draw=black, circle, inner sep=2pt},
    ghost/.style={fill=black!10, draw=black!45, circle, inner sep=1.4pt},
    proj/.style={dashed, black!55, thin},
    ghproj/.style={dashed, black!30, thin},
    pert/.style={<->, line width=0.7pt, >=Stealth, black!85},
    basisarrow/.style={->, line width=0.9pt, >=Latex, black},
]

\begin{scope}[shift={(0,0)}]
  \node[font=\bfseries] at (0, 1.85) {Tangent perturbation};

  \draw[black, thick, smooth, domain=-1.8:1.8]
    plot (\x, {-0.22*\x*\x - 1.2});
  \node[anchor=west] at (1.85, {-0.22*1.8*1.8 - 1.2}) {$\mathcal{M}$};

  \node[ghost] at (-0.85, 0.7) {};
  \node[ghost] at  (0.85, 0.7) {};
  \node[ghost] at (-0.62, {-0.22*0.62*0.62 - 1.2}) {};
  \node[ghost] at  (0.62, {-0.22*0.62*0.62 - 1.2}) {};
  \draw[ghproj] (-0.85, 0.7) -- (-0.62, {-0.22*0.62*0.62 - 1.2});
  \draw[ghproj]  (0.85, 0.7) --  (0.62, {-0.22*0.62*0.62 - 1.2});

  \node[pt, label={[xshift=3pt]right:$\hat y$}] (yhat) at (0, 0.7) {};
  \node[pt, label={[yshift=-2pt]below:$\tilde y^{(1)}$}] (ytil) at (0, -1.2) {};

  \draw[basisarrow] (0, -1.2) -- (0.85, -1.2) node[right=1pt, font=\footnotesize] {$u$};
  \draw[basisarrow] (0, -1.2) -- (0, -0.35) node[left=2pt, font=\footnotesize] {$v$};

  \draw[pert] (-0.85, 1.15) -- (0.85, 1.15) node[midway, above=2pt, font=\footnotesize] {$\delta u$};
\end{scope}

\begin{scope}[shift={(5.2,0)}]
  \node[font=\bfseries] at (0, 1.85) {Normal perturbation};

  \draw[black, thick, smooth, domain=-1.8:1.8]
    plot (\x, {-0.22*\x*\x - 1.2});
  \node[anchor=west] at (1.85, {-0.22*1.8*1.8 - 1.2}) {$\mathcal{M}$};

  \node[ghost] at (0, 0.0) {};
  \node[ghost] at (0, 1.4) {};
  \draw[ghproj] (0, 0.0) -- (0, -1.2);
  \draw[ghproj] (0, 1.4) -- (0, -1.2);

  \node[pt, label={[xshift=4pt]right:$\hat y$}] (yhat2) at (0, 0.7) {};
  \node[pt, label={[yshift=-2pt]below:$\tilde y^{(1)}$}] (ytil2) at (0, -1.2) {};

  \draw[basisarrow] (0, -1.2) -- (0.85, -1.2) node[right=1pt, font=\footnotesize] {$u$};
  \draw[basisarrow] (0, -1.2) -- (0, -0.35) node[left=2pt, font=\footnotesize] {$v$};

  \draw[pert] (0.85, 0.0) -- (0.85, 1.4) node[midway, right=2pt, font=\footnotesize] {$\delta v$};
\end{scope}
\end{tikzpicture}%
}
\caption{Geometric intuition. The projector eliminates gradient flow for $f(\tilde y^{(1)})$ along direction $v$ (orthogonal to local constraint manifold $\mathcal{M}$) because, for an idealized projector, $\partial \tilde y^{(1)} / \partial \hat y \cdot \delta v = 0$. This leaves only gradients of $f(\tilde y^{(1)})$ which are \emph{tangent} to the constraint manifold (direction $u$) to be backpropagated through the projection step into the neural backbone $N_\theta$.}
\label{fig:tangent_normal_intuition}
\end{figure}
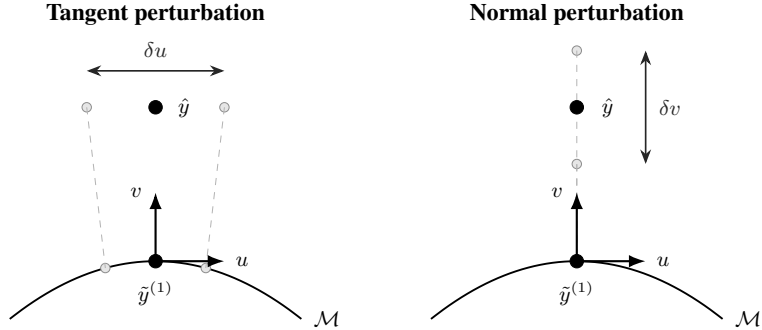

\paragraph{Empirical Demonstration.} In Fig.~\ref{fig:fhat_vs_ftilde}, we show three variants of \ours{} trained on the toy problem described in Appendix~\ref{app:intuition_setup}: (i) loss at $f(\hat y)$ (\ours{} default), (ii) loss using $f(\tilde y^{(1)})$, and (iii) loss using $f(\hat y)$ for the first half of training, switched to $f(\tilde y^{(1)})$ at half the epochs. Variant (i) converges normally. Variant (ii) is stuck near initialisation for the entire run: the gradient magnitude is roughly four orders of magnitude smaller than in (i). Variant (iii) shows that the failure mode is specific to early training: once the network has been pulled close to the feasible manifold under $f(\hat y)$, the projector's normal-attenuation no longer harms training.

\begin{figure}[!htbp]
\centering
\includegraphics[width=\linewidth]{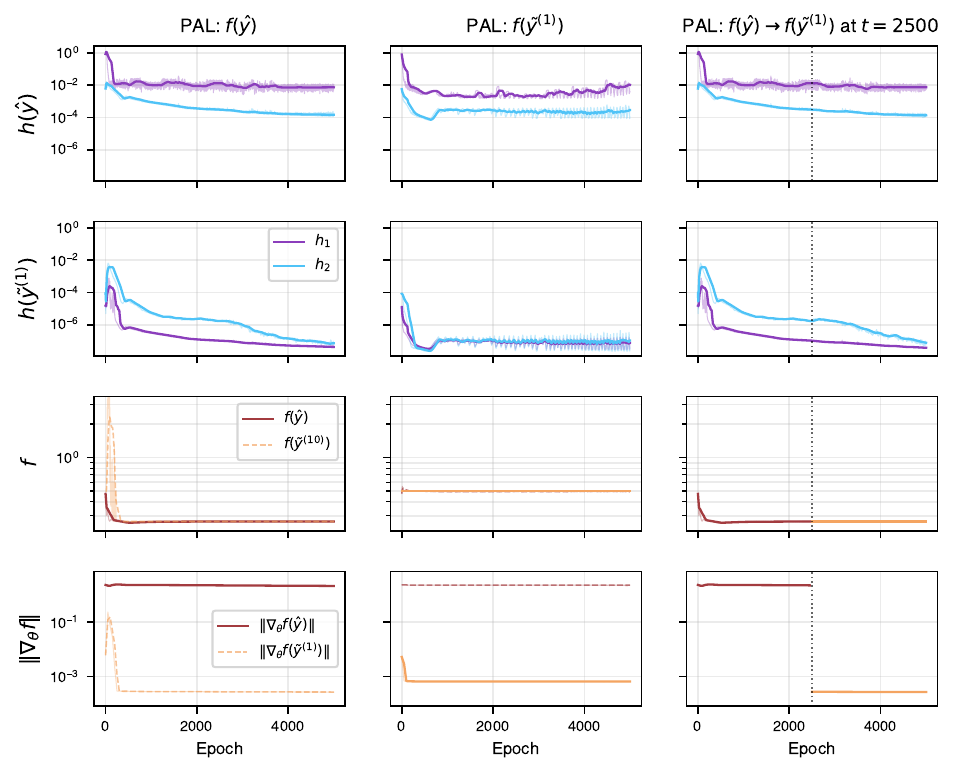}
\caption{\ours{} loss-placement comparison on the toy problem of Appendix~\ref{app:intuition_setup}. Rows from top: pre-projection residuals $h(\hat y)$, post-projection residuals $h(\tilde y^{(1)})$, objective values $f$, and objective gradient magnitudes (computed on a fixed evaluation batch). Columns: $f(\hat y)$ throughout (\ours{} default), $f(\tilde y^{(1)})$ throughout, and a midway switch from $f(\hat y)$ to $f(\tilde y^{(1)})$ at $t=2500$ (dotted vertical line). When using $f(\tilde y^{(1)})$ throughout, the objective never escapes the initialisation neighbourhood due to vanishing gradients normal to the manifold. The right column, having trained under $f(\hat y)$ first, retains its objective and feasibility quality after the switch, demonstrating that the pathology is specific to early training.}
\label{fig:fhat_vs_ftilde}
\end{figure}

\paragraph{Why do prior methods not report this?} We hypothesize that the same gradient-routing pathology affects every projection-based method that places the objective at the post-repair iterate $\tilde y^{(N)}$, but that each \ac{SOTA} method in the literature implicitly mitigates this issue with warm-up procedures: Both SnareNet \citep{chu2026snarenet} and ENFORCE \citep{lastrucci2025enforce} defer hard projection to late training. SnareNet uses adaptive constraint relaxation, and ENFORCE activates its projection module only after a warm-up phase. We hypothesize that DC3 \citep{donti2021dc3} and FSNet \citep{nguyen2025fsnet} are unaffected because they rely on gradient descent (FSNet wholly, DC3 for inequalities) where the differential of $N$ small gradient-descent steps does not suffer from the issues described above. 

\paragraph{Implications for \ours{}.} In practice, to circumvent this issue with \ours{}, there are three approaches to deal with the effect described above:

\begin{enumerate}
    \item \emph{Pre-projection objective}. This is what we report on using throughout in this paper and, while not perfect in theory, is the most straightforward solution in practice. Since we enforce $\|\Delta y^{(1)}\| < \tau_d$ using our adaptive loss (see weight update in Eq.~\ref{eq:pal_weight_update}), we assume $f(\hat y) \approx f(\tilde y^{(1)})$ for $\tau_d \ll 1$ and simply use this.
    \item \emph{Warm-up}. One could consider using $f(\hat y)$ in early training and switching to $f(\tilde y^{(1)})$ in late stages of training to combine the advantages of both. However, as Fig.~\ref{fig:fhat_vs_ftilde} shows, there is virtually no advantage over simply using $f(\hat y)$ throughout training when $\tau_d \ll 1$.
    \item \emph{Warm-up with Penalty Method}. Alternatively, one could simply skip projection in early training like many \ac{SOTA} methods do, and use either $f(\hat y)$ or $f(\tilde y^{(1)})$ later in training. For our experiments and algorithm presentation we omit this since it would add more complexity to our method. However, we would like to note that there would be some benefits to doing so also in terms of efficiency, since the additional compute of the single projection step would be eliminated during the warm-up phase. This could be especially relevant for $K \gg 1$, where assembly of the constraint Jacobian $J$ is expensive.

\end{enumerate}

\section{\acl{LM} Projector}
\label{app:projector}
We write $c \in \mathbb{R}^K$ for the constraint violation vector, with entries $|h_k(y)|$ for equalities and $\max(g_k(y), 0)$ for inequalities, and $K$ the number of constraints. 

\paragraph{Active Set.} The active constraint vector $c_\mathcal{A}$ collects all violated constraints $c(y) > 0$. $J_\mathcal{A}(y) := \partial c_\mathcal{A} / \partial y$ is the Jacobian of the active constraint vector. The local feasible manifold of the active constraints is $\mathcal{M}_\mathcal{A} := \{y : c_\mathcal{A}(y) = 0\}$.

The \ac{LM} step treats all active constraints as equalities and does not consider the inactive inequalities, i.e., they are deleted from the linear system. A mask is used for this using $c(\tilde y^{(i)})$ before every iterate. In effect, an inequality that a step overshoots enters the active set at the next iterate, and one that becomes satisfied leaves it. The active set can therefore switch between iterates at inference. In practice, we did not experience any issues with active-set switching. The convergence statement below assumes a fixed active set. The two alternative repair steps (\ac{SQP} and \acl{IP}) in Appendix~\ref{app:repair_variants} do not use this masking, since they both have native ways to account for non-violated inequalities (quadratic subproblem for \ac{SQP}, barrier weight for \ac{IP}). 

\paragraph{\acs{LM} Repair Step.} Starting from $\tilde y^{(0)} = \hat y$, the \ac{LM} projection applies:
\begin{equation}
\tilde y^{(i+1)}
=
\tilde y^{(i)}
-
J_\mathcal{A}(\tilde y^{(i)})^\top
\Bigl(
J_\mathcal{A}(\tilde y^{(i)})J_\mathcal{A}(\tilde y^{(i)})^\top + \lambda_i I
\Bigr)^{-1}
c_\mathcal{A}(\tilde y^{(i)}),
\label{eq:pal_projection_loop}
\end{equation}
with Yamashita--Fukushima~\citep{yamashita2001rate} damping
\[
\lambda_i = \delta \cdot \|c_\mathcal{A}(\tilde y^{(i)})\|^2,
\]
where $\delta > 0$ is a constant (we use $10^{-3}$). The projection is computed independently per batch element.

Each step of Eq.~\ref{eq:pal_projection_loop} equivalently solves
\begin{equation}
\Delta y^{(i)}
=
\arg\min_{\Delta y}
\tfrac{1}{2}\bigl\| c_\mathcal{A}(\tilde y^{(i)}) + J_\mathcal{A}(\tilde y^{(i)})\,\Delta y \bigr\|^{2}
+ \tfrac{\lambda_i}{2}\|\Delta y\|^{2}.
\label{eq:damped_ls}
\end{equation}
The Tikhonov term $\tfrac{\lambda_i}{2}\|\Delta y\|^{2}$ regularizes $J_\mathcal{A} J_\mathcal{A}^\top$ and bounds the step where the linearization is inaccurate. As $c_\mathcal{A} \to 0$ this term vanishes, \eqref{eq:damped_ls} reduces to the minimum-norm solution of $J_\mathcal{A}\,\Delta y=-c_\mathcal{A}$, i.e., a first-order approximation to orthogonal projection onto $\mathcal{M}_\mathcal{A}$.
The \ac{LM} projector is insensitive to the damping hyperparameter $\delta$ over multiple orders of magnitude but removing damping entirely ($\delta=0$ and $\lambda_{\min}=0$) degrades the objective on some seeds (see Appendix \ref{app:ablations}). On ill-conditioned constraints the undamped step fails, as we observe for the undamped completion of DC3 in Table~\ref{tab:methods-sweep-synthetic}.

Expanding $c_\mathcal{A}$ around $\tilde y^{(i)}$:
\[
c_\mathcal{A}(\tilde y^{(i+1)})
=
c_\mathcal{A}(\tilde y^{(i)})
+ J_\mathcal{A}(\tilde y^{(i)})\bigl(\tilde y^{(i+1)} - \tilde y^{(i)}\bigr)
+ \mathcal{O}(\|\Delta y^{(i)}\|^2).
\]
With damping the \ac{LM} step does not drive the linearized residual to zero, so $R_{i+1} = c_\mathcal{A}(\tilde y^{(i+1)})$ collects two contributions: a damping residual and the higher-order Taylor remainder $\mathcal{O}(\|\Delta y^{(i)}\|^2)$ under $c_\mathcal{A} \in C^1$ with locally Lipschitz Jacobian.

For the idealized fixed-active-set smooth case (Assumption~\ref{ass:pal_local}),~\citet{yamashita2001rate} establish local quadratic convergence of Eq.~\ref{eq:pal_projection_loop} to the active feasible set $\mathcal{M}_\mathcal{A}$ from a \emph{sufficiently close initialization}, which we use as motivation for our adaptive weight update rule.

\subsection{Convergence}
We fix a neighborhood $U \subset \mathbb{R}^d$ on which the active set $A$ is constant, and let
$c_\mathcal{A} : U \to \mathbb{R}^{K_\mathcal{A}}$
denote the corresponding smooth active residual map, with Jacobian
$J_\mathcal{A}(y) := \nabla c_\mathcal{A}(y)$.
The local active feasible set is
$\mathcal{M}_\mathcal{A} := \{ y \in U : c_\mathcal{A}(y)=0 \}$.

\begin{assumption}[Local active-set regularity]
\label{ass:pal_local}
There exists a point $y^\star \in \mathcal{M}_\mathcal{A}$ and a neighborhood $U$ of $y^\star$ such that:
\begin{enumerate}
\item the active set $A$ is fixed on $U$.
\item $c_\mathcal{A}$ is continuously differentiable on $U$, and $J_\mathcal{A}$ is Lipschitz continuous on $U$.
\item $c_\mathcal{A}$ satisfies a local error bound on $U$: there exists $\kappa>0$ such that
\[
\operatorname{dist}(y,\mathcal{M}_\mathcal{A}) \le \kappa \|c_\mathcal{A}(y)\| \qquad \forall y \in U.
\]
\end{enumerate}
\end{assumption}

The error bound in (3) is strictly weaker than requiring $J_\mathcal{A}$ to have full row rank on $\mathcal{M}_\mathcal{A}$, and accommodates redundant or rank-deficient constraint systems.

Under Assumption~\ref{ass:pal_local}, Theorem~2.1 of~\citet{yamashita2001rate} applies to the idealized fixed-active-set iteration: for all $\tilde y^{(0)}$ sufficiently close to $\mathcal{M}_\mathcal{A}$, the iterates of Eq.~\ref{eq:pal_projection_loop} converge quadratically to $\mathcal{M}_\mathcal{A}$. In particular, there exist $C, \rho > 0$ such that whenever $\operatorname{dist}(\tilde y^{(i)},\mathcal{M}_\mathcal{A}) \le \rho$,
\[
\operatorname{dist}(\tilde y^{(i+1)},\mathcal{M}_\mathcal{A})
\le
C \, \operatorname{dist}(\tilde y^{(i)},\mathcal{M}_\mathcal{A})^2.
\]
Combining with the error bound and the local Lipschitz property of $c_\mathcal{A}$, this yields $r_{i+1} \le M r_i^2$ with $M := LC\kappa^2$, where $r_i := \|c_\mathcal{A}(\tilde y^{(i)})\|$~\citep{yamashita2001rate}. This applies to the smooth fixed-active-set map $c_\mathcal{A}$, not automatically to the deployed practical projector with active-mask updates and numerical damping floor $\lambda_{\min}$.

\paragraph{Practical Proxy for the Local Basin.}
The radius $\rho$ is problem-specific and not directly observable. Accordingly, we do not claim that $R_{1,k} < \tau_c$ is sufficient for the Yamashita--Fukushima basin condition for the idealized fixed-active-set iteration. Instead, we use $R_{1,k}$ only as a practical heuristic for local convergence. If a single linearized repair step from $\hat y$ drives all constraint residuals below $\tau_c$, then the local linear model is accurate at $\hat y$, and the projector is likely operating in a regime of reliable local convergence. This is the signal used by \ours{}'s weight rule (Eq.~\ref{eq:pal_weight_update}), which treats constraint $k$ as easy to repair when its single-step residual falls below $\tau_c$. Empirically, for sufficiently small problem-dependent choices of $\tau_c$ (typically around $10^{-4}$), this proxy has aligned well with stable projector behavior across our experiments up to extremely non-linear constraints (high curvature) as shown in \S~\ref{sec:curvature}.

\section{{Inference Precision}}
\label{app:precision}

Here, we use checkpoints from the paper on the six synthetic benchmarks \texttt{S1}--\texttt{S6} which were trained with $\tau_c = \tau_d = 10^{-4}$. 
We show inference with iterates of the repair loop for $N \in \{1, 2, 5, 10, 100\}$ steps and record the maximum constraint violation over the test set of 64 queries per seed, averaged over 10 seeds. We report $\texttt{float64}$ and $\texttt{float32}$ implementation variants. 

Table~\ref{tab:precision} shows that for both $\texttt{float64}$ and $\texttt{float32}$, every benchmark plateaus at the numerical precision floor from $N \approx 2$ to $5$ onward, except for $\texttt{float64}$ on benchmark $\texttt{S5}$. We believe this is due to $\texttt{S5}$'s overdetermined active set. The violation decreases monotonically in $N$ on every benchmark.

Overall, this illustrates that the first-step residual target $\tau_c = 10^{-4}$ and the actual precision that can be reached for $N>1$ are not to be confused, and that \ours{} can consistently reach arbitrary precision. 

\begin{table}[H]
\centering
\footnotesize
\setlength{\tabcolsep}{6pt}
\caption{Maximum constraint violation over the test set against the number of inference repair steps $N$, over 10 seeds, for the trained \ours{} models from the paper on the synthetic benchmarks \texttt{S1}--\texttt{S6}.}
\label{tab:precision}
\begin{tabular}{lccccc}
\toprule
 & $N{=}1$ & $N{=}2$ & $N{=}5$ & $N{=}10$ & $N{=}100$ \\
\midrule
\multicolumn{6}{l}{\texttt{float64}} \\
\texttt{S1} (sphere track)      & $8.48\cdot10^{-4}$ & $2.46\cdot10^{-7}$ & $7.55\cdot10^{-16}$ & $7.55\cdot10^{-16}$ & $7.55\cdot10^{-16}$ \\
\texttt{S2} (active-set switch) & $1.36\cdot10^{-4}$ & $3.41\cdot10^{-6}$ & $2.22\cdot10^{-16}$ & $2.22\cdot10^{-16}$ & $2.22\cdot10^{-16}$ \\
\texttt{S3} (ill-cond.\ tube)   & $8.42\cdot10^{-6}$ & $1.03\cdot10^{-7}$ & $3.54\cdot10^{-15}$ & $2.22\cdot10^{-17}$ & $2.01\cdot10^{-17}$ \\
\texttt{S4} (qv coupling)       & $1.01\cdot10^{-2}$ & $1.26\cdot10^{-5}$ & $8.44\cdot10^{-16}$ & $9.10\cdot10^{-16}$ & $8.22\cdot10^{-16}$ \\
\texttt{S5} (overdetermined)    & $6.77\cdot10^{-3}$ & $4.98\cdot10^{-4}$ & $6.33\cdot10^{-8}$  & $8.81\cdot10^{-9}$  & $8.83\cdot10^{-11}$ \\
\texttt{S6} (redundant ineq.)   & $2.16\cdot10^{-3}$ & $1.45\cdot10^{-5}$ & $1.67\cdot10^{-16}$ & $1.39\cdot10^{-16}$ & $1.33\cdot10^{-16}$ \\
\midrule
\multicolumn{6}{l}{\texttt{float32}} \\
\texttt{S1} (sphere track)      & $8.48\cdot10^{-4}$ & $5.81\cdot10^{-7}$ & $3.58\cdot10^{-7}$ & $3.58\cdot10^{-7}$ & $3.58\cdot10^{-7}$ \\
\texttt{S2} (active-set switch) & $1.36\cdot10^{-4}$ & $3.54\cdot10^{-6}$ & $1.43\cdot10^{-7}$ & $1.43\cdot10^{-7}$ & $1.43\cdot10^{-7}$ \\
\texttt{S3} (ill-cond.\ tube)   & $8.49\cdot10^{-6}$ & $1.47\cdot10^{-7}$ & $3.13\cdot10^{-8}$ & $3.05\cdot10^{-8}$ & $2.76\cdot10^{-8}$ \\
\texttt{S4} (qv coupling)       & $1.01\cdot10^{-2}$ & $1.37\cdot10^{-5}$ & $3.96\cdot10^{-7}$ & $4.49\cdot10^{-7}$ & $4.15\cdot10^{-7}$ \\
\texttt{S5} (overdetermined)    & $6.74\cdot10^{-3}$ & $5.74\cdot10^{-4}$ & $2.80\cdot10^{-7}$ & $2.62\cdot10^{-7}$ & $2.38\cdot10^{-7}$ \\
\texttt{S6} (redundant ineq.)   & $2.16\cdot10^{-3}$ & $1.50\cdot10^{-5}$ & $1.24\cdot10^{-7}$ & $1.19\cdot10^{-7}$ & $1.19\cdot10^{-7}$ \\
\bottomrule
\end{tabular}
\end{table}

\section{Residual Intuition on a Toy Problem}
\label{app:intuition_setup}

Fig.~\ref{fig:residual_intuition} compares three constraint-handling strategies on a single 2D problem with one stiff linear equality and one mildly nonlinear equality:
\[
\min_y \; (y_1 - 0.5)^2 + (y_2 - 0.5)^2
\quad \text{s.t.} \quad
\begin{aligned}
& h_1(y) = 10\,y_1 = 0, \\
& h_2(y) = 0.01\sin(5\,y_2) + 0.1\,y_1 = 0.
\end{aligned}
\]
The objective pulls the prediction toward $(0.5, 0.5)$. $h_1$ is stiff and linear (well-suited to a single \ac{LM} step), while $h_2$ is mildly nonlinear in $y_2$ and weakly couples back to $y_1$. The neural backbone is a 3-layer \acs{MLP} of width 128 with ReLU activations, trained with Adam at learning rate $10^{-4}$ for 5000 epochs and batch size 100. Conditions $z$ are sampled from $\mathcal{N}(0, I_2)$ and supplied as input but do not enter the problem.

All three methods use the \emph{same} \ac{LM} projector with identical hyperparameters (damping $\delta = 10^{-3}$, floor $10^{-6}$) both during training (where invoked) and at evaluation. They differ only in (i) how often the projector is invoked during training, and (ii) the training loss:

\begin{itemize}
    \item \textbf{ALM+Bolt-On.} No projection during training. MDMM-style dual ascent~\citep{platt1988constrained} with fixed quadratic penalty: $\mathcal{L} = f + \lambda^\top h + \tfrac{\rho}{2}\|h\|^2$, $\lambda \leftarrow \lambda + \eta_\lambda\,h$, with $\eta_\lambda = 10^{-2}$ and $\rho = 1$. The projector is bolted on only at evaluation. (Basir \& Senocak's APU~\citep{basir2023adaptive} was unstable on this toy problem, so we substitute MDMM as a closely related fixed-penalty variant.)
    \item \textbf{Soft Loss.} Ten chained \ac{LM} projection steps in training with live gradient through the displacement chain, trained with $\mathcal{L} = f + \lambda_d\,\|\hat y - \tilde y^{(10)}\|_2 + \lambda_c\,\|h(\tilde y^{(10)})\|_2$ at $\lambda_d = 0.5$, $\lambda_c = 1.0$.
    \item \textbf{\ours{}.} A single \ac{LM} projection step in training, with the gated log-decade weight update of Eq.~\ref{eq:pal_weight_update}: $\gamma = 10^{-2}$, $D_{\max} = 1$, $\tau_d = 10^{-1}$.
\end{itemize}

\section{Synthetic Benchmark Details}
\label{app:synthetic_setups}

We provide formal definitions for the conditional synthetic benchmarks. All follow the convention $h(y, z) = 0$ for equalities and $g(y, z) \le 0$ for inequalities. 

Five of the six benchmarks (\texttt{S1}--\texttt{S4}, \texttt{S6}) share a \emph{degenerate-objective} construction: the design variable $y$ is partitioned into a \emph{head} block on which the objective depends and a \emph{tail} block that does not appear in the objective but is coupled to the head through one or more constraints. Projecting an arbitrary $\hat y$ onto the constraint set generally moves the head coordinates away from their optimum, so the backbone network $N_\theta$ must drive the constraint residual toward zero on its own, since projection alone is not enough. \texttt{S5} is an overdetermined consistent equality system whose feasible set is finite per $z$, isolating the projection geometry from objective effects.

\paragraph{\texttt{S1} (Sphere constraint).}
Design variable $y \in \mathbb{R}^4$, $K=1$, condition $z \in [0.5, 2]$. One equality, no inequalities.
\[
\min_y \; \bigl(1 - (y_1 - s_1(z))\bigr)^2 + 100\bigl((y_2 - s_2(z)) - (y_1 - s_1(z))^2\bigr)^2
\quad \text{s.t.} \quad
\textstyle\sum_{i=1}^{4} y_i^2 = z^2,
\]
where the shift $s(z)$ is chosen so the global minimizer is $y^\star(z) = (z/\sqrt{2}, z/\sqrt{2}, 0, 0)$. The objective depends only on $(y_1, y_2)$, and the sphere projection scales all four coordinates uniformly, so any tail offset displaces the head off-optimum after projection.

\paragraph{\texttt{S2} (Probability simplex).}
Design variable $y \in \mathbb{R}^{10}$, $K=11$, condition $z \in [-1, 1]^2$. One equality and ten non-negativity inequalities.
\[
\min_y \; \bigl(s(z) - y_1\bigr)^2 + 10\sum_{i=1}^{4}(y_{i+1} - y_i^2)^2
\quad \text{s.t.} \quad
\textstyle\sum_{i=1}^{10} y_i = 1, \quad y_i \ge 0,
\]
with $s(z) = 0.4 + 0.05\,z_1 + 0.02\,z_2$. Only the head $(y_1, \ldots, y_5)$ enters the chained-Rosenbrock objective, and $\sum_{i \le 5} y_i^\star \approx 0.586$, so the simplex equality forces nontrivial mass onto the tail $(y_6, \ldots, y_{10})$, which does not enter the objective.

\paragraph{\texttt{S3} (Ill-conditioned head with mixing equality).}
Design variable $y \in \mathbb{R}^{20}$, $K=6$, condition $z \sim \mathrm{LogU}(0.1, 1)$. One mixing equality, one nonlinear tube inequality, four near-parallel linear inequalities.
\[
\min_y \; \tfrac{1}{2}\sum_{i=1}^{10} z^{2(i-1)/9}\,\bigl(y_i - t_i(z)\bigr)^2
\quad \text{s.t.} \quad
\begin{aligned}
& a_0^\top y = a_0^\top t(z), \quad \|y_{11:20}\|^2 \le z^4, \\
& a_j^\top y \le a_j^\top t(z) + 0.2 \quad (j = 1, \ldots, 4),
\end{aligned}
\]
with target $t_i(z) = z$ for $i \le 10$ and $t_i(z) = z^2/\sqrt{10}$ for $i > 10$. The head Hessian is diagonal with condition number $z^{-2}$, ranging up to 100 at $z = 0.1$. The mixing equality couples the head $y_{1:10}$ to the tail $y_{11:20}$ (which does not enter the objective), so any tail movement to satisfy the equality feeds back into the head through the projection step.

\paragraph{\texttt{S4} (Quadratic objective with sinusoidal coupling).}
Design variable $y = (u, v) \in \mathbb{R}^{15} \times \mathbb{R}^{15}$, $K=26$, condition $z \in [0.5, 1.5]^5$. Twenty-five equalities and one inequality.
\[
\min_y \; \tfrac{1}{2}\bigl(v - v^\star(z)\bigr)^\top W_v \bigl(v - v^\star(z)\bigr)
\quad \text{s.t.} \quad
\begin{aligned}
& A\,u = \tilde A\,z, \quad v_j = \sin(u_j) + (Bz)_j, \\
& \|v\|^2 \le \|v^\star(z)\|^2 + 1,
\end{aligned}
\]
with $A \in \mathbb{R}^{10 \times 15}$ rank $10$ and $W_v$ SPD with $\mathrm{cond}(W_v) \approx 5$. The block $u$ does not enter the objective, but the main equality is underdetermined in $u$, so projecting an arbitrary $\hat u$ onto the constraint produces a point that generically differs from $u^\star$, and the sinusoidal coupling then forces $v$ off its quadratic optimum. 

\paragraph{\texttt{S5} (Overdetermined system).}
Design variable $y \in \mathbb{R}^4$, $K=5$, condition $z \in [0.5, 1.5]^5$. Five equalities ($m > n$), no inequalities, mutually consistent at a $y^\star(z) \in \mathbb{R}^4$ parameterized by $z$.
\[
\min_y \; \tfrac{1}{2}\bigl(y - y^\star(z)\bigr)^\top P \bigl(y - y^\star(z)\bigr)
\quad \text{s.t.} \quad
\begin{aligned}
& y_1^2 + y_2^2 = z_1^2, \quad y_3^2 + y_4^2 = z_2^2, \\
& y_1 y_3 = \alpha(z), \quad y_2 y_4 = \beta(z), \\
& y_1 y_4 - y_2 y_3 = \delta(z),
\end{aligned}
\]
where $\alpha(z) = y^\star_1 y^\star_3$, $\beta(z) = y^\star_2 y^\star_4$, and $\delta(z) = y^\star_1 y^\star_4 - y^\star_2 y^\star_3$ are the corresponding bilinear forms of $y^\star(z)$. Because $m > n$, the implicit-function partition required by DC3 does not apply. The feasible set is $\{\pm y^\star(z)\}$, so any successful projection lands at one of these two points regardless of $\hat y$.

\paragraph{\texttt{S6} (Coupling equalities with redundant inequalities).}
Design variable $y \in \mathbb{R}^{50}$, $K=90$, condition $z \in [0.5, 1.5]^{20}$. Fifteen coupling equalities and 75 inequalities (25 distinct $+$ 50 near-redundant copies).
\[
\min_y \; \tfrac{1}{2}\sum_{i=16}^{50}\bigl(y_i - t_i(z)\bigr)^2
\quad \text{s.t.} \quad
\begin{aligned}
& y_i - \tfrac{1}{2} y_{i+15} = z_{(i \bmod 20)} - \tfrac{1}{2} z_{((i+15) \bmod 20)}, \;\; i = 1, \ldots, 15, \\
& a_j^\top y \le a_j^\top t(z) + 0.2, \;\; j = 1, \ldots, 25,
\end{aligned}
\]
The objective depends only on the tail $y_{16:50}$, and the head $y_{1:15}$ is pinned to the tail via the coupling equalities. The redundant inequality block stresses projector behavior (rank-deficient active sets).

\section{Chip Placement Case Study}
\label{app:b105}

Here, we show one additional application where the constraint and objective evaluation is relatively cheap, so that methods with unrolled repair in training are memory-feasible. This application is a macro-placement problem inspired by Intel's FloorSet~\citep{floorset2024}. The task is to place $B\!=\!30$ rectangular blocks on a $200\!\times\!200$ canvas to minimise a wirelength proxy (a weighted sum of pairwise Manhattan distances between connected block centres and boundary pins) plus a soft bounding-box area penalty (LSE-smoothed).
A single equality constraint requires the total overlap volume across all $B(B{-}1)/2$ block pairs to be zero.
The design vector $y \in \mathbb{R}^{3B}$ ($3B\!=\!90$ variables) encodes block centres $(x_i, y_i)$ and raw widths $s_i$. Widths decode as $w_i\!=\!\mathrm{softplus}(s_i)+0.5$, and heights follow from a fixed per-instance target area, keeping all boxes non-degenerate and area-preserving. We want to point out that this benchmark is a lightweight component placement task and a strongly simplified FloorSet derivative. The full FloorSet suite additionally considers grouping, multi-instantiation, and pre-placement constraints.

\begin{figure}[H]
\noindent
\begin{minipage}[c]{0.40\textwidth}
\centering
\includegraphics[width=\linewidth]{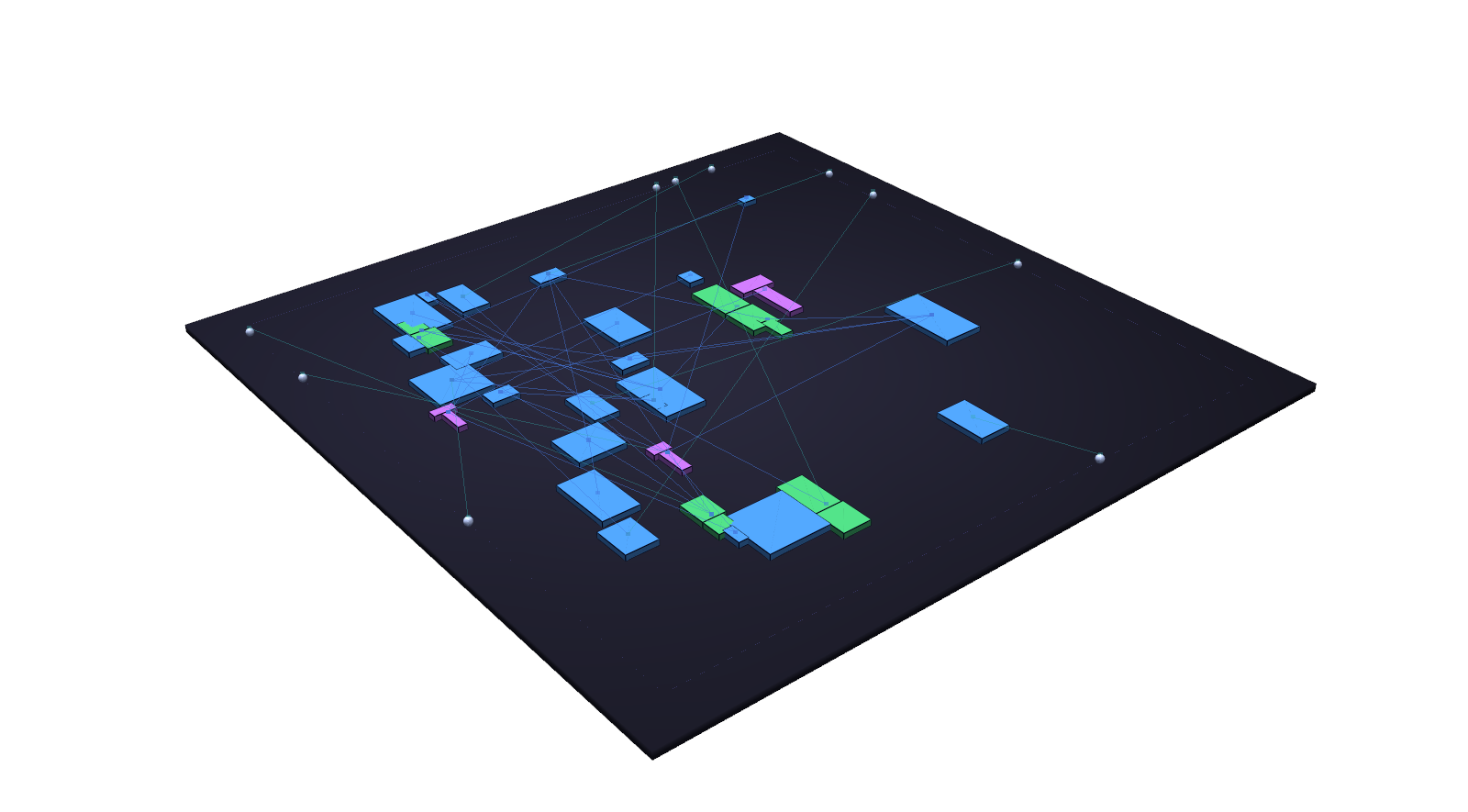}
\end{minipage}\hspace{0.02\textwidth}%
\begin{minipage}[c]{0.57\textwidth}
\centering
\footnotesize
\setlength{\tabcolsep}{3pt}
\begin{tabular}{@{}lcccc@{}}
\toprule
& Obj.\ $\times 10^{3}$\,$\downarrow$ & Feas.\ (\%)\,$\uparrow$ & Wall-Clock\,$\downarrow$ & $n$ \\
\midrule
\textbf{\ours}   & $\mathbf{1.16{\pm}0.04}$ & $\mathbf{100{\pm}0}$ & $1\mathrm{m}18\mathrm{s}{\pm}6\mathrm{s}$ & 10 \\
ALM+Bolt-On      & $1.19{\pm}0.05$ & $80{\pm}42$ & $\mathbf{40\mathrm{s}{\pm}2\mathrm{s}}$ & 10 \\
FSNet            & $5.27{\pm}0.31$ & $0{\pm}0$ & $1\mathrm{h}36\mathrm{m}{\pm}6\mathrm{m}$ & 5 \\
SnareNet         & $1.35{\pm}0.02$ & $0{\pm}0$ & $1\mathrm{h}26\mathrm{m}{\pm}1\mathrm{m}$ & 5 \\
\bottomrule
\end{tabular}
\end{minipage}
\caption{Candidate chip placement for \texttt{E4} (left) and full results (right). \ours{} and ALM+Bolt-On run at 10 seeds, FSNet and SnareNet at 5 seeds. All methods use \emph{canonical default hyperparameters without per-method tuning}.}
\label{fig:b105_appendix}
\end{figure}

\paragraph{Results.} \ours{} reaches 100\% feasibility while FSNet and SnareNet reach 0\% feasibility at substantially higher wall-clock time than \ours{} due to the compute required by the unrolled repair. We show ALM, \ours{}, and SnareNet with 2000 training epochs each, while FSNet is shown with only 100 training epochs in order to keep wall-clock times manageable. We use canonical default hyperparameters for each method, i.e., no per-method tuning. Despite the problem having only one inequality constraint, both SnareNet and FSNet reach $0\%$ feasibility. We did not investigate in further detail why this is the case, but believe both FSNet and SnareNet could perform better on this task if their hyperparameters were tuned extensively. However, orthogonal to this, we want to point out that \ours{} is \emph{significantly} faster to train and achieves 100\% feasibility despite not being tuned for the problem at hand. Computationally, this benchmark is very well suited to \ours{} since it only has a single constraint ($K=1$), meaning that the constraint Jacobian $J$ is cheap to build. ALM+Bolt-On reaches 100\% feasibility on 8 of 10 seeds, but the remaining 2 seeds carry very high constraint violations, which is what drives the mean to $80\%\!\pm\!42\%$ in Fig.~\ref{fig:b105_appendix}. This experiment was run in parallel (across seeds and methods) on a 288-core ARM CPU node. Cross-method ratios for wall-clock time are reliable, but absolute wall-clock numbers are hardware- and contention-specific.

\section{Engineering Benchmark Details}
\label{app:engineering_setups}

\subsection{Unmanned Aircraft}
\label{app:b101}

\begin{figure}[h!]
  \centering
  \includegraphics[width=\linewidth]{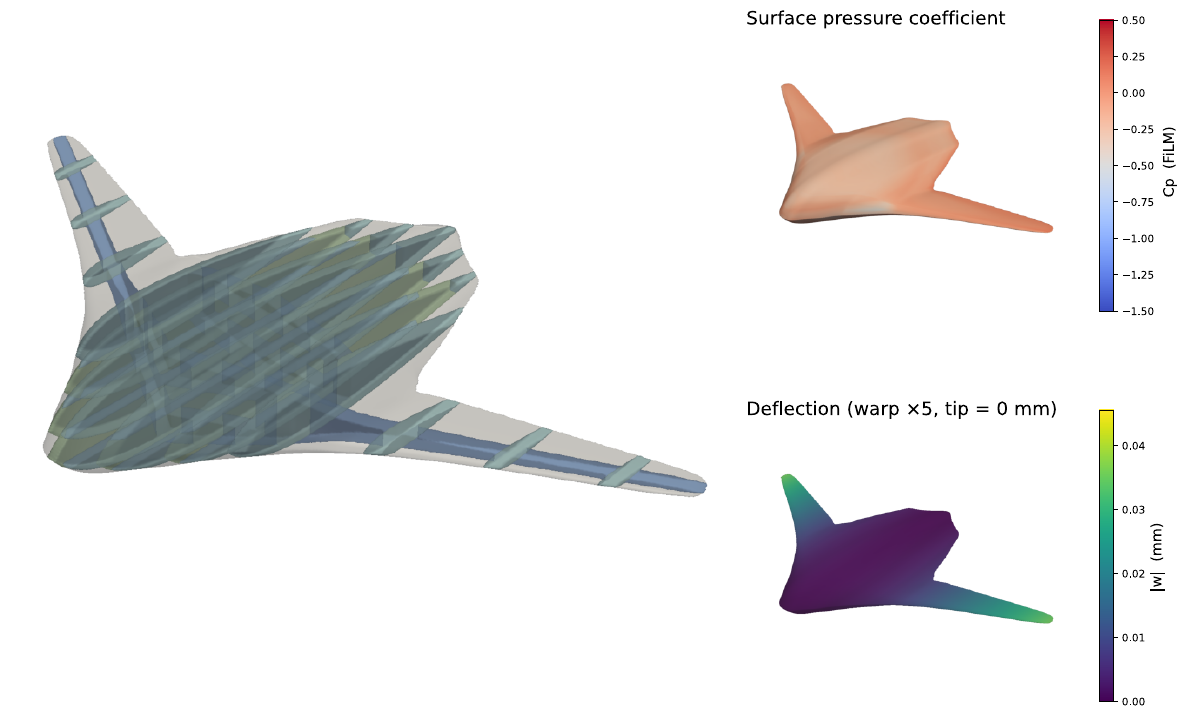}
  \caption{\texttt{E1} unmanned aircraft benchmark, candidate design.
  \textbf{Left:} outer mould-line geometry from the analytical B\'ezier
  representation, with the decoded internal wingbox (skin, spars, ribs)
  overlaid in blue/green. \textbf{Top right:} surface
  pressure coefficient $c_p$ predicted by the FiLM-modulated BlendedNet
  surrogate $\Phi^{\theta}_{\text{cp}}$~\citep{sung2025blendednet} at iso-sliced surface points.
  \textbf{Bottom right:} structural tip deflection $|w(y)|$ from beam
  integration of the pressure loads through the predicted section properties,
  warp-amplified $5\times$ for visualisation. The tip-deflection constraint
  reads $w_{\text{tip}}$ at $y=b/2$.}
  \label{fig:b101_appendix}
\end{figure}

\begin{figure}[h!]
  \centering
  \resizebox{\linewidth}{!}{\begingroup
\tikzset{
  font=\small,
  node distance=6mm and 10mm,
  box/.style={draw, rounded corners=1pt, minimum height=6mm, inner sep=3pt, align=center},
  geo/.style={box, densely dashed},
  ana/.style={box},
  sur/.style={box, fill=black!10},
  gen/.style={box, pattern=north east lines, pattern color=black!35},
  res/.style={box, double, double distance=0.6pt},
  arr/.style={-{Latex[length=2mm]}, semithick},
  lbl/.style={font=\scriptsize, fill=white, inner sep=1pt},
  leg/.style={font=\scriptsize, minimum height=4mm, minimum width=6mm, inner sep=1pt, rounded corners=1pt},
}
\tikzset{
  inp/.style={font=\small, inner sep=2pt, align=right},
  outv/.style={font=\normalsize, inner sep=2pt},
  bus/.style={semithick},
}
\begin{tikzpicture}[x=1mm,y=1mm]
\node[inp] (in1) at (0,6)   {$\mathbf{x}_{\mathrm{skin}},\,L,\,\alpha_{cr}$};
\node[inp] (in3) at (0,-11) {$\mathbf{x}_{\mathrm{bat}}$};
\node[inp] (in2) at (0,-24) {$\mathbf{z},\,\mathbf{x}_{\mathrm{walls}}$};
\draw[decorate,decoration={brace,amplitude=1.5mm,mirror}] ([xshift=-1.5mm]in1.west|-in1.north) -- ([xshift=-1.5mm]in1.west|-in2.south) node[midway,left=2mm,rotate=90,anchor=south]{Design $y$};
\node[sur] (aero) at (27,17)  {Aero-\\dynamics\\$\Phi^{\theta}_{\mathrm{aero}}$};
\node[sur] (sdf)  at (27,0)   {Neural\\implicit\\SDF};
\node[geo] (iso)  at (49,0)   {Iso-\\slicing};
\node[sur] (cp)   at (73,0)   {Surface\\pressure\\$\Phi^{\theta}_{\mathrm{cp}}$};
\node[sur] (sec)  at (27,-24) {Structural\\properties\\$\Phi^{\theta}_{\mathrm{sec}}$};
\node[ana] (beam) at (101,0)   {Beam\\integration\\\& mass};
\node[res] (range)  at (129,30)  {Range};
\node[res] (lift)   at (129,17)  {Lift\\balance};
\node[res] (strain) at (129,0)   {Strain};
\node[res] (tip)    at (129,-14) {Tip\\deflection};
\node[outv] (f) at (145,30) {$f$};
\node[outv] (c) at (145,0)  {$c$};
\draw[arr] (in1.east) -- ++(3,0) |- (aero.west);
\draw[arr] (in1.east) -- ++(3,0) |- (sdf.west);
\draw[arr] (in2.east) -- (sec.west);
\draw[arr] (in3.east) -| ([xshift=-3.5mm]beam.south);
\draw[arr] (sdf) -- node[lbl,above]{$\mathbf{d}$} (iso);
\draw[arr] (iso) -- node[lbl,above]{$\mathbf{r}_s,\mathbf{n}_s$} (cp);
\draw[arr] (cp) -- node[lbl,above]{$q_z, m, x_{cp}$} (beam);
\draw[arr] (sec.east) -| node[lbl,pos=0.3,above]{$A, I, J$} ([xshift=3.5mm]beam.south);
\draw[arr] (beam) -- node[lbl,above]{$\sigma_{vm}$} (strain);
\draw[arr] (beam.east) -- ++(4,0) |- node[lbl,pos=0.8,above]{$w_{\mathrm{tip}}$} (tip.west);
\draw[arr] (beam.north) |- node[lbl,pos=0.7,below]{$m_{\mathrm{tot}}$} ([yshift=-1.2mm]lift.west);
\draw[arr] (aero.east) -- ++(4,0) |- node[lbl,pos=0.6,above]{$C_L$} ([yshift=1.2mm]lift.west);
\draw[arr] (aero.north) |- node[lbl,pos=0.75,above]{$C_L/C_D$} (range.west);
\draw[arr] (range.east) -- (f);
\draw[bus] (lift.east) -- (141,17) coordinate (bl);
\draw[bus] (tip.east) -- (141,-14) coordinate (bt);
\draw[bus] (bl) -- (bt);
\draw[arr] (strain.east) -- (c);
\end{tikzpicture}
\endgroup}
  \caption{\texttt{E1} evaluator $\Phi$, from the design vector $y$ to the objective $f$ and constraints $c$. Dashed: geometry processing. Plain: analytical. Grey: neural surrogate. Hatched: generative model. Double border: constraint or objective.}
  \label{fig:e1_evaluator}
\end{figure}
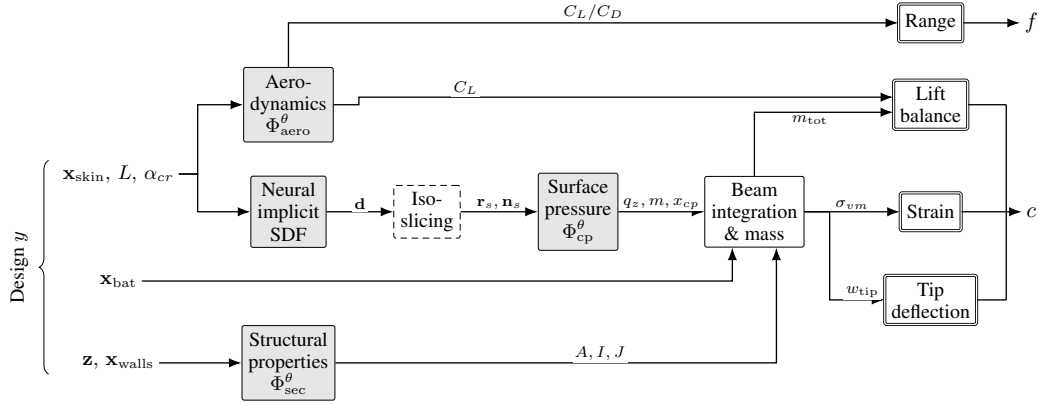

We now describe the benchmark $\texttt{E1}$ in detail. This application aims to maximize the range on an unmanned \ac{BWB} aircraft under structural constraints. Here, we mostly describe the evaluator $\Phi$, which is made up of multiple \emph{pre-trained} generative models and neural surrogates, specialized for various aspects of aircraft design. Fig.~\ref{fig:e1_evaluator} shows the arrangement of these submodels. Fig.~\ref{fig:b101_appendix} shows a candidate design (geometry, surface pressure, tip deflection). The data and trained models are available in the supplied code repository.

\subsubsection{Geometry} 
The geometry for this application is based on BlendedNet~\citep{sung2025blendednet}. In order to achieve differentiability for our evaluator, we use geometry representations based on implicits (\acp{SDF}). We employ two different representations: a neural implicit \ac{SDF} net, and an analytical implicit shape representation. 

\paragraph{Neural Implicit BWB Geometry.} First, we use a neural implicit representation for fast, differentiable signed-distance queries used inside the optimization loop during training (iso-slicing, surface sampling of the FiLM model, geometry-dependent constraints). We use a conditional \acs{MLP} with GELU activations and Fourier positional encoding on the spatial input \citep{tancik2020fourier,mildenhall2020nerf}. This maps spatial points $\mathbf{r}\in\mathbb{R}^3$ and the 9-D shape conditioner $\mathbf{x}_{\text{skin}}=(B_1,B_2,B_3,C_2,C_3,C_4,S_1,S_2,S_3)$, which describe body widths, chords, sweeps, as described by \citet{sung2025blendednet}. We use an additional parameter $L$ for isotropic scaling of the geometry. For each query (in practice, this is highly parallelized), a signed distance $\mathbf{d}\in\mathbb{R}^1$ (the distance to the surface) is returned. We use symmetry as a prior for added computational efficiency, such that $y\!\to\!|y|$ is enforced inside the network. The neural implicit representation is trained on the geometry dataset (meshes) provided in BlendedNet~\citep{sung2025blendednet}. 

\paragraph{Analytical B\'ezier SDF.} Similarly, this maps $\mathbf{r},  \mathbf{x}_{\text{skin}} \mapsto \mathbf{d}$ and serves as a high-fidelity reference geometry used for visualization (rendered with PyVista~\citep{sullivan2019pyvista}). This representation is much more computationally expensive because each query
rebuilds B\'ezier patches and six CSG operations, and is therefore not used live in training. The 9 design parameters drive three spanwise wing panels, each a bicubic B\'ezier surface patch (4$\times$4$\times$3 control net) per upper/lower surface, anchored at four spanwise stations with fitted airfoil control points. For tangent handling between these patches at the trailing and leading edges (to match the planform described by \citet{sung2025blendednet}), we use AdvLink-style in/out sweep tangents per station that replicate the OpenVSP~\citep{mcdonald2022openvsp} tangent conventions. We use per-panel surface SDF $\cap$ y-slab (smooth) $\cap$ swept TE half-space (sharp) to make a smooth union over panels, then mirror this across the $xz$ plane to make a smooth union of the two halves into the full aircraft. To find the bicubic B\'ezier control points, we performed regression on the airfoil described by \citet{sung2025blendednet}. Since the splines only use 4 control points, we cannot model the airfoil exactly, resulting in a small RMSE of $\approx 2.7$\,mm/m (body) and $\approx 2.0$\,mm/m (tip).

\paragraph{Coordinate Frame.} We denote x-direction as the direction of travel for the \ac{BWB}, y-direction spanwise, and $z$-direction up/down (thickness-wise). 

\subsubsection{Aerodynamics}
We use a simple MLP to model global aerodynamic parameters given a geometry ($ \mathbf{x}_{\text{skin}}$) and flight state, i.e., no geometry processing is used for these predictions (instead, geometry is needed for aero-structural coupling, more on this below). We model lift, drag, and pitching-moment coefficients $(C_L, C_D, C_M)$ with a simple \acs{MLP} $\Phi^{\theta}_{\text{aero}}$ that feeds a lift-balance constraint and a Breguet-range objective. Inputs are the 9-D shape conditioner $\mathbf{x}_{\text{skin}}\in[-1,1]^9$ from \citet{sung2025blendednet}. This is the same geometry parametrization the implicit representation above consumes. Additionally, we use the 4-D flight state $(\mathrm{Ma},\,h,\,\alpha_{cr}^{\deg},\,\log_{10} L)$, where $\mathrm{Ma}$ is derived from $(V,h)$ via a standard-atmosphere helper, $\alpha_{cr}$ is converted from radians, and the characteristic length $L$ is log-transformed (length scales span roughly one decade across training). For training data, we use the cases provided by \citet{sung2025blendednet}. Since this is a pure \acs{MLP} with smooth activations, we can differentiate $(C_L, C_D, C_M)$ w.r.t.\ $(\mathrm{Ma},\,h,\,\alpha_{cr}^{\deg},\, L)$.

\subsubsection{Surface Pressure}

We use the model denoted as $\Phi^{\theta}_{\text{cp}}$ to determine surface pressure on the \ac{BWB} (top-right panel of Fig.~\ref{fig:b101_appendix}), which is then used to determine wing deformation and strain in conjunction with our structural model (see below). This model is an exact training replica of the FiLM-modulated \acs{MLP} from BlendedNet~\citep{sung2025blendednet}. The model is queried by the iso-slice mechanism (described below) with surface points $\mathbf{r}_s$ and surface
normals $\mathbf{n}_s$ and a condition vector containing $\alpha_{cr}$, the flow state, and shape parameters. The model returns the surface pressure coefficient $c_p$ together with chordwise and vertical shear coefficients $(c_{f,x}, c_{f,z})$. These are combined into the per-point surface force density $\mathbf{f} = -c_p\,\mathbf{n} + (c_{f,x}, 0, c_{f,z})$ before chordwise quadrature, so both pressure and skin friction enter the structural load path. Global drag is still taken from $\Phi^{\theta}_{\text{aero}}$ (see above), not integrated from the FiLM shear field.

\subsubsection{Iso-slicing}
In order to couple the surface pressure model (FiLM BlendedNet) and our structural model, we employ a deterministic geometry extraction algorithm that queries the neural SDF using a quadtree-based marching-squares level-set extraction to find a discrete set of surface query points  $\mathbf{r}_s$ with outward normals $\mathbf{n}$ and arc-length integration weights, suitable for chordwise quadrature of the FiLM predictions.
At each spanwise station $y_i$, the BWB geometry is sliced with this iso-contour primitive $\texttt{prog.isocontour(plane='xz', stations=y\_unit, normal\_mode='3d')}$, which returns 2D polylines on the chordwise plane plus matching 3D outward surface normals and per-segment arc-length weights.
Default quadtree settings use base resolution 8 with 5 levels of refinement, so the SDF is queried repeatedly per slice. After extraction, each chordwise polyline $(x, z)$ is lifted to 3D by inserting the unit-frame spanwise station value as $y$. The station coordinate is detached from the gradient graph since it is a sampling location, not a design variable, such that the gradient graph only routes to $\mathbf{x}_{\text{skin}}, L, \alpha_{cr}$. At queried stations beyond the local span the iso-contour returns no polyline. The corresponding station's load contributions are zeroed and stay differentiable. Downstream, the outputs of the iso-slicing procedure $(\mathbf{r}_s, \mathbf{n}, w)$ are fed into the surface pressure model $\Phi^{\theta}_{\text{cp}}$ to determine $c_p$, which is then integrated chordwise with the arc-length weights into per-station aerodynamic loads. These are shear load, moment, and center of pressure $(q_z, m, x_{cp})$. 

\subsubsection{Structural Model}

The structural model of the \ac{BWB} consists of a generative model, $\Phi^{\theta}_{\text{dec}}$ (\ac{VAE}~\citep{wellingvae} decoder) to encode valid structural internals layouts conditional on the aircraft skin geometry $\textbf{x}_{\text{skin}}$ and a surrogate model $\Phi^{\theta}_{\text{sec}}$ (\acs{MLP}) to predict structural properties directly from the latent of this encoder/decoder. Data to train the \ac{VAE} was obtained through rejection sampling of valid/invalid internal layout designs. 

\paragraph{Architecture.} The \ac{VAE} encoder (used in training only) maps $(\textbf{x}_{\text{struct}}, \textbf{x}_{\text{skin}}) \to \textbf{z}\in \mathbb{R}^d$. The decoder maps $\Phi^{\theta}_{\text{dec}}$: $(\mathbf{z}, \textbf{x}_{\text{skin}}) \to \textbf{x}_{\text{struct}}\in \mathbb{R}^9$, the reconstructed raw wingbox layout (skin, spar, rib geometry), shown overlaid in the left panel of Fig.~\ref{fig:b101_appendix}. The decoder is used for training and visualization only. In the evaluator, the latent $\mathbf{z}$ is part of the design vector and feeds the structural properties predictor directly. As alluded to, the structural properties predictor maps $\Phi^{\theta}_{\text{sec}}$: $(\mathbf{z}, \textbf{x}_{\text{skin}}, \mathbf{x}_{\text{walls}}, y) $ to structural properties, where $\mathbf{x}_{\text{walls}}$ is the design vector of (uniform) wall thickness of the \ac{BWB} skin, front spar, and rear spar. For any given design at each training epoch, the structural model is used to evaluate structural properties per station. Additionally, we use a dedicated head to predict the overall volume of the inner ribs, which we assume do not contribute to the global stiffness constraints we evaluate, which is a conservative assumption. This dedicated head maps $(\mathbf{z}, \textbf{x}_{\text{skin}}, t_{\text{rib}}) \to V_{\text{ribs}}$, where the rib volume $V_{\text{ribs}}$ is a single scalar. This head is not used in the evaluator, and rib mass is not included in the mass aggregation below. All four components of the structural model architecture are plain GELU \acsp{MLP}. We use latent dimension $\dim(\textbf{z})=16$. The model explicitly does not handle the thickness of the ribs and spars, since skin thickness does not affect the geometric validity of structural internals and would have made rejection sampling much more expensive (curse of dimensionality) without additional benefit. Instead, these are concatenated after the decoder. For efficiency, the predictor is called once flattened over $B \cdot N_y$ stations: $\mathbf{z}$ and $\textbf{x}_{\text{skin}}$ are broadcast across $N_y$ Sobol-sampled spanwise stations and the predictor handles these in a single batched forward.
Training data is based on quadtree-based marching-squares material point assignment and computing geometric section properties on the resulting cross-sections. The surrogate's ground truth is therefore based on SDF evaluation and beam-section integrals over the built-up wingbox, not FEA. The 11-D output covers cross-section area $A$ (used to estimate structural mass via spanwise integration), centroid offsets $(u_{cg}, v_{cg})$, principal bending stiffnesses $I_{uu}$ and $I_{vv}$, the per-axis first moments $Q_u^{\max}, Q_v^{\max}$ (used in the transverse-shear formula below, with $Q_{\max}\!=\!\max(Q_u^{\max}, Q_v^{\max})$), and torsional stiffness $J$. The shear centre is not surrogate-predicted. We assume $x_{sc}\!\equiv\!x_{cg}$ under doubly-symmetric thin-walled idealisation. As mentioned, we assume that the ribs do not contribute to stiffness and only compute their volume for mass estimation. Thus, stiffness stems from the main spar, as well as the skin. The skin is extremely thin but its contributions to torsional and bending stiffness properties cannot be neglected, so in data generation we use very fine resolutions ($\sim$ 5$\times$ skin thickness) when performing quadtree-based marching-squares in order to properly resolve the skin. 

\subsubsection{Helper Functions}

We use deterministic differentiable helper functions for planform and atmospheric conditions. The planform evaluation ($S_{\text{ref}}, b/2$) is a closed-form trapezoid integral over the four-station BWB chord distribution. The atmosphere helper is used to determine $\rho, q_\infty, \mathrm{Ma}, \mathrm{Re}$ using the ISA troposphere model (lapse rate to $11$\,km), in torch-native and autograd-safe formulations so all four quantities differentiate through $(h, V)$ (altitude and velocity). 

\subsubsection{Beam Quadrature, Mass, and Constraints}

\paragraph{Spanwise Sampling.} We use Sobol-quasi-random stations on $[0, b/2]$ per station query in each training epoch. Stations are sorted ascending and gradients flow through them via $b/2$ (the random draws themselves are detached).

\paragraph{Beam Integration.}
With the models described above, we obtain per-station distributed loads $q_z, m, x_{cp}$ from $\Phi^{\theta}_{\text{cp}}$ and section properties $A, I_{uu}, I_{vv}, J, u_{cg}, v_{cg}$ from the structural prediction $\Phi^{\theta}_{\text{sec}}$. 

First, we integrate the loads:

\[V(y) = \int_y^{y_{\text{tip}}} q_z\,dy'\]
\[M(y) = \int_y^{y_{\text{tip}}} V\,dy'\]
\[T(y) = \int_y^{y_{\text{tip}}} t(y')\,dy'\]

with distributed torsion $t = q_z\,(x_{cp}-u_{cg}) + m$ under the doubly-symmetric assumption $x_{sc} \equiv x_{cg}$. 

To determine stress components per station, we use bending:
\[\sigma_b = |M|\,z_{\max}/I_{uu} \]
with $z_{\max} = \sqrt{3}\sqrt{I_{uu}/A}$ (rectangular-section equivalent fibre distance, used as a placeholder rather than a surrogate output), transverse shear $\tau_s = |V|\,Q_{\max}/(I_{uu}\,t_{\text{skin}})$, and Bredt torsion $\tau_t = |T|/(2 A_{\text{enc}}\,t_{\text{skin}})$ with $A_{\text{enc}} = 20\,A$ pinned as a coarse thin-walled wingbox placeholder. We use composite von~Mises stress for the \ac{CFRP} material: $\sigma_{vm} = \sqrt{\sigma_b^2 + 3(\tau_s + \tau_t)^2}$, summing the two shear contributions at a common point as a conservative envelope. Tip deflection is evaluated by double forward cumulative trapezoid of $M/(E\,I_{uu})$ from the cantilever root, returning the full $w(y)$ profile. The benchmark reads $w_{\text{tip}} = w(y_{\text{tip}})$ for the tip-deflection constraint. In terms of materials, we assume quasi-isotropic CFRP laminate with $E=55$\,GPa, $\rho=1600$\,kg/m$^3$, allowable strain $\varepsilon_{\text{allow}}=4\!\times\!10^{-3}$. We are aware that these are gross simplifications, but we consider the above choices adequate for the application demo at hand.

\paragraph{Structural Mass.} We use trapezoid: $m_{\text{struct}} = \rho_{\text{mat}} \int_0^{b/2} A_{\text{phys}}(y)\,dy$.
For the range computation, we use battery mass: $m_{\text{bat}} = \rho_{\text{bat}}\,V_{\text{bat}}$ with $V_{\text{bat}} = (w\,d\,h)\,L^3$ from the 6-D battery design vector $\mathbf{x}_{\text{bat}} = (c_x, c_y, c_z, w, d, h)$ (the first three slots are placement coordinates, the last three are the box dimensions in unit-scale ratios). We use $\rho_{\text{bat}} = 2400$\,kg/m$^3$ for an advanced silicon-anode/NMC pack. For the sake of simplicity and because the battery tends to be small, we do not evaluate overlap constraints between the battery and internal structures. Overall mass is then evaluated as: 

\[m_{\text{tot}} = m_{\text{struct}} + m_{\text{bat}} + m_{\text{payload}}\] with payload pinned at 3\,kg.

\subsubsection{Constraints.}
Overall, there are three constraints evaluated with the models described above. First, we use a lift balance equality constraint:
\[(W - L_{\text{lift}})/W = 0\]
with $W = m_{\text{tot}}\,g $ and $L_{\text{lift}} = C_L\,q_\infty\,S_{\text{ref}}$.
    
This single constraint is what implicitly bounds battery mass: a heavier battery raises $W$ and forces the geometry to produce more lift, producing more drag, reducing the benefit in range. The cruise angle of attack $\alpha_{cr}$ is independently restricted to $\pm 3^{\circ}$ as a hard design-vector box bound (not an inequality constraint).

Next, we evaluate a maximum material strain constraint, which is evaluated as per-station residuals $c_i = \sigma_{vm}(y_i)/(E\,\varepsilon_{\text{allow}}) - 1$, and aggregated as $\tfrac{1}{N}\sum_i \mathrm{ReLU}(c_i)$. The aggregate is exactly zero if every station satisfies its strain bound, and the $K{=}1$ formulation gives a smaller Jacobian than one constraint per station, which eases compute requirements drastically for projection-based methods as described in \S~\ref{sec:experiments}.

Lastly, we constrain tip deflection: $|w_{\text{tip}}|/(0.10\cdot b/2) - 1 \le 0$.

\subsubsection{Objective}

For the objective, we use pure Breguet range:

\[R = (\eta_{\text{prop}}/g)\,E^*_{\text{bat}}\,(L/D)\,(m_{\text{bat}}/m_{\text{tot}})\]
with $\eta_{\text{prop}}=0.8$, specific energy $E^*_{\text{bat}} = 300$\,Wh/kg, and $L/D = C_L/C_D$ from $\Phi^{\theta}_{\text{aero}}$. The reported objective is
$-R/R_{\text{ref}}$ with $R_{\text{ref}}=10^7$\,m, sized so $\nabla\!\text{obj}$ stays $\mathcal{O}(1)$. $R_{\text{ref}}$ is a normalization constant for gradient scaling, not a physical target. Achieved $R$ during optimization is roughly an order of magnitude smaller.

\subsection{Urban Planning}
\label{app:b102}

\begin{figure}[h!]
  \centering
  \includegraphics[width=\linewidth]{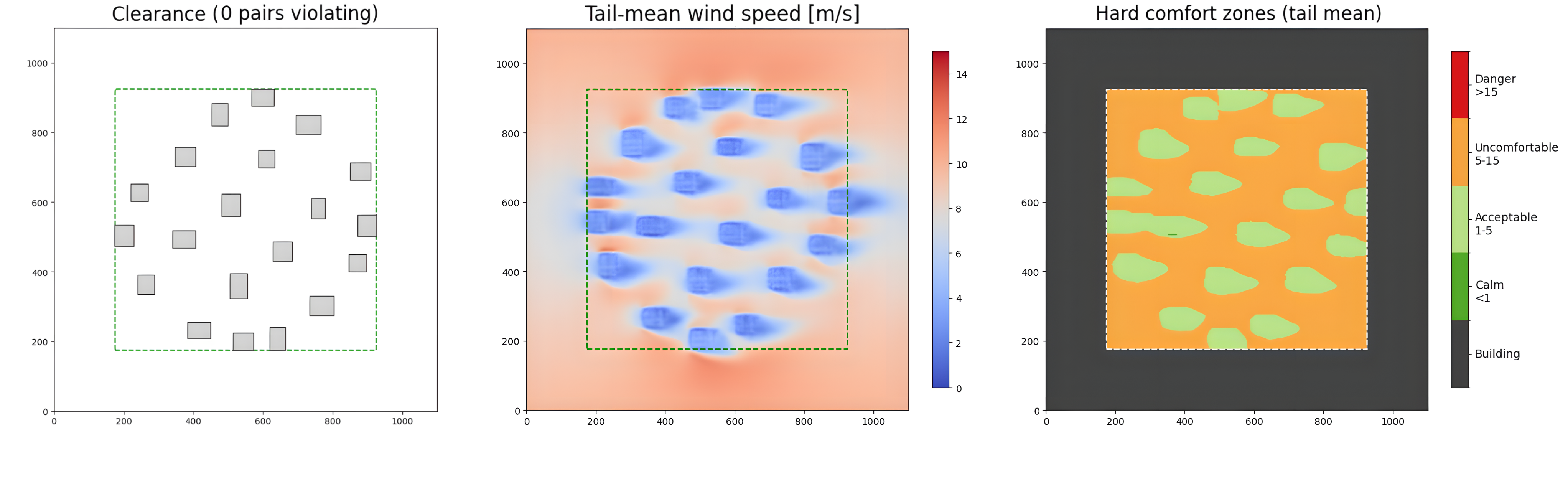}
  \caption{\texttt{E2} urban planning benchmark, three views of a candidate design.
  \textbf{Left:} top-down site plan with the $750\times750\,$m buildable plot
  (green dashed) and the 10 building footprints. Clearance is evaluated
  pairwise and reported as the number of pairs with overlapping
  $d_{\min}$-inflated AABBs. \textbf{Middle:} tail-mean wind speed predicted
  by the WinDiNet surrogate over the full $1100\times1100\,$m raster domain.
  \textbf{Right:} hard pedestrian-comfort zones derived from the tail-mean
  speed (calm $<1\,$m/s, acceptable $1$--$5$, uncomfortable $5$--$15$, danger
  $>15$). The danger constraint forbids any cell entering the red band.}
  \label{fig:b102_appendix}
\end{figure}

\begin{figure}[h!]
  \centering
  \includegraphics[width=0.9\linewidth]{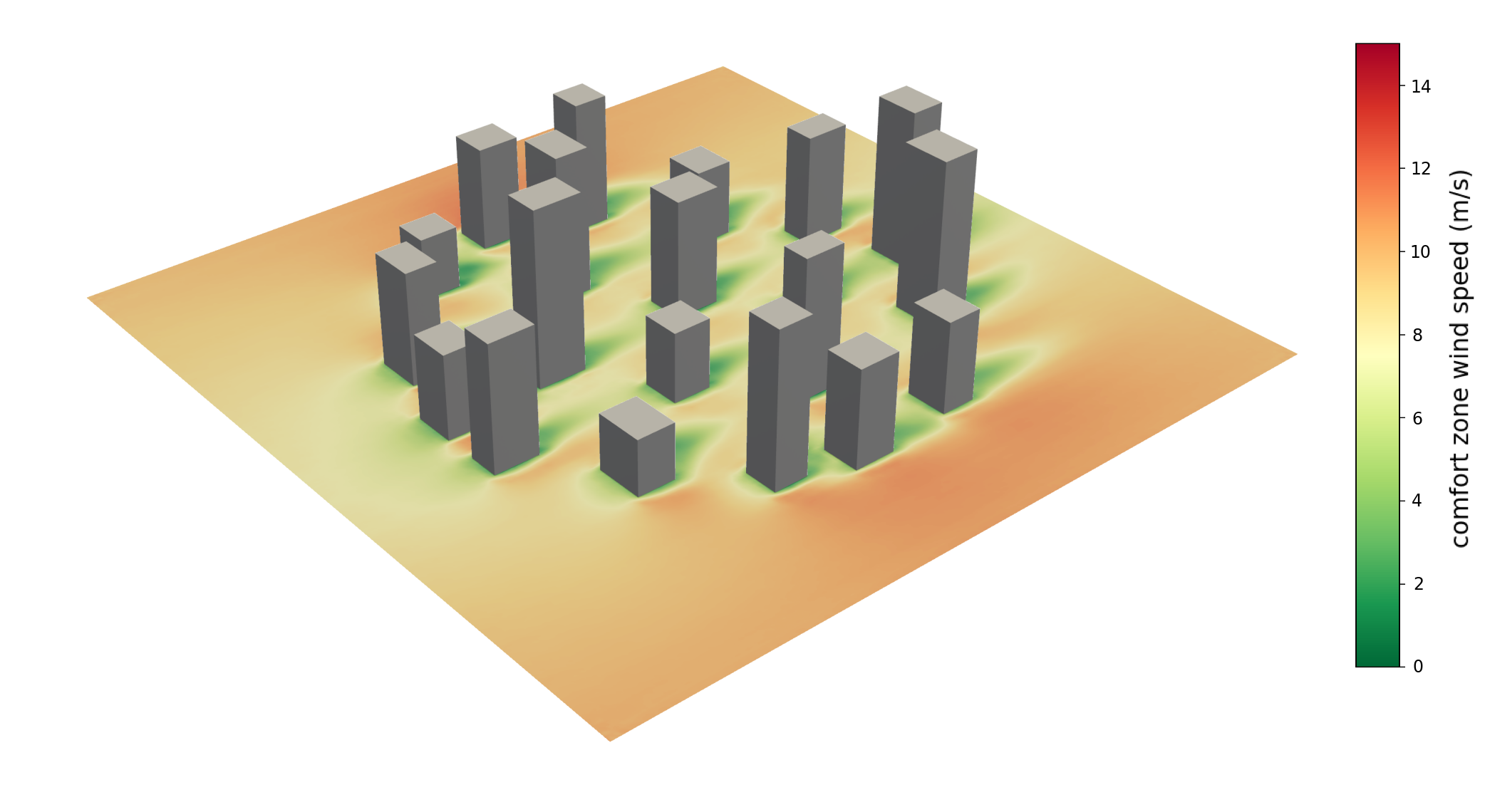}
  \caption{\texttt{E2} candidate design rendered in 3D with PyVista~\citep{sullivan2019pyvista}, with the pedestrian-comfort
  field draped at ground level. Building heights are decoded from the design
  vector. The colour map shows tail-mean wind speed over the comfort range
  $0$--$15\,$m/s, with green/yellow indicating calm and acceptable zones and
  red indicating the danger band that the surrogate constraint forbids.}
  \label{fig:b102_3d_comfort}
\end{figure}

\begin{figure}[h!]
  \centering
  \resizebox{\linewidth}{!}{\begingroup
\tikzset{
  font=\small,
  node distance=6mm and 10mm,
  box/.style={draw, rounded corners=1pt, minimum height=6mm, inner sep=3pt, align=center},
  geo/.style={box, densely dashed},
  ana/.style={box},
  sur/.style={box, fill=black!10},
  gen/.style={box, pattern=north east lines, pattern color=black!35},
  res/.style={box, double, double distance=0.6pt},
  arr/.style={-{Latex[length=2mm]}, semithick},
  lbl/.style={font=\scriptsize, fill=white, inner sep=1pt},
  leg/.style={font=\scriptsize, minimum height=4mm, minimum width=6mm, inner sep=1pt, rounded corners=1pt},
}
\tikzset{
  inp/.style={font=\small, inner sep=2pt, align=right},
  outv/.style={font=\normalsize, inner sep=2pt},
  bus/.style={semithick},
}
\begin{tikzpicture}[x=1mm,y=1mm]
\node[inp] (in1) at (0,3.5)  {$c_x, c_y$};
\node[inp] (in2) at (0,-3.5) {$w_{\mathrm{raw}}, d_{\mathrm{raw}}, h_{\mathrm{raw}}$};
\draw[decorate,decoration={brace,amplitude=1.5mm,mirror}] ([xshift=-1.5mm]in2.west|-in1.north) -- ([xshift=-1.5mm]in2.west|-in2.south) node[midway,left=2mm,rotate=90,anchor=south]{Design $y$};
\node[font=\scriptsize, below=0mm of in2] {10 buildings};
\node[ana] (dec)  at (24,0)   {Softplus\\decoding};
\node[geo] (mask) at (52,16)  {Occupancy\\mask};
\node[gen] (wind) at (79,16)  {WinDiNet};
\node[ana] (tail) at (105,16)  {Tail-mean\\field};
\node[res] (danger) at (131,16)  {Danger};
\node[res] (clear)  at (131,6)   {Clearance};
\node[res] (site)   at (131,-5)  {Site, Height,\\Coverage};
\node[res] (vol)    at (131,-16) {Volume};
\node[outv] (c) at (147,6)   {$c$};
\node[outv] (f) at (147,-16) {$f$};
\draw[arr] (in1.east) -- ++(3,0) |- (dec.west);
\draw[arr] (in2.east) -- ++(3,0) |- (dec.west);
\draw[arr] (dec.east) -- node[lbl,above]{boxes} ++(8,0) |- (mask.west);
\draw[arr] (dec.east) -- ++(8,0) |- (clear.west);
\draw[arr] (dec.east) -- ++(8,0) |- (site.west);
\draw[arr] (dec.east) -- ++(8,0) |- (vol.west);
\draw[arr] (mask) -- node[lbl,above]{$M$} (wind);
\draw[arr] (wind) -- node[lbl,above]{$u, v$} (tail);
\draw[arr] (tail) -- node[lbl,above]{$\bar s$} (danger);
\draw[arr] (mask.north) -- ++(0,5) -| node[lbl,pos=0.25,above]{$M$} (danger.north);
\draw[bus] (danger.east) -- (143,16) coordinate (bd);
\draw[bus] (site.east) -- (143,-5) coordinate (bs);
\draw[bus] (bd) -- (bs);
\draw[arr] (clear.east) -- (c);
\draw[arr] (vol.east) -- (f);
\end{tikzpicture}
\endgroup}
  \caption{\texttt{E2} evaluator $\Phi$, from the design vector $y$ to the objective $f$ and constraints $c$. Dashed: geometry processing. Plain: analytical. Grey: neural surrogate. Hatched: generative model. Double border: constraint or objective.}
  \label{fig:e2_evaluator}
\end{figure}
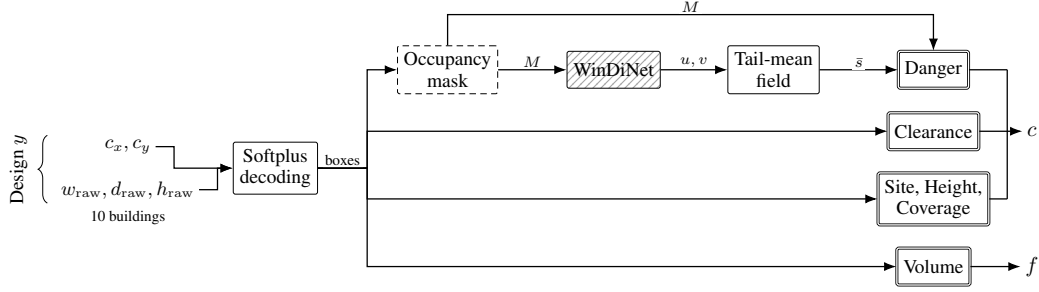

\paragraph{Setup.}
Fig.~\ref{fig:e2_evaluator} shows the arrangement of the evaluator. Place $B=10$ rectangular buildings on a $750\times750\,$m site to maximise total
built volume under spacing, height, coverage, and pedestrian wind comfort constraints. The
design variable $y$ is $50$D, with each building contributing $(c_x, c_y, w_{\mathrm{raw}}, d_{\mathrm{raw}}, h_{\mathrm{raw}})$ and a softplus decoding on $w, d, h$ to keep dimensions positive. The objective is $-\sum_i w_i\, d_i\, h_i / V_{\mathrm{ref}}$ with $V_{\mathrm{ref}}=10^6\,$m$^3$, which keeps the loss in $\mathcal{O}(1)$.

\paragraph{Wind Surrogate.}
The pedestrian-comfort constraint is evaluated through WinDiNet~\citep{perini2026windinet}, an LTX-Video 2B~\citep{hacohen2024ltxvideo} diffusion backbone fine-tuned to roll out 2D pedestrian-level velocity fields over the full $1100\times1100\,$m raster domain, conditioned on a building-footprint occupancy mask and a fixed inlet velocity. The surrogate is differentiable end-to-end, so gradients flow from each constraint back to the design vector as has been shown by \citet{perini2026windinet}. We run $57$ diffusion frames at $2$ inference steps and score comfort on the
tail-mean field over the last $10$ frames, $\bar{s}(p) = \lVert (\bar u(p), \bar v(p)) \rVert_2$.

\paragraph{Constraints.} All five constraints are inequalities with per-constraint tolerance $10^{-4}$.
\begin{itemize}
  \item \textbf{Site} (footprint inside $[175, 925]\,$m in y- and x-direction): softplus-smoothed
    squared-mean over the $4N$ box-edge primitives,
    \[
      c_{\mathrm{site}} = \frac{1}{4N}\sum_{i,e}
        \mathrm{softplus}_\beta(p_{i,e})^2,
    \]
    with $p_{i,e}\in\{L_0 - x_i^-,\ x_i^+ - L_1,\ L_0 - y_i^-,\ y_i^+ - L_1\}$.
  \item \textbf{Height} (cap at $h_{\max}=250\,$m):
    $c_{\mathrm{height}} = \tfrac{1}{B}\sum_i \mathrm{relu}(h_i - h_{\max})$.
  \item \textbf{Coverage} (site-clipped footprint fraction $\le 0.5$):
    $c_{\mathrm{cov}} = \mathrm{relu}\!\big(\tfrac{1}{A_{\mathrm{site}}}\sum_i A_i^{\mathrm{clip}} - 0.5\big)$.
  \item \textbf{Danger} (no fluid cell exceeds $15\,$m/s in the tail mean): a
    fluid-mask--weighted mean of $\mathrm{relu}(\bar s - 15)$ over inner-plot
    fluid cells, multiplied by $w_d=100$ to bring the typical excess into the
    $\mathcal{O}(10^{-2})$ band. In raw m/s the cell-averaged excess sits below
    the $10^{-4}$ tolerance even when the integrated violation is non-trivial.
  \item \textbf{Clearance} (no pair within $d_{\min}=10\,$m): kept as
    $\binom{B}{2}=45$ \emph{per-pair} constraints,
    \[
      c_{ij} =
      \mathrm{relu}\!\big(\tfrac{w_i+w_j}{2} + \tfrac{d_{\min}}{2} - |c_x^{(i)}-c_x^{(j)}|\big)
      \cdot
      \mathrm{relu}\!\big(\tfrac{d_i+d_j}{2} + \tfrac{d_{\min}}{2} - |c_y^{(i)}-c_y^{(j)}|\big),
    \]
    i.e., the m$^2$ overlap of the $d_{\min}$-inflated AABBs of pair $(i,j)$.
\end{itemize}
We now provide some more details on why the last of these constraints, \textbf{Clearance}, was evaluated pairwise instead of aggregated into a mean or max score like the other constraints in this example. 

\paragraph{Details on Pairwise Clearance Constraints.}
Aggregating the 45 pairwise overlaps into a single mean-pooled scalar weakened training across methods, which we believe is due to \emph{gradient cancellation}: a building wedged between violating neighbours on opposite sides receives
near-opposite push vectors that partially cancel under mean aggregation, so the gradient with respect to its center $(c_x, c_y)$ is small even though the design is grossly infeasible. The per-pair formulation gives each pair its own multiplier and its own Jacobian row, restoring directional information and per-pair pressure. However, this comes at a cost, especially for projection-based methods: With $B$ buildings, the number of \textbf{Clearance} constraints is $\frac{1}{2}B(B-1)$. Projection-based methods do need to build the constraint Jacobian $J$ on each projection step, the cost of which scales with $K$ (the overall number of constraints). However, in this case, this additional cost is attenuated by the fact that most constraints are such that only the graph relating to the \textbf{Clearance} constraint needs to be backpropagated back through, i.e., the expensive diffusion model for wind prediction is only backpropagated through once (for the \textbf{Danger} constraint), the cost of which dominates.

\subsection{{Power Grid Benchmark}}
\label{app:b104}

\texttt{E3} is the AC Optimal Power Flow (ACOPF) benchmark, instantiated on the IEEE 57-bus system. Apart from the differences we list here, it is analogous to the application example in DC3~\citep{donti2021dc3}. ACOPF determines generator setpoints that meet bus-level demand at minimum generation cost while respecting the nonlinear physics of the AC power grid.

The conditioning vector $x = (p_d, q_d) \in \mathbb{R}^{2L}$ specifies real and reactive power demand at each load bus, and the design vector $y = (p_g, q_g, v_m, v_a) \in \mathbb{R}^{2G + 2B}$ collects generator setpoints together with bus voltages in polar form (magnitude $v_m$, angle $v_a$).
For IEEE-57 ($B\!=\!57$ buses, $G\!=\!7$ generators, $E\!=\!80$ branches), this gives $\dim(y) = 128$. The objective is the standard quadratic generation cost,
\begin{equation}
f(y) = \sum_{i=1}^{G} \left( c_{0,i} + c_{1,i}\, p_{g,i} + c_{2,i}\, p_{g,i}^{2} \right),
\end{equation}
normalised by $\bar{g}^{\,2}$, where $\bar{g}$ is the mean generator base power. This applies DC3's normalisation formula (the absolute scale differs slightly because pandapower's per-generator MBASE for the slack bus deviates from PYPOWER's). The equality constraints enforce real and reactive power balance at every bus ($2B = 114$ residuals, nonlinear in $(v_m, v_a)$). The inequality constraints box-bound generator outputs ($p_g, q_g$), bus voltage magnitudes, branch apparent power flows ($|S_{ij}|$ at both ends of each branch), and per-branch voltage angle differences. In total, we count 114 equalities and 462 inequalities, i.e., $K=576$ constraints overall.

\paragraph{Differences from DC3 Implementation.} Our variant of this benchmark is more difficult than the one in \citet{donti2021dc3}, with a larger number of constraints. DC3's released code only enforces generator and voltage box bounds, $4G + 2B = 142$ inequalities. We additionally retain MATPOWER's~\citep{matpower2011} thermal line limits and angle-difference bounds, $4E = 320$ extra inequalities, bringing the total to $462$. On IEEE-57 these added slots are mostly inactive: the 160 angle-difference residuals are vacuous because the case data marks every branch as angle-unconstrained ($\pm 360^\circ$) and our output box bounds each $v_a \in [-\pi, \pi]$, so $|v_{a,i} - v_{a,j}| \le 2\pi$ holds by construction. The 160 thermal residuals are well-defined but slack ($S_{ij}^{\max} = 99$~pu after $\texttt{baseMVA}$ normalisation), so they almost never bind on sampled OPF solutions. They do become active on extreme off-manifold predictions and therefore still contribute to the feasibility metric. For grid data we use $\texttt{pandapower}$~\citep{pandapower.2018} and for constraint residuals we use $\texttt{ML4OPF}$~\citep{ml4opf}, whereas DC3 calls PYPOWER~\citep{pypower} directly. The underlying MATPOWER~\citep{matpower2011} cases and admittance physics are aligned to within ${\sim}10^{-5}$~pu, but pandapower's per-generator MBASE and slack-bus bound conventions differ slightly from PYPOWER's, requiring small loader-side adjustments. Our condition pool is regenerated under our loader: per-sample correlated uniform load perturbations around the nominal demand, retaining only samples for which an offline ACOPF solve with thermal limits relaxed converges. Every method evaluated here, including the DC3 baseline, draws its conditions from this same pool, so the head-to-head comparisons in this paper are internally fair in terms of conditioning.

\paragraph{Cost and Hardware.} With $K = 576$ ($114$ equalities and $462$ inequalities), this is the largest constraint count among our engineering benchmarks. Per training step, \ours{}'s cost is dominated by Jacobian assembly, one backward pass through $\Phi$ per constraint row, while DC3 assembles the Jacobian over the $114$ equalities only but takes several inner correction steps per training step, which results in higher wall-clock times overall. ALM's per-step cost carries no dependence on $K$ (one forward and one backward through $\Phi$), which is why it trains in about a minute. \ours{} finishes roughly $2.5$ times faster than DC3. This benchmark was run on a 288-core ARM CPU node. Cross-method ratios are reliable, but absolute wall-clock numbers are hardware-specific.

\section{Failure Modes of Baselines in Engineering Benchmarks}
\label{app:failure-modes}

This appendix collects some of our hypotheses for the failure modes observed on \texttt{E1}--\texttt{E3} for the baselines. We do not study each failure in depth.

\paragraph{DC3 on \texttt{E1} and \texttt{E2}.} DC3 is structurally inapplicable here. Its completion step solves the equality constraints for a dependent subset of the outputs and requires a known dependence on that subset. The single equality constraint of \texttt{E1} and \texttt{E2} is a neural surrogate of the full design vector, so no such subset exists, and DC3 would reduce to a non-adaptive penalty loss followed by gradient-correction steps. We therefore do not run it on \texttt{E1} and \texttt{E2}.

\paragraph{IPOPT on \texttt{E2}.} No IPOPT restart converges on \texttt{E2}. The best iterate of every restart is feasible, but only $31\%$ of final iterates are, and all restarts terminate with a local-infeasibility status. The aggregated constraints of \texttt{E2} apply ReLU transforms, so the Jacobian rows of inactive constraints vanish identically and the constraint qualification fails at most iterates ($95.6\%$ of Jacobian rows are zero at the final iterates, and the remaining rows span roughly $35$ orders of magnitude). We speculate that this is the pathology \citet{yang2022nnsurrogates} report for IPOPT on embedded ReLU networks, where dual infeasibility is never driven to zero and the solver terminates at spurious points. On \texttt{E1}, IPOPT reaches feasible points in every restart but also never exits with a converged status.

\paragraph{SnareNet on \texttt{E3}.} On \texttt{E3}, the number of constraints $K = 576$ is much larger than $\dim(y) = 128$. We believe that SnareNet was conceived around $K \leq \dim(y)$, since its convergence proofs require a constraint Jacobian of full row rank \citep{chu2026snarenet}. In Appendix~G of the original SnareNet paper, \citet{chu2026snarenet} show experiments where SnareNet is trained under $K>\dim(y)$. They show that, on synthetic non-convex programs with $K=500$ and $\dim(y) = 100$, SnareNet exhausts its projector budget of 100 steps without converging. However, we have no intuition for why this occurs, as \citet{chu2026snarenet} do not elaborate on the suspected reasons for this failure beyond noting that SnareNet suits problems with $K \ll \dim(y)$. On \texttt{E3}, each repair step reduces the relaxed residual by roughly $2\%$, so we cap the repair budget at $5$ steps, and all ten seeds diverge to non-finite losses within the first $300$ epochs.

\paragraph{ENFORCE on \texttt{E3}.} ENFORCE provides an optional gate during training. This gate rejects any projection step in training if it does not lower the soft loss. We find that, without the gate, training diverges catastrophically on \texttt{E3}, with all ten seeds producing a non-finite loss in the first epoch. The gate is therefore needed to stabilize training. With the gate enabled, however, we observe that all projection steps on \texttt{E3} are rejected. Effectively, ENFORCE then trains without any projection awareness. At test time, the projection is applied regardless, and we then observe divergence, resulting in numerical errors on $2$ to $13$ of the $64$ test queries per seed.

\paragraph{FSNet on \texttt{E3}.} FSNet seeks feasibility by minimizing the summed constraint violation with L-BFGS. We speculate that this is much weaker than projection on \texttt{E3} because of the extremely high number of constraints. Amalgamating them into a single score discards the per-constraint information that a full Jacobian retains. Additionally, the single score assumes somewhat comparable scales across constraints, but \ac{ACOPF} constraints (voltages, thermal limits, \ldots) have mixed units and potentially very different scales. The FSNet paper itself does not use L-BFGS on its \ac{ACOPF} benchmark but switches to \ac{LM} for the feasibility-seeking step \citep[Appendix~B]{nguyen2025fsnet}, which requires the constraint Jacobian in every step. This variant is not part of the public code, so our port runs the generic L-BFGS solver on \texttt{E3}. We read the authors' switch as confirmation that first-order feasibility seeking is not viable on \texttt{E3}.

\paragraph{ENFORCE on \texttt{E1} and \texttt{E2}.} We run variants with and without penalty-method warm-up. On \texttt{E1} and \texttt{E2}, and we used the IFT variant for better memory efficiency, but find it does not reduce peak memory usage. $39$ of $40$ seeds run \ac{OOM} while assembling this Jacobian. On \texttt{E1}, a single ENFORCE seed finishes training, but runs \ac{OOM} in evaluation due to a longer repair unroll.

\section{{Hyperparameter Tuning}}
\label{app:hp_tuning}

All baselines and \ours{} were tuned with Bayesian optimization on the synthetic suite \texttt{S1}--\texttt{S6}. Generally, the budget per method was set to $8$ Sobol and $24$ Bayesian-optimization trials, i.e., $32$ trials per method, with 3 seeds per trial. Two exceptions are ALM+Bolt-On (8+12 due to the very low-dimensional search space) and FSNet (stopped early due to plateau and very long run times). No sweep improved on its best trial after trial 24. The top three trials per method were then re-run with $8$ seeds, and the winner was selected lexicographically: highest feasibility, then lowest objective, then lowest constraint violation. The ten reported seeds in the results are disjoint from the tuning seeds and were never used in selection. A run counts as diverged if it crashes catastrophically or its post-repair objective exceeds $10^2$. Table~\ref{tab:hp-winners} shows the tuned hyperparameters per method. 

\begin{table}[H]
\centering
\small
\caption{{Selected hyperparameter values per method from the Bayesian optimization sweep. $\dagger$: For ALM+Bolt-On, the optimized hyperparameters of frozen ALM models were reused, i.e., only hyperparameters of the projection were tuned for ALM+Bolt-On itself.}}
\label{tab:hp-winners}
\begin{tabular}{@{}lp{9.6cm}@{}}
\toprule
\textbf{Method} & \textbf{BO winner configuration} \\
\midrule
ALM & \texttt{lr}~$8.3{\cdot}10^{-5}$; $\gamma$~$8.8{\cdot}10^{-4}$; $\alpha$~$0.995$; \texttt{eps}~$10^{-4}{}$ \\
ALM+Bolt-On & \texttt{proj\_delta}~$10^{-5}{}$; \texttt{proj\_max\_iters}~$20$ $\dagger$ \\
ENFORCE (v4) & \texttt{lr}~$2.5{\cdot}10^{-4}{}$; \texttt{eps\_chol}~$2.3{\cdot}10^{-5}$; \texttt{max\_it}~$100$; \texttt{training\_tolerance}~$0.01$; \texttt{weight\_loss\_displacement}~$5.0$; \texttt{weighting\_option}~$5$; \texttt{ada\_np\_auto\_activation}~true (fixed: \texttt{epoch\_start\_hard\_constrained}~$0$, \texttt{eps}~$10^{-8}$, \texttt{inference\_tolerance}~$10^{-6}$) \\
DC3 & \texttt{lr}~$1.75{\cdot}10^{-5}{}$; \texttt{soft\_weight}~$10$; \texttt{corr\_train\_steps}~$10$; \texttt{corr\_lr}~$1.8{\cdot}10^{-6}$ \\
FSNet & \texttt{lr}~$2.0{\cdot}10^{-4}$; \texttt{dist\_weight}~$5.0$; \texttt{weight\_decay}~$3.0{\cdot}10^{-5}$ \\
SnareNet & \texttt{lr}~$2.5{\cdot}10^{-4}{}$; $\lambda$~$2.6{\cdot}10^{-3}$; \texttt{soft\_weight}~$1.0$; \texttt{soft\_epochs}~$1000$; \texttt{decay\_epochs}~$1000$ \\
PAL [Ours] & \texttt{lr}~$7.6{\cdot}10^{-5}$; \texttt{rate}~$1.1{\cdot}10^{-3}{}$; \texttt{max\_decades}~$5.0$ \\
\bottomrule
\end{tabular}
\end{table}

\section{Reproducibility}
\label{app:engineering_benches_hyperparams_and_settings}

Here, we state the hardware used in our experiments and provide rough wall-clock times, and then show the hyperparameters used for \ours{} and ALM in the engineering benchmarks. Hyperparameters on the synthetic benchmarks come from the equal-budget Bayesian optimization sweep in Appendix~\ref{app:hp_tuning}. On \texttt{E1}--\texttt{E4} we generally use paper defaults. FSNet, SnareNet, and ENFORCE hyperparameters are not shown for the engineering benchmarks \texttt{E1} and \texttt{E2}. These were not tuned since these methods consistently fail immediately due to memory intractability anyway.

On \texttt{E3}, all learned methods follow the published DC3 schedule for \ac{ACOPF}, i.e., a $200{\times}2$ backbone, 1000 training samples, 2000 epochs, batch size 200, and learning rate $10^{-3}$. DC3 runs at its published hyperparameters, \ours{} at defaults with $\lambda_{\min} = 10^{-4}$. FSNet and ENFORCE otherwise run at their published defaults. SnareNet runs at its published defaults except for a repair budget of 5 steps, since its repair layer reduces the residual by roughly $2\%$ per step and 100 steps would exceed 200 hours per seed.

\subsection{Hardware and Wall-Clock}

Runs were generally executed on GH200 Grace-Hopper nodes. The engineering benchmarks \texttt{E1} and \texttt{E2} used 4$\times$\,GH200 per seed via DDP. Synthetic benchmarks (\texttt{S1}--\texttt{S6}), \texttt{E3}, \texttt{E4}, and ablations were run CPU-only. Wall-clock times for \texttt{E3} and \texttt{E4} are reported separately in Appendices~\ref{app:b104} and~\ref{app:b105}.

Tables~\ref{tab:approx-wallclock-synth} and~\ref{tab:approx-wallclock-engi} report median end-to-end wall-clock per run over 10 seeds ($n{=}5$ where noted in the corresponding results tables). These figures cover configuration, training, final evaluation, and artifact saving. Training dominates in every cell shown.

\begin{table}[h!]
\centering
\footnotesize
\setlength{\tabcolsep}{4pt}
\caption{Wall-clock times on \texttt{S1}--\texttt{S6}. The total time reported includes initialization, training, evaluation, and artifact saving. The overall prediction time is for 64 samples (queries). We report the median over $10$ seeds (counting non-diverged seeds only). ``---'' indicates fully diverged or structurally inapplicable cells. $^{\dagger}$For ALM+Bolt-On, training reuses ALM checkpoints and the reported total time is inference-time projection only.}
\label{tab:approx-wallclock-synth}
\begin{tabular}{@{}llrrrr@{}}
\toprule
& \textbf{Method} & \textbf{Train (s)} & \textbf{Total (s)} & \textbf{Predict (s)} & \textbf{$/$query (ms)} \\
\midrule
\multirow{7}{*}{\texttt{S1}}
  & \textbf{\ours} & 106 & 247 & 0.0119 & 0.185 \\
  & ALM & 58 & 131 & 0.00734 & 0.115 \\
  & ALM+Bolt-On & 58$^{\dagger}$ & 10 & 0.0507 & 0.792 \\
  & ENFORCE v4 & 57 & 63 & 0.0152 & 0.237 \\
  & FSNet & 1720 & 1909 & 0.176 & 2.75 \\
  & SnareNet & 149 & 310 & 0.0506 & 0.791 \\
  & DC3 & --- & --- & --- & --- \\
\midrule
\multirow{7}{*}{\texttt{S2}}
  & \textbf{\ours} & 118 & 261 & 0.018 & 0.282 \\
  & ALM & 65 & 136 & 0.00917 & 0.143 \\
  & ALM+Bolt-On & 65$^{\dagger}$ & 49 & 0.0627 & 0.98 \\
  & ENFORCE v4 & 104 & 110 & 0.0422 & 0.659 \\
  & FSNet & 1881 & 2074 & 0.0395 & 0.617 \\
  & SnareNet & 152 & 292 & 0.175 & 2.73 \\
  & DC3 & 261 & 417 & 0.0166 & 0.259 \\
\midrule
\multirow{7}{*}{\texttt{S3}}
  & \textbf{\ours} & 126 & 276 & 0.0159 & 0.249 \\
  & ALM & 71 & 126 & 0.00775 & 0.121 \\
  & ALM+Bolt-On & 71$^{\dagger}$ & 50 & 0.408 & 6.37 \\
  & ENFORCE v4 & 72 & 80 & 0.0448 & 0.700 \\
  & FSNet & 1593 & 1791 & 0.0174 & 0.272 \\
  & SnareNet & 216 & 277 & 0.156 & 2.44 \\
  & DC3 & 253 & 410 & 0.0213 & 0.333 \\
\midrule
\multirow{7}{*}{\texttt{S4}}
  & \textbf{\ours} & 179 & 339 & 0.0485 & 0.758 \\
  & ALM & 91 & 196 & 0.00776 & 0.121 \\
  & ALM+Bolt-On & 91$^{\dagger}$ & 52 & 0.367 & 5.74 \\
  & ENFORCE v4 & 183 & 189 & 0.950 & 14.8 \\
  & FSNet & 8045 & 8268 & 0.0688 & 1.08 \\
  & SnareNet & 460 & 483 & 0.788 & 12.3 \\
  & DC3 & --- & --- & --- & --- \\
\midrule
\multirow{7}{*}{\texttt{S5}}
  & \textbf{\ours} & 135 & 303 & 0.023 & 0.36 \\
  & ALM & 116 & 260 & 0.00744 & 0.116 \\
  & ALM+Bolt-On & 116$^{\dagger}$ & 34 & 0.094 & 1.47 \\
  & ENFORCE v4 & --- & --- & --- & --- \\
  & FSNet & 15772 & 15968 & 0.238 & 3.73 \\
  & SnareNet & 142 & 167 & 0.14 & 2.19 \\
  & DC3 & --- & --- & --- & --- \\
\midrule
\multirow{7}{*}{\texttt{S6}}
  & \textbf{\ours} & 367 & 516 & 0.332 & 5.19 \\
  & ALM & 77 & 227 & 0.0137 & 0.214 \\
  & ALM+Bolt-On & 77$^{\dagger}$ & 41 & 0.736 & 11.5 \\
  & ENFORCE v4 & 1243 & 1248 & 0.432 & 6.75 \\
  & FSNet & 1794 & 2042 & 0.271 & 4.23 \\
  & SnareNet & 1493 & 1542 & 5.79 & 90.5 \\
  & DC3 & --- & --- & --- & --- \\
\bottomrule
\end{tabular}
\end{table}

\begin{table}[h!]
\centering
\footnotesize
\setlength{\tabcolsep}{4pt}
\caption{End-to-end wall-clock time on engineering benchmarks. On \texttt{E2}, \ours{} suffers from sequential Jacobian assembly (loop over $K$).}
\label{tab:approx-wallclock-engi}
\begin{tabular}{@{}llr@{}}
\toprule
Benchmark & Method & Median wall-clock \\
\midrule
\texttt{E1} BWB        & \textbf{\ours}    & $\sim$8\,min   \\
                         & ALM               & $\sim$70\,s    \\
\midrule
\texttt{E2} Urban Planning & \textbf{\ours}    & $\sim$106\,min \\
                         & ALM+Bolt-On       & $\sim$65\,min  \\
                         & ALM               & $\sim$32\,s   \\
\bottomrule
\end{tabular}
\end{table}

\subsection{Hyperparameters}
For the neural backbone $N_\theta$, all methods use Adam with a simple \acs{MLP} $512{\times}4$ prediction head. ALM+Bolt-On shares ALM's training runs for the engineering benchmarks \texttt{E1} and \texttt{E2}. At inference, it adds a single Levenberg--Marquardt projector pass that uses \ours{}'s projector defaults ($N{=}10$, $\delta{=}10^{-3}$). The benchmark sets the residual gap target $\tau_c$ used by \ours{}'s log-gap rule ($\tau_c{=}10^{-3}$ on \texttt{E1} and $\tau_c{=}10^{-4}$ on \texttt{E2}). We treat $\tau_c$ as a property of the benchmark, not a method-specific hyperparameter. For decay, \checkmark{} marks the bidirectional log-gap update (the default, where multipliers $\bar w_{c,k}$ and $\bar w_d$ can decay), and \xmark{} marks monotone-only growth, which we enable only on \texttt{E1} since we observe moderate oscillations between $\bar w_c$ and $c(\tilde y^{(1)})$ which we want to avoid. We are unsure what causes these.

\begin{table}[h!]
\centering
\footnotesize
\setlength{\tabcolsep}{4pt}
\caption{Hyperparameters for \ours{} and ALM on \texttt{E1} and \texttt{E2}. Bold marks deviations from defaults. ALM defaults follow \citet{basir2023adaptive}. For both benchmarks, we performed minor hand-tuning of $\gamma_{\mathrm{ALM}}$ and $\alpha$ but found defaults to work relatively well.}
\label{tab:hparams-b101-b102}
\begin{tabular}{@{}lccc@{}}
\toprule
& Default & \texttt{E1} (Unmanned Aircraft) & \texttt{E2} (Urban Planning) \\
\midrule
Training epochs            & ---       & 200             & 500 \\
Batch size                 & ---       & 32              & 16 \\
Learning rate ($N_\theta$) & ---       & $10^{-4}$       & $10^{-3}$ \\
\midrule
\multicolumn{4}{c}{\textbf{\ours{}}}\\
\midrule
$\gamma$         & $10^{-2}$  & $\bm{10^{-1}}$  & $\bm{10^{-1}}$ \\
$D_{\max}$       & $1.0$      & $\bm{5.0}$      & $\bm{5.0}$ \\
$\tau_d$         & $10^{-4}$  & $10^{-4}$       & $10^{-4}$ \\
$\delta$         & $10^{-3}$  & $10^{-3}$       & $10^{-3}$ \\
$\lambda_{\min}$ & $10^{-6}$  & $10^{-6}$       & $10^{-6}$ \\
Decay            & \checkmark & \textbf{\xmark} & \checkmark \\
\midrule
\multicolumn{4}{c}{\textbf{ALM}}\\
\midrule
$\gamma_{\mathrm{ALM}}$ & $10^{-2}$ & $\bm{10^{-3}}$ & $10^{-2}$ \\
$\alpha$                & $0.99$    & $0.99$         & $0.99$ \\
$\mu_0$                 & $1.0$     & $1.0$          & $1.0$ \\
$\lambda_0$             & $1.0$     & $1.0$          & $1.0$ \\
\bottomrule
\end{tabular}
\end{table}

FSNet runs the authors' implementation with its default truncated unrolled differentiation, unrolling up to $50$ repair steps and backpropagating through the first $30$, which lies inside the recommended range ($10$ to $50$) reported in the FSNet ablations~\citep{nguyen2025fsnet}. Like this default, all remaining FSNet settings are left at the authors' values, apart from the tuned hyperparameters listed in Appendix~\ref{app:hp_tuning}.

\section{Compute Scaling}
\label{app:compute_scaling}
In this appendix, we provide some thoughts on how \ours{} performs on problems with a large number of constraints, and how complexity scales with problem dimensionality, the number of constraints, and constraint cost heterogeneity. Table \ref{tab:methods-sweep-synthetic-full} shows the full results table of the synthetic benchmarking results, including full $N_b ^\phi$ and $N_f ^\phi$, which Table \ref{tab:methods-sweep-synthetic} showed only excerpts of. 

\begin{table}[h]
\centering
\footnotesize
\setlength{\tabcolsep}{3pt}
\renewcommand{\arraystretch}{1.15}
\caption{Full version of Table~\ref{tab:methods-sweep-synthetic}: synthetic stress-test experiments for five baselines and \ours{}, including evaluator passes per training step on all benchmarks. Means and standard deviations across 10 seeds. ``---'' indicates failure on 2 or more seeds.}
\label{tab:methods-sweep-synthetic-full}
\begin{tabular}{@{}ll *{6}{M}@{}}
\toprule
 & \textbf{Bench.}
  & \makecell{\textbf{ALM+}\\\textbf{Bolt-On}} & \textbf{ENFORCE} & \textbf{DC3} & \textbf{FSNet} & \textbf{SnareNet} & \makecell{\textbf{PAL}\\{}[Ours]} \\
\midrule
\multirow{7}{*}{\rotatebox[origin=c]{90}{Feas.\ (\%)}}
  & \texttt{S1} & \cc{100} $100\pm0$ & \cc{100} $100\pm0$ & — & \cc{100} $100\pm0$ & \cc{100} $100\pm0$ & \cc{100} $100\pm0$ \\
  & \texttt{S2} & \cc{100} $100\pm0$ & \cc{100} $100\pm0$ & \cc{84} $97\pm7$ & \cc{100} $100\pm0$ & \cc{100} $100\pm0$ & \cc{100} $100\pm0$ \\
  & \texttt{S3} & \cc{100} $100\pm0$ & \cc{100} $100\pm0$ & \cc{100} $100\pm0$ & \cc{100} $100\pm0$ & \cc{100} $100\pm0$ & \cc{100} $100\pm0$ \\
  & \texttt{S4} & \cc{98} $100\pm1$ & \cc{100} $100\pm0$ & — & \cc{100} $100\pm0$ & \cc{100} $100\pm0$ & \cc{100} $100\pm0$ \\
  & \texttt{S5} & \cc{100} $100\pm0$ & — & — & \cc{4} $51\pm31$ & \cc{100} $100\pm0$ & \cc{100} $100\pm0$ \\
  & \texttt{S6} & \cc{100} $100\pm0$ & \cc{100} $100\pm0$ & — & \cc{100} $100\pm0$ & \cc{100} $100\pm0$ & \cc{100} $100\pm0$ \\
\cmidrule(lr){2-8}
  & \textit{Mean} & \cc{100} \best{$100$} & — & — & \cc{66} $92$ & \cc{100} \best{$100$} & \cc{100} \best{$100$} \\
\midrule
\multirow{7}{*}{\rotatebox[origin=c]{90}{Obj.}}
  & \texttt{S1} & \cc{44} $0.07\pm0.12$ & \cc{100} $0.00\pm0.00$ & — & \cc{100} $0.00\pm0.00$ & \cc{12} $0.78\pm0.10$ & \cc{55} $0.03\pm0.04$ \\
  & \texttt{S2} & \cc{26} $0.27\pm0.17$ & \cc{100} $0.00\pm0.00$ & \cc{100} $0.00\pm0.00$ & \cc{100} $0.00\pm0.00$ & \cc{33} $0.16\pm0.00$ & \cc{100} $0.00\pm0.00$ \\
  & \texttt{S3} & \cc{100} $0.00\pm0.00$ & \cc{100} $0.00\pm0.00$ & \cc{100} $0.00\pm0.00$ & \cc{100} $0.00\pm0.00$ & \cc{70} $0.01\pm0.00$ & \cc{100} $0.00\pm0.00$ \\
  & \texttt{S4} & \cc{0} $19.47\pm5.25$ & \cc{70} $0.01\pm0.01$ & — & \cc{100} $0.00\pm0.00$ & \cc{41} $0.09\pm0.01$ & \cc{100} $0.00\pm0.00$ \\
  & \texttt{S5} & \cc{100} $0.00\pm0.00$ & — & — & \cc{0} $4.41\pm2.16$ & \cc{100} $0.00\pm0.00$ & \cc{100} $0.00\pm0.00$ \\
  & \texttt{S6} & \cc{35} $0.14\pm0.03$ & \cc{100} $0.00\pm0.00$ & — & \cc{100} $0.00\pm0.00$ & \cc{16} $0.60\pm0.06$ & \cc{55} $0.03\pm0.01$ \\
\cmidrule(lr){2-8}
  & \textit{Mean} & \cc{0} $3.32$ & — & — & \cc{13} $0.73$ & \cc{26} $0.27$ & \cc{70} \best{$0.01$} \\
\midrule
\multirow{7}{*}{\rotatebox[origin=c]{90}{$N_\Phi^{\mathrm{f}}$}}
  & \texttt{S1} & \cc{100} $1.0\pm0.0$ & \cc{53} $6.2\pm0.1$ & — & \cc{0} $72\pm77$ & \cc{9} $35.0\pm5.0$ & \cc{82} $2.0\pm0.0$ \\
  & \texttt{S2} & \cc{100} $1.0\pm0.0$ & \cc{54} $6.0\pm0.0$ & \cc{10} $34.5\pm0.0$ & \cc{25} $18.5\pm0.1$ & \cc{23} $20.5\pm1.0$ & \cc{82} $2.0\pm0.0$ \\
  & \texttt{S3} & \cc{100} $1.0\pm0.0$ & \cc{58} $5.2\pm0.1$ & \cc{10} $34.5\pm0.0$ & \cc{41} $10.1\pm0.1$ & \cc{3} $44.1\pm5.9$ & \cc{82} $2.0\pm0.0$ \\
  & \texttt{S4} & \cc{100} $1.0\pm0.0$ & \cc{53} $6.4\pm0.0$ & — & \cc{0} $77.7\pm0.6$ & \cc{0} $61.9\pm2.8$ & \cc{82} $2.0\pm0.0$ \\
  & \texttt{S5} & \cc{100} $1.0\pm0.0$ & — & — & \cc{0} $595\pm50$ & \cc{0} $50.3\pm3.6$ & \cc{82} $2.0\pm0.0$ \\
  & \texttt{S6} & \cc{100} $1.0\pm0.0$ & \cc{50} $7.1\pm0.1$ & — & \cc{30} $15.5\pm0.0$ & \cc{0} $81.1\pm2.9$ & \cc{82} $2.0\pm0.0$ \\
\cmidrule(lr){2-8}
  & \textit{Mean} & \cc{100} \best{$1.0$} & — & — & \cc{0} $131$ & \cc{1} $48.8$ & \cc{82} $2.0$ \\
\midrule
\multirow{7}{*}{\rotatebox[origin=c]{90}{$N_\Phi^{\mathrm{b}}$}}
  & \texttt{S1} & \cc{100} $1.0\pm0.0$ & \cc{64} $4.1\pm0.1$ & — & \cc{39} $11.0\pm5.6$ & \cc{10} $33.6\pm5.0$ & \cc{82} $2.0\pm0.0$ \\
  & \texttt{S2} & \cc{100} $1.0\pm0.0$ & \cc{32} $14.1\pm0.0$ & \cc{21} $22.0\pm0.0$ & \cc{48} $7.8\pm0.1$ & \cc{20} $22.7\pm1.3$ & \cc{36} $12.0\pm0.0$ \\
  & \texttt{S3} & \cc{100} $1.0\pm0.0$ & \cc{49} $7.4\pm0.2$ & \cc{21} $22.0\pm0.0$ & \cc{56} $5.6\pm0.1$ & \cc{1} $47.7\pm6.7$ & \cc{50} $7.0\pm0.0$ \\
  & \texttt{S4} & \cc{100} $1.0\pm0.0$ & \cc{11} $32.5\pm0.2$ & — & \cc{6} $39.1\pm0.2$ & \cc{0} $96.9\pm5.3$ & \cc{16} $27.0\pm0.0$ \\
  & \texttt{S5} & \cc{100} $1.0\pm0.0$ & — & — & \cc{0} $87.8\pm1.0$ & \cc{0} $53.4\pm4.1$ & \cc{54} $6.0\pm0.0$ \\
  & \texttt{S6} & \cc{100} $1.0\pm0.0$ & \cc{0} $128\pm2$ & — & \cc{49} $7.2\pm0.0$ & \cc{0} $253\pm12$ & \cc{0} $91.0\pm0.0$ \\
\cmidrule(lr){2-8}
  & \textit{Mean} & \cc{100} \best{$1.0$} & — & — & \cc{16} $26.4$ & \cc{0} $84.6$ & \cc{19} $24.2$ \\
\bottomrule
\end{tabular}
\end{table}

\paragraph{Cost Model.} Per training step, \ours{} costs $2 C_f + (K+1) C_b$ evaluator passes, where $C_f$ and $C_b$ are the costs of one forward and one backward pass (VJP) through the evaluator $\Phi$, respectively. This is because two forward passes are needed, one before the projection at $\hat y$ and one after the projection at $\tilde y^{(1)}$. $K$ VJP backward passes are needed to build the Jacobian $J$, and one backward pass is needed to backpropagate the training loss (weighted constraints and the objective). 

For the special case of $K \gg \dim(y)$, $J$ can be assembled from forward-mode JVPs instead of reverse-mode VJPs, meaning the assembly cost is $\min(K, \dim(y))$ passes. Additionally, \ours{} requires $O(K^2 \dim(y) + K^3)$ for the damped normal system of the \ac{LM} projector. In the regime we target, the evaluator passes dominate and the latter is negligible. The recorded number of constraint evaluator calls in Table \ref{tab:methods-sweep-synthetic} fits this model exactly (e.g., 91 backward calls at $K=90$ on \texttt{S6}).

\paragraph{Comparison with Full-Unroll Projection Methods.} Every projection-based method (DC3, ENFORCE, SnareNet) assembles the same $K$-row Jacobian, but unrolled methods do so at each of $N$ repair iterates. This results in an overall cost of $N(C_f + K C_b)$ versus \ours{}'s single assembly. As disclosed in item (3) of the Limitations section, this cost is shared by all projection-based methods, and \ours{} is the cheapest member of that class.

\paragraph{Comparison with FSNet.} FSNet sidesteps Jacobian assembly and requires only $N(C_f + C_b)$, but unrolls many more repair steps (large $N$). The break-even point between \ours{} and FSNet is roughly at $K \approx 2N$. FSNet must still store the unrolled tape, which can be memory-intractable (see Table~\ref{fig:engineering_results}).

\paragraph{Heterogeneous Constraint Complexity.} In some applications, one or more constraints might be extremely expensive to compute and dominate compute cost. For example, \texttt{E2}'s wind-speed danger constraint shown in Appendix~\ref{app:b102} is evaluated using diffusion model unrolls, while other constraints in this problem are trivial to compute (e.g., box-to-box overlaps, analytical closed-form constraints, etc.). In this situation, the Jacobian assembly count $K$ virtually drops out of \ours{}'s cost while FSNet still pays the full unroll.

Other than cost, we do not see any specific reasons \ours{} might struggle with large-$K$. Benchmark \texttt{E3} in Table~\ref{fig:engineering_results} illustrates that \ours{} can be applied without any modifications to applications with a very high number of constraints ($K=576$) and trains faster than DC3 thanks to single-step training, despite building the full Jacobian (which DC3 does not, as it only projects equalities).

\end{document}